\documentclass{article}

\usepackage{arxiv}

\usepackage[utf8]{inputenc} 
\usepackage[T1]{fontenc}    
\usepackage{hyperref}       
\usepackage{url}            
\usepackage{booktabs}       
\usepackage{amsfonts}       
\usepackage{nicefrac}       
\usepackage{microtype}      
\usepackage{lipsum}		
\usepackage{graphicx}
\usepackage{doi}

\title{TMBuD: A dataset for urban scene building detection}

\author{ \href{https://orcid.org/0000-0002-0071-958X}{\includegraphics[scale=0.06]{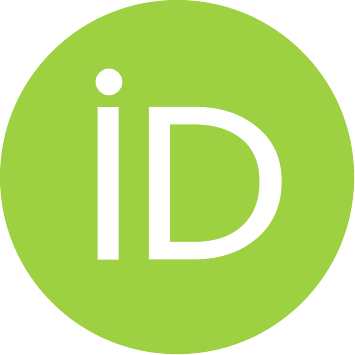}\hspace{1mm}Ciprian~Orhei} \\
	Politehnica University of Timi\c{s}oara\\
	Timi\c{s}oara, Romania\\
	\texttt{ciprian.orhei@cm.upt.ro} \\
	\And
	\href{https://orcid.org/0000-0003-2394-4859}{\includegraphics[scale=0.06]{orcid.pdf}\hspace{1mm}Silviu~Vert} \\
	Politehnica University of Timi\c{s}oara\\
	Timi\c{s}oara, Romania\\
	\texttt{silviu.vert@upt.ro} \\
	\And
	\href{https://orcid.org/0000-0002-0203-1519}{\includegraphics[scale=0.06]{orcid.pdf}\hspace{1mm}Muguras~Mocofan} \\
	Politehnica University of Timi\c{s}oara\\
	Timi\c{s}oara, Romania\\
	\texttt{muguras.mocofan@upt.ro} \\
	\And
	\href{https://orcid.org/0000-0003-1185-1997}{\includegraphics[scale=0.06]{orcid.pdf}\hspace{1mm}Radu~Vasiu} \\
	Politehnica University of Timi\c{s}oara\\
	Timi\c{s}oara, Romania\\
	\texttt{radu.vasiu@upt.ro} \\
}

\renewcommand{\shorttitle}{Accepted in 27th International Conference, ICIST 2021}

\hypersetup{
pdftitle={TMBuD: A dataset for urban scene building detection},
pdfsubject={cs.CV, cs.AI, cs.ML},
pdfauthor={Ciprian~Orhei, Silviu~Vert, Muguras~Mocofan, Radu~Vasiu},
pdfkeywords={Building dataset, facade detection, edge detection, semantic segmentation, edge detection ground-truth, semantic segmentation ground-truth},
}

\begin{document}
\maketitle

\begin{abstract}
Building recognition and 3D reconstruction of human made structures in urban scenarios has become an interesting and actual topic in the image processing domain. For this research topic the Computer Vision and Augmented Reality areas intersect for creating a better understanding of the urban scenario for various topics. In this paper we aim to introduce a dataset solution, the TMBuD, that is better fitted for image processing on human made structures for urban scene scenarios. The proposed dataset will allow proper evaluation of salient edges and semantic segmentation of images focusing on the street view perspective of buildings. The images that form our dataset offer various street view perspectives of buildings from urban scenarios, which allows for evaluating complex algorithms. The dataset features 160 images of buildings from Timi\c{s}oara, Romania, with a resolution of 768 x 1024 pixels each.


\end{abstract}
\section{Introduction}

Computer Vision (CV) aims to create computational models that can mimic the human visual system. From an engineering point of view, CV aims to build autonomous systems which could perform some of the tasks that the human visual system is able to accomplish \cite{huang1996computer}.

Urban scenarios reconstruction and understanding of it is an area of research with several applications nowadays: entertainment industry, computer gaming, movie making, digital mapping for mobile devices, digital mapping for car navigation, urban planning, driving. Understanding urban scenarios has become much more important with the evolution of Augmented Reality (AR). AR is successfully exploited in many domains nowadays, one of them being culture and tourism, an area in which the authors of this paper carried multiple research projects \cite{vert2014relevant}, \cite{vert2017augmented}, \cite{vert2019augmented}.

Automatic urban scene object recognition describes the process of segmentation and classification of buildings, trees, cars and so on. This job is done using a fixed number of categories on which a model is trained for classifying scene components \cite{babahajiani2014object}. Object detection, recognition and estimation in 3D images have gained momentum due to the availability of more complex sensors and an increase in large scale 3D data. Visual recognition of buildings can be a problematic task due to image distortions, image saturation or obstacles that are blocking the line of sight. The assumption that local shape structures are sufficient to recognise objects and scenes is largely invalid in practice since objects may have a similar shape \cite{fu2014indoor}.

\begin{figure}[h!]
\centering
\begin{tabular}{ccc}
\includegraphics[height=0.1\textheight]{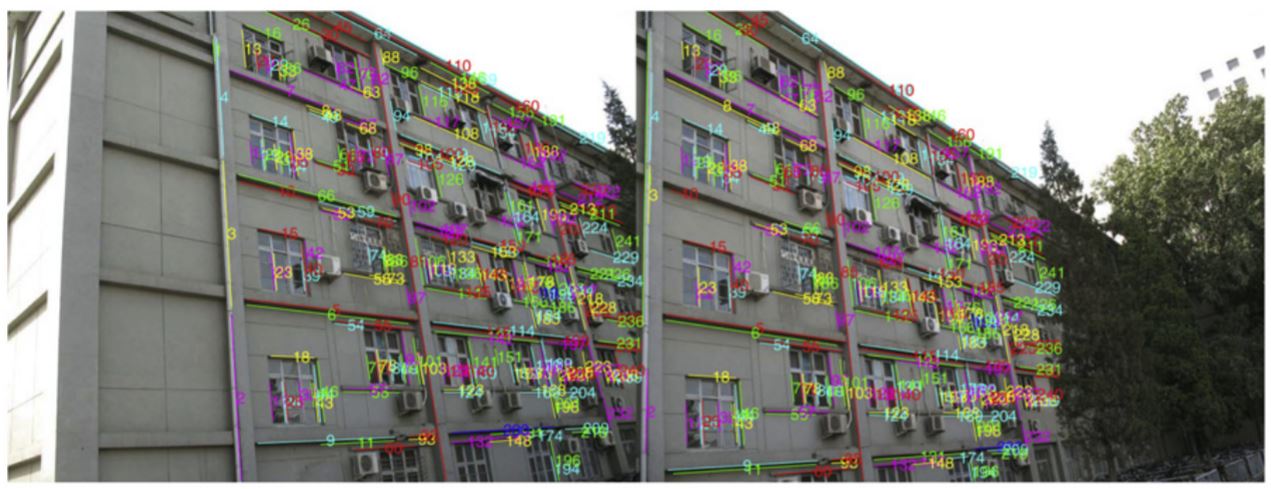}
& \includegraphics[height=0.1\textheight]{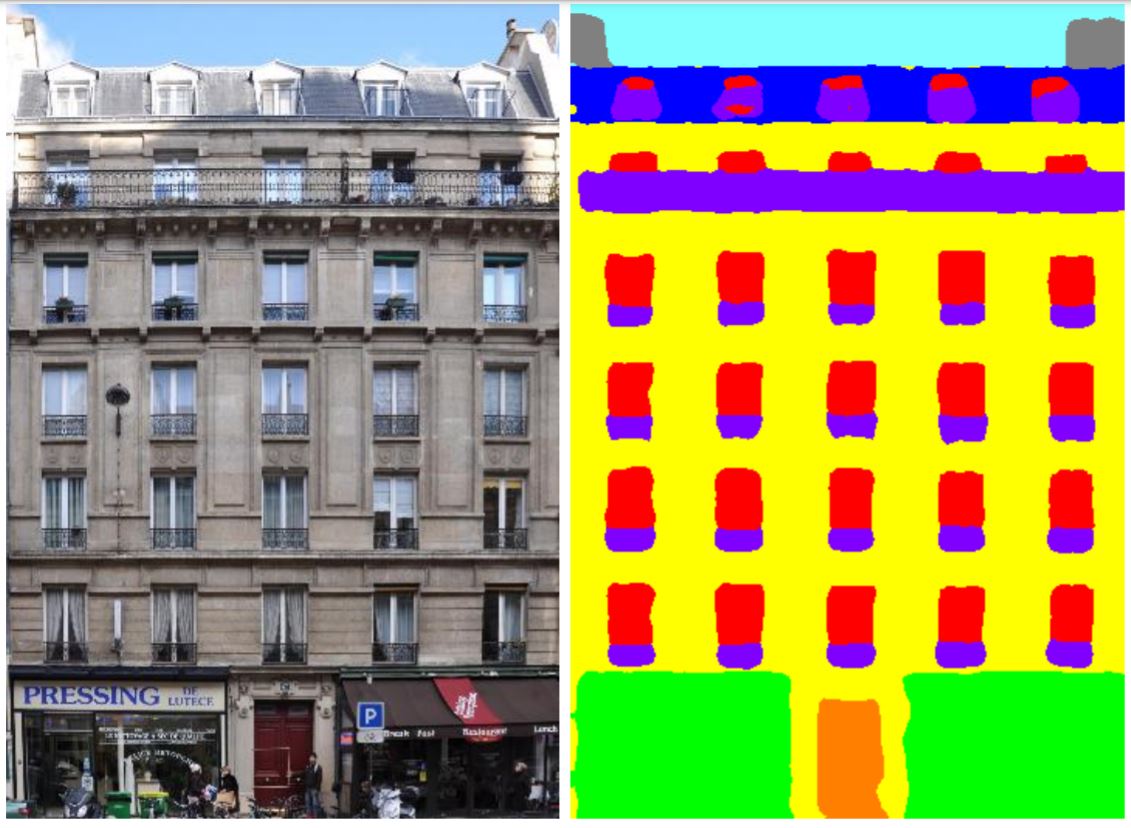}
& \includegraphics[height=0.1\textheight]{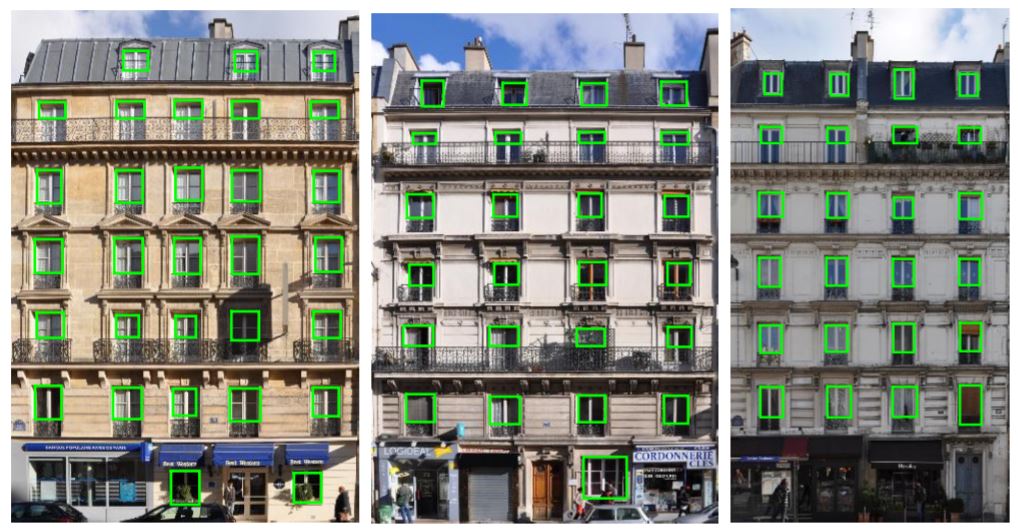}
\\
A
&
B
&
C
\end{tabular}
\caption[center]{A: Example of line matching algorithm \cite{line_match_ex2};
B: Example of semantic segmentation algorithm \cite{liu2017deepfacade};
C: Example of window detection algorithm \cite{liu2017deepfacade}}
\label{fig:algo_example}
\end{figure}

In the last decades research in this domain has increased; annually, multiple new approaches and algorithms are presented in literature regarding urban building detection. The variety of solutions used to reach the detection goal can be a combination of any of the following: edge detection algorithms \cite{2020Orhei}, \cite{edge3}, line detection \cite{line1}, \cite{line2}, line matching features \cite{line_match_ex1}, \cite{line_match_ex2}, semantic segmentation and so on. In Figure \ref{fig:algo_example} we present snapshots of steps of a building detection algorithm. All proposed algorithms bring to the table a novel approach to solve corner cases of an existing problem.

We believe that this dataset will help enhance the novel algorithms in this domain because of the gap of edges and structural details on buildings images used in other benchmarks. This need occurred when trying to develop algorithms that focus not only on boundaries or contours but on details present in the facade of buildings. For example, in Figure \ref{fig:diff_dataset_example} we present an image from the popular BSDS500 \cite{BSDS500} dataset, alongside the ground-truth for that image offered by them. In parallel we labeled the same image in our proposed concept of ground-truth. We can observe that the focus of the annotated edges is different: they focus more on boundaries and not on facade details and in our case the other way around. This difference can be an impediment to evaluate correctly the results of an algorithm depending on the scope of it: a general edge detection algorithm will not perform properly if tuned on a building-oriented dataset and of course the other way around.

The Timi\c{s}oara Building Dataset - TMBuD - \cite{CMBDT} is composed of 160 images with the resolution of 768x1024 pixels. Our motivation for this is the belief that this resolution is a good balance between the processing resources needed for manipulating the image and the actual resolution of pictures made with smart devices. Moreover, this is the actual video resolution for filming using a smartphone, the main sensor for building detection systems.

The paper is organized as following: in Section \ref{Sec:EdgeDataset} we will present popular existing edge detection datasets with ground-truth and in Section \ref{Sec:SemSegDataset} we will present similar semantic annotated datasets. In the end, in Section \ref{Sec:Proposed} we will describe our proposed dataset and the issues that we observed that resulted in the need of this new dataset.

\section{Edge detection annotated datasets}
\label{Sec:EdgeDataset}

In this section we present the existing datasets for evaluating edge detection algorithms. Even if edges do not serve as stand alone features in the new CV universe, they still represent a fundamental block for line feature detection.

\begin{figure}[h!]
\centering
\setlength{\tabcolsep}{0.5pt}
\begin{tabular}{ccccc}
& \includegraphics[height=0.1\textheight]{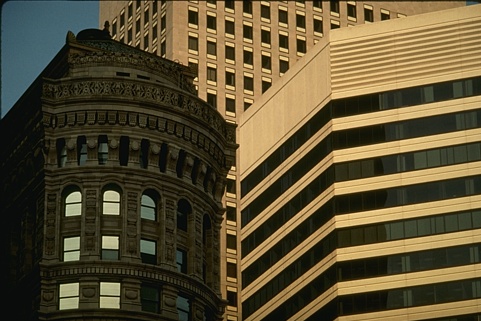}
& \includegraphics[height=0.1\textheight]{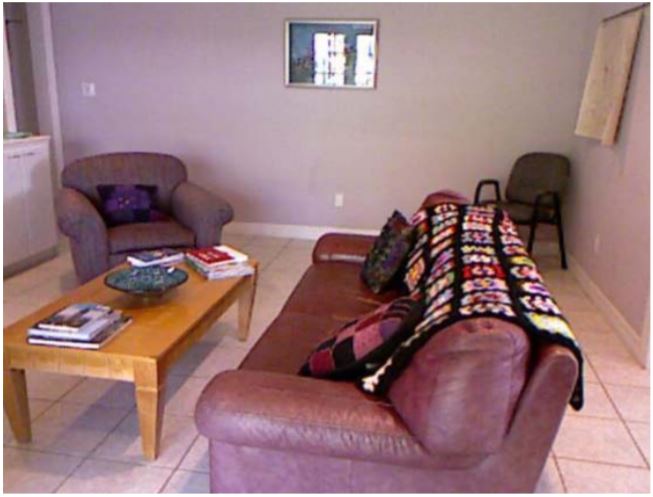}
& \includegraphics[height=0.1\textheight]{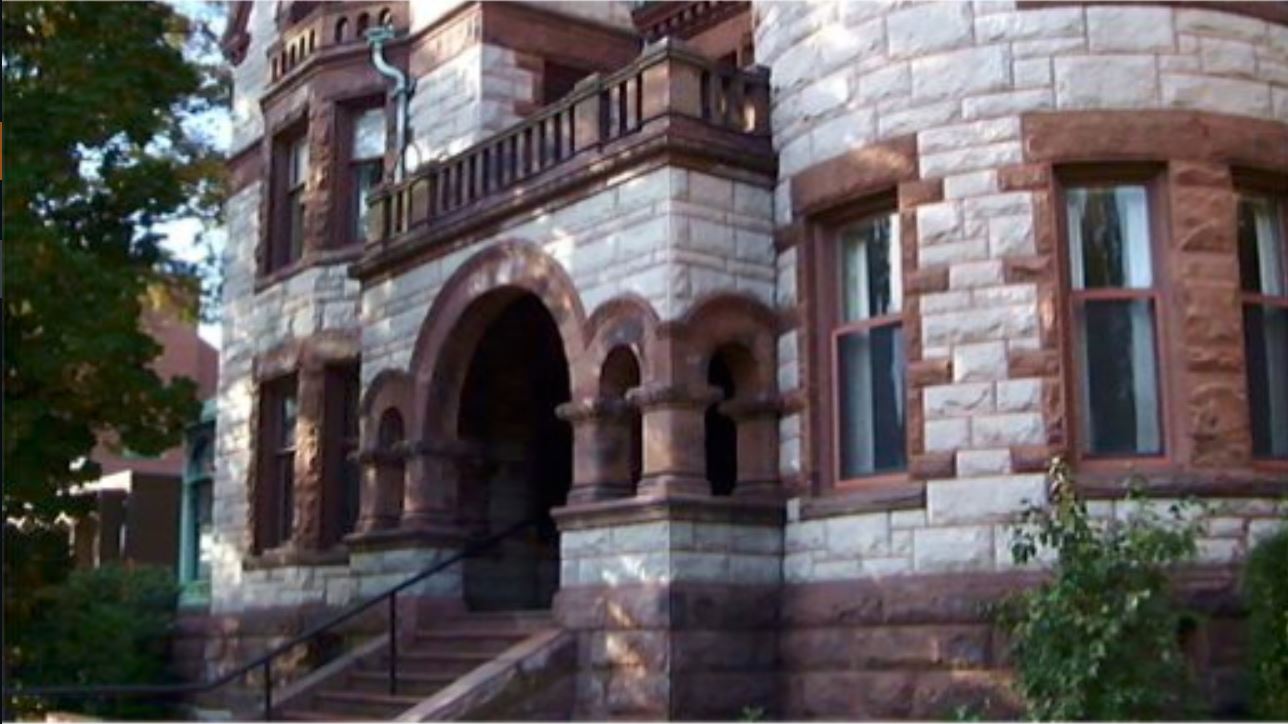}
& \includegraphics[height=0.1\textheight]{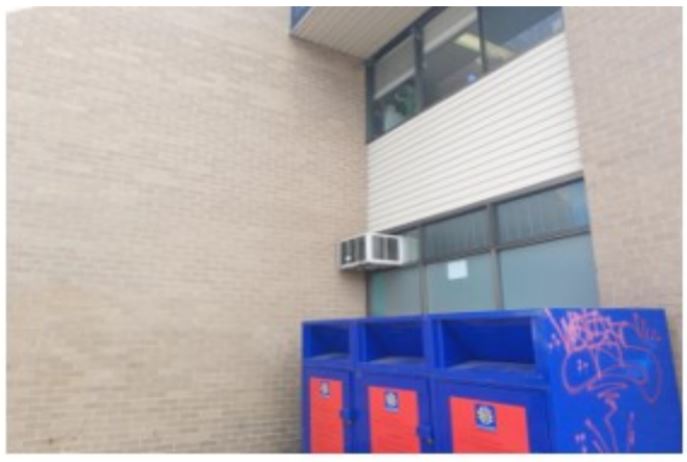}
\\
& \includegraphics[height=0.1\textheight]{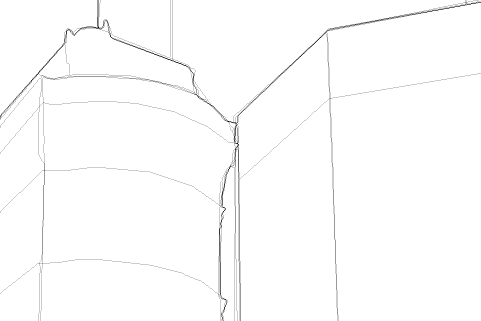}
& \includegraphics[height=0.1\textheight]{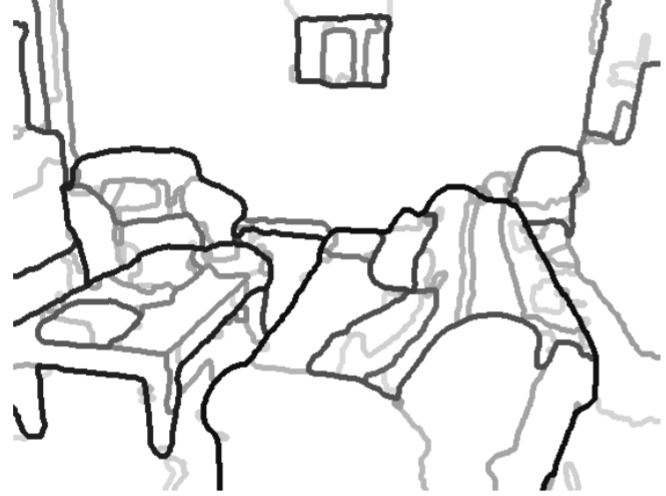}
& \includegraphics[height=0.1\textheight]{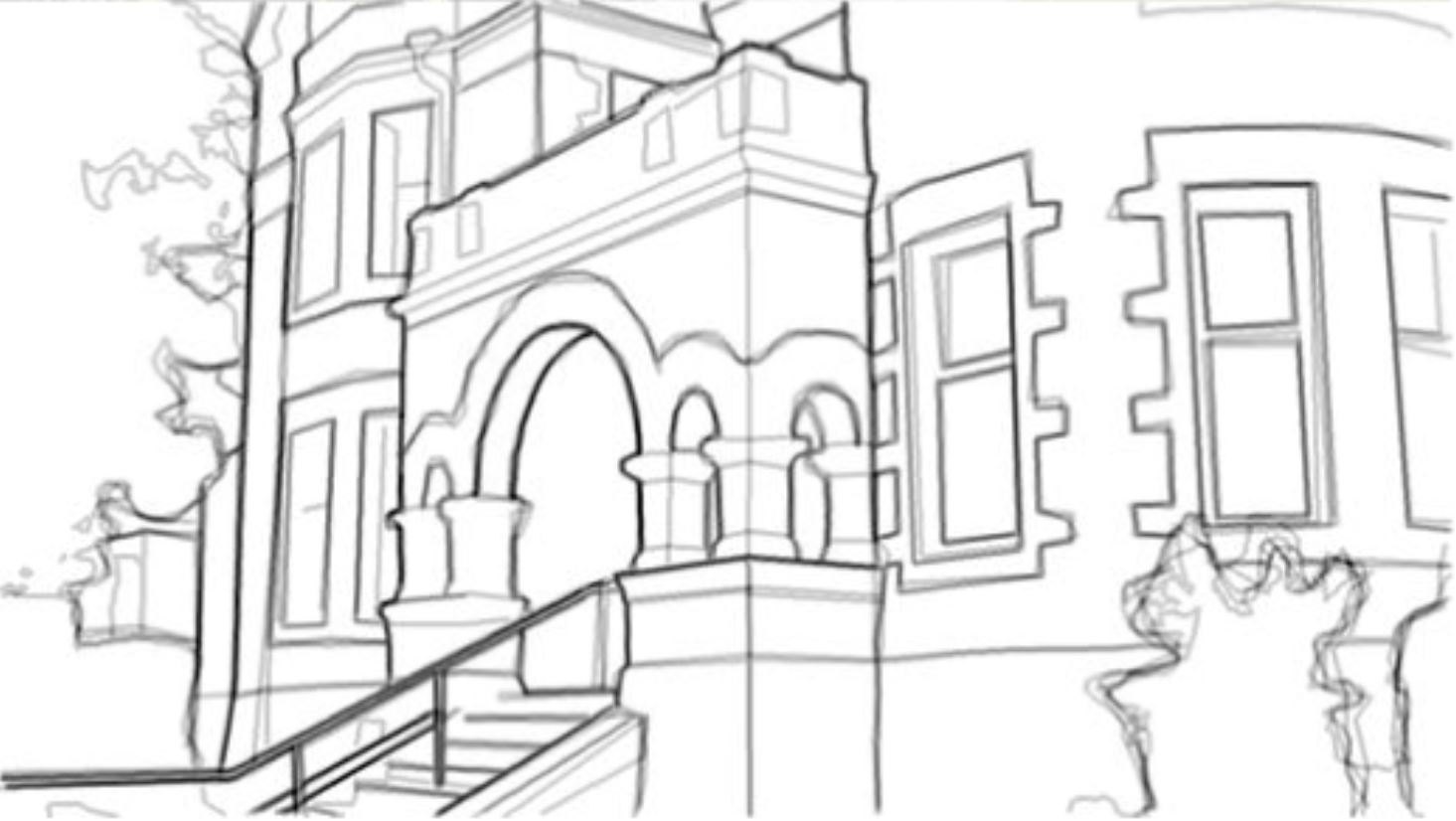}
& \includegraphics[height=0.1\textheight]{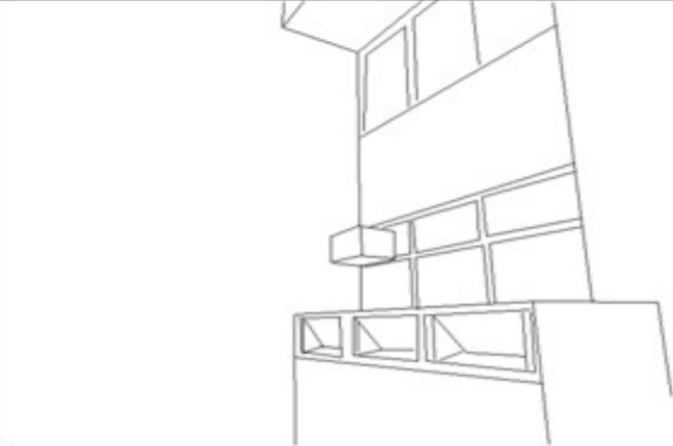}
\end{tabular}
\caption{Examples of images and equivalent ground-truth. Rows: original image, ground-truth; Coloumns: BSDS500 \cite{BSDS300}, \cite{BSDS500}, NYUDV2 \cite{nyv2}, MCUE \cite{multicue}, StructED \cite{structural}}
\label{fig:edge_example}
\end{figure}

\textbf{The Berkeley Segmentation Data Set} \cite{BSDS300}, \cite{BSDS500} is one of the most cited paper benchmarks. This benchmark is often used to compare algorithm generated contours or segmentations to human ground-truth data. For the Berkeley database, 1000 representative images of 481x321 RGB images from the Corel image database were chosen. The main criterion for selecting images was that it contains at least one distinctive object \cite{BSDS300}.

\textbf{NYU Depth Dataset V2} \cite{nyv2} consists of 1449 RGBD images comprising of commercial and residential buildings in three different cities from US. The image dataset contains 464 different indoor scenes across 26 scene classes. Each image has a dense per-pixel depth labeling using Microsoft Kinect. If a scene contained multiple instances of an object class, each instance received a unique instance label.

\textbf{The multi-cue boundary detection dataset} \cite{multicue} concerns to study the interaction of several early visual cues (luminance, color, stereo, motion) during boundary detection in challenging natural scenes. They considered a variety of places (from university campuses to street scenes and parks) and seasons to minimize possible biases. The dataset contains 100 scenes, each consisting of a left and right view short (10-frame) color sequence. Each sequence was sampled at a rate of 30 frames per second. Each frame has a resolution of 1280 by 720 pixels.

\textbf{The Structural Edge dataset} \cite{structural} propose a new concept of structural edge. Structural edges include occluding contours of objects as well as orientation discontinuities in surfaces are important for understanding the 3D structure of objects and environments. The validity of structural edges was tested using an eye tracking test. The structural edge dataset contains 600 images in natural indoor and outdoor scenes. The structural edges are labeled manually and validated by eye-tracking data from 10 participants with overall 20 trials.

In Figure \ref{fig:edge_example} we present images with the ground-truth from the dataset presented in this section. The mentioned datasets don't focus on certain domain of images of future specific scope to be used. This is a positive point when concerning with a wide range scope algorithm evaluation but is a negative aspect when focusing on a single use case, as we concern ourselves. 

\section{Semantic Segmentation annotated datasets}
\label{Sec:SemSegDataset}

In this section we will present existing semantic segmentation datasets that focus on urban scenarios and that would be a good candidate to be used in constructing a building detection algorithm in the end. 

The datasets which are selected and used by system designers play a very important role in the quality of the trained model and thereby system performance. So, selecting an appropriate dataset for a task can be one of the most challenging steps at the beginning of the research process \cite{Zlateski2018OnTI}. 

\begin{figure}[h!]
\setlength{\tabcolsep}{2pt}
\centering
\begin{tabular}{ccccccc} 
		\includegraphics[height=2.5cm,width=1.6cm]{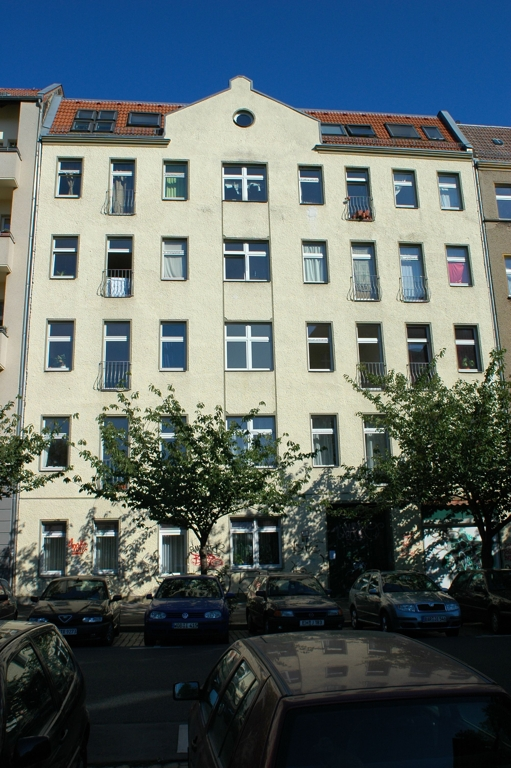}
	&
		\includegraphics[height=2.5cm,width=1.6cm]{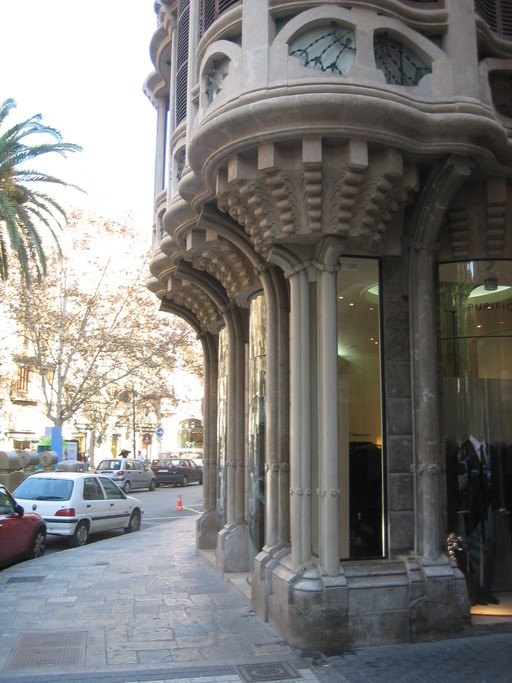}
	&
	    \includegraphics[height=2.5cm,width=1.6cm]{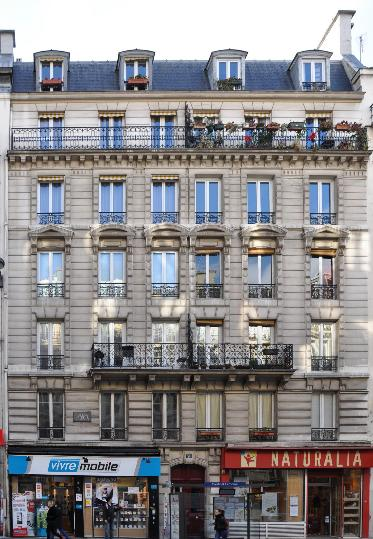}
	&
		\includegraphics[height=2.5cm,width=1.6cm]{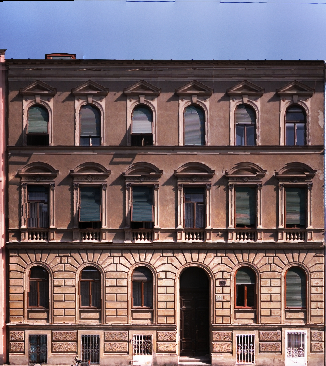}
	&
		\includegraphics[height=2.5cm,width=1.6cm]{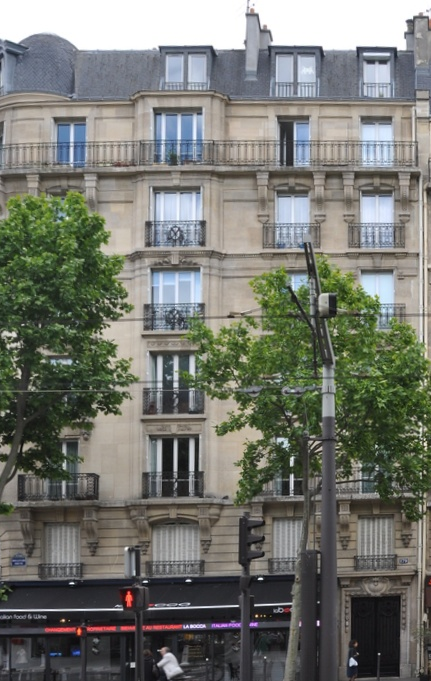}
	&
		\includegraphics[height=2.5cm,width=1.6cm]{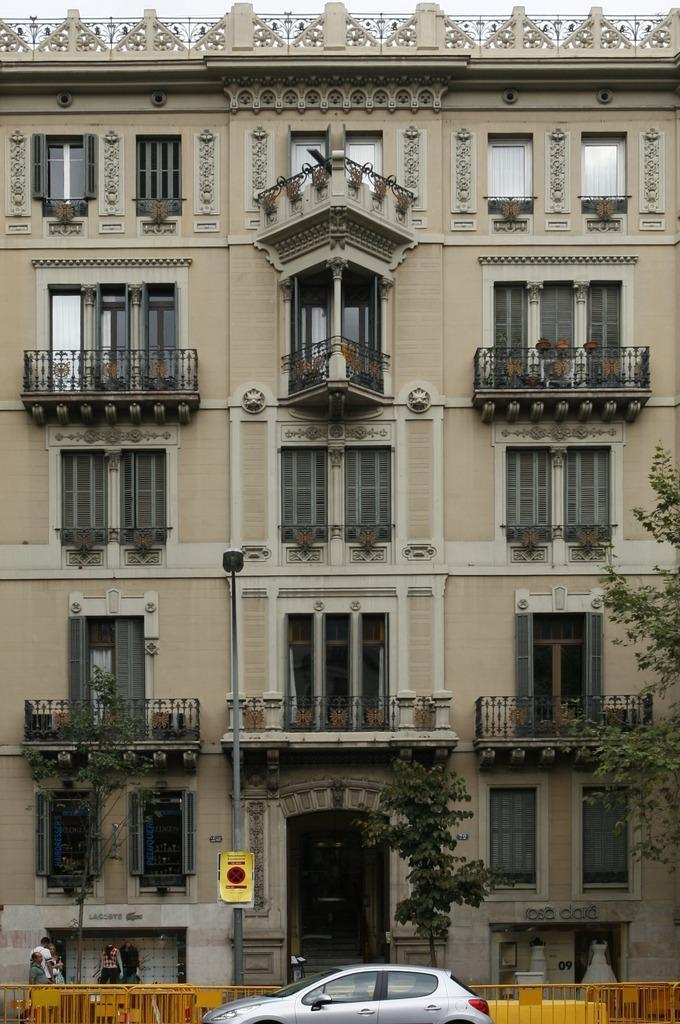}
	&
		\includegraphics[height=2.5cm,width=1.6cm]{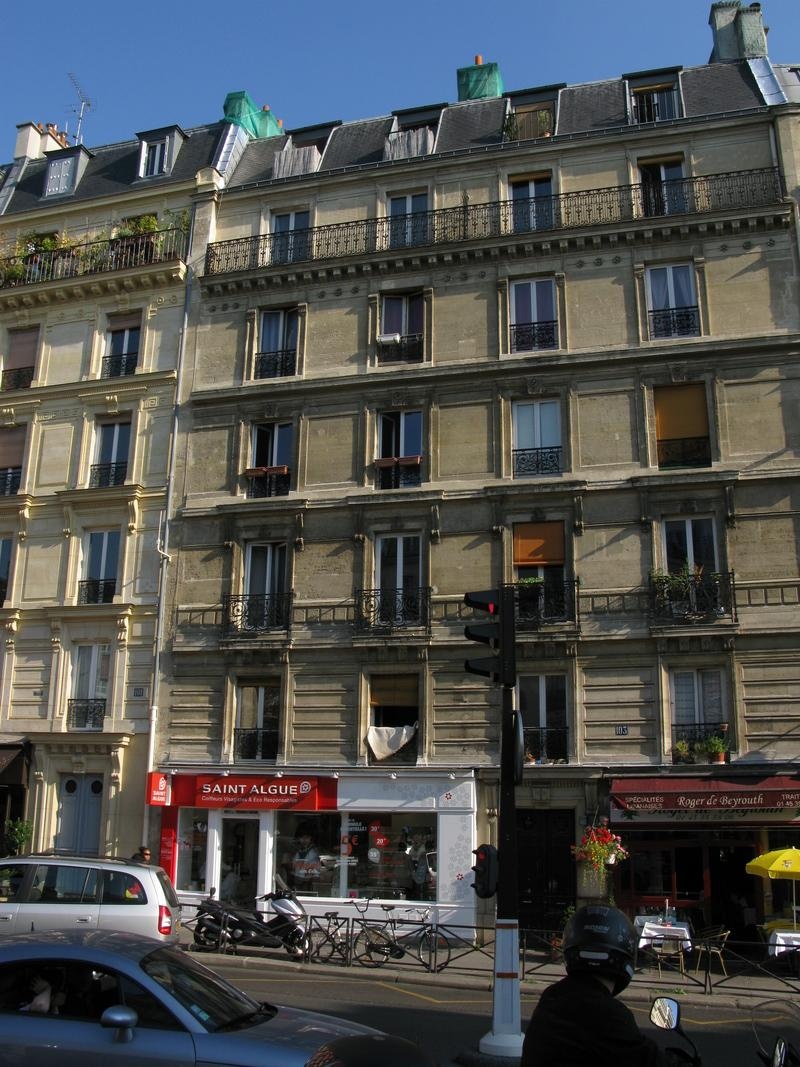}
\\
        \includegraphics[height=2.5cm,width=1.6cm]{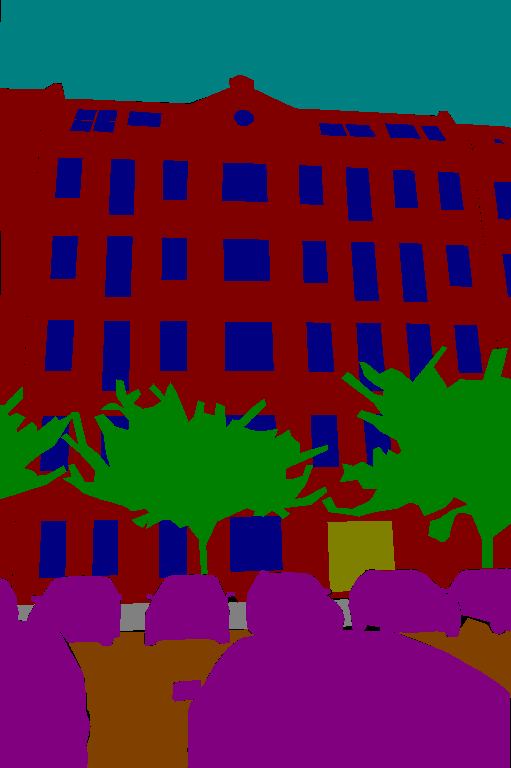}
	&
		\includegraphics[height=2.5cm,width=1.6cm]{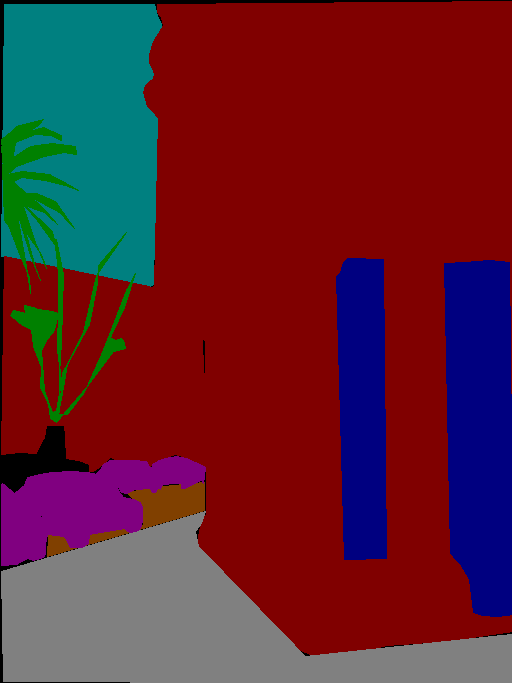}
	&
	    \includegraphics[height=2.5cm,width=1.6cm]{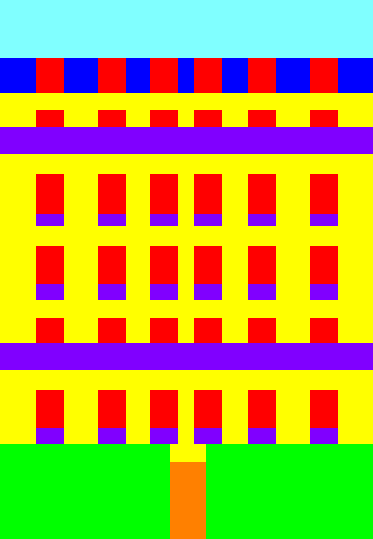}
	&
		\includegraphics[height=2.5cm,width=1.6cm]{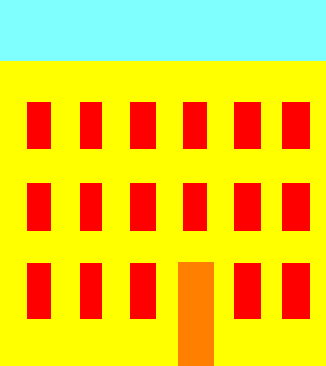}
	&
		\includegraphics[height=2.5cm,width=1.6cm]{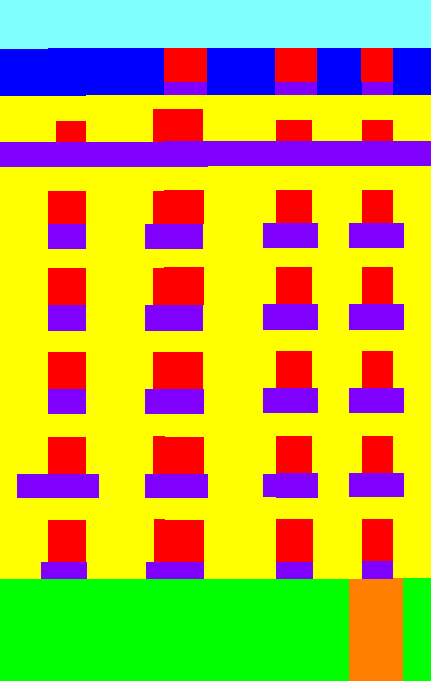}
	&
		\includegraphics[height=2.5cm,width=1.6cm]{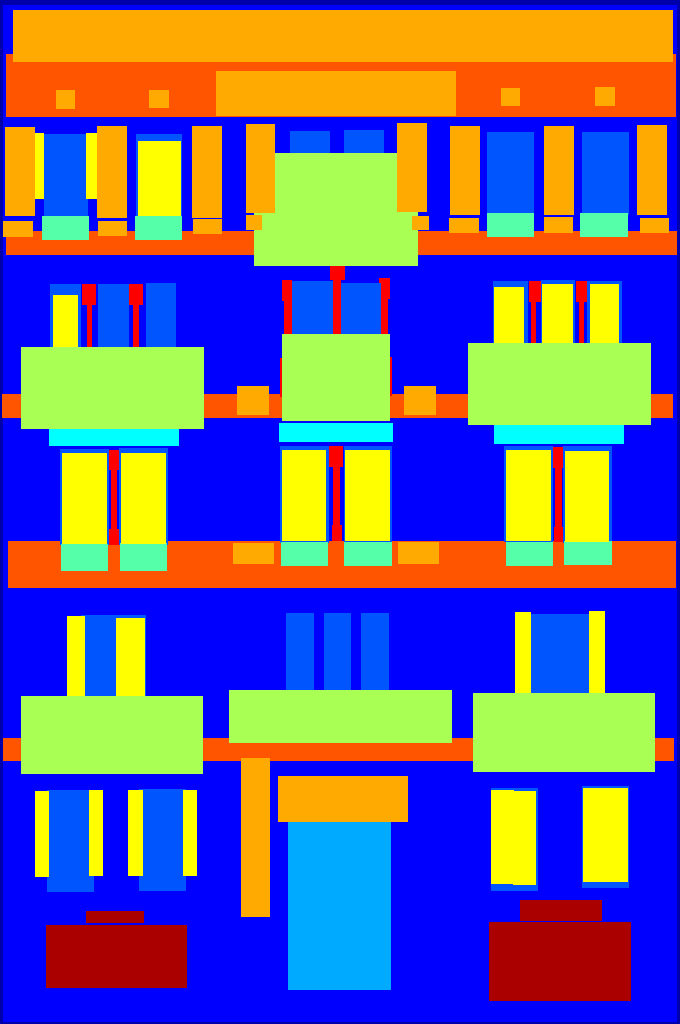}
	&
		\includegraphics[height=2.5cm,width=1.6cm]{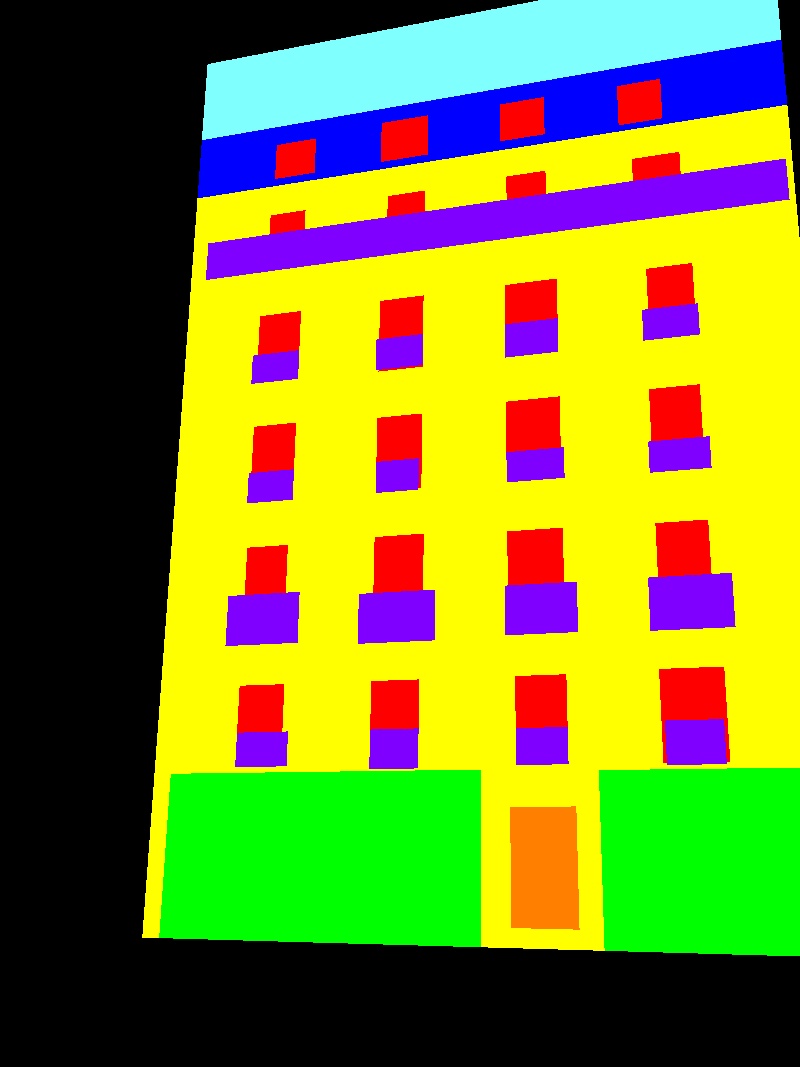}

\\
	    \includegraphics[height=2.5cm,width=1.6cm]{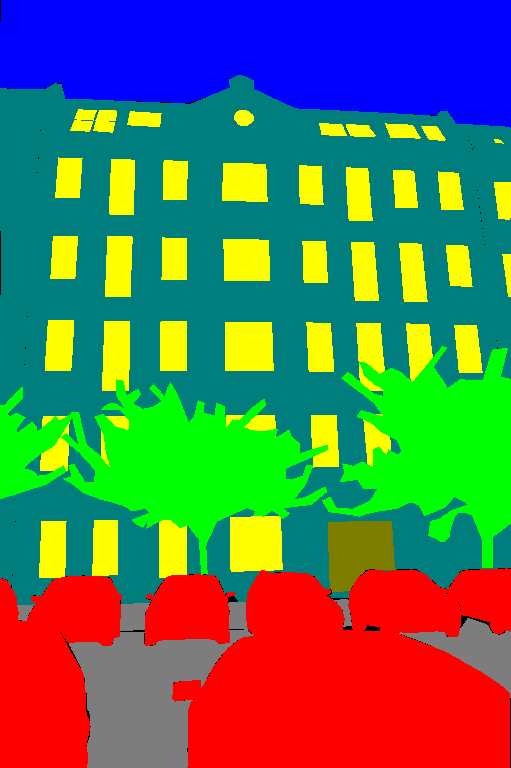}
	&
		\includegraphics[height=2.5cm,width=1.6cm]{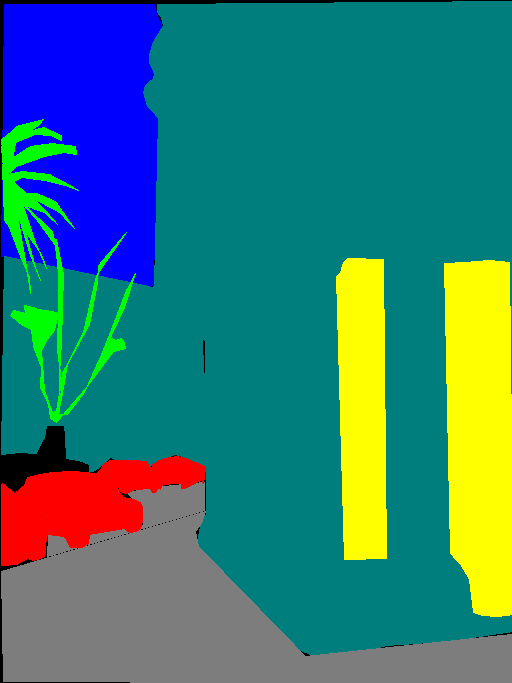}
	&
        \includegraphics[height=2.5cm,width=1.6cm]{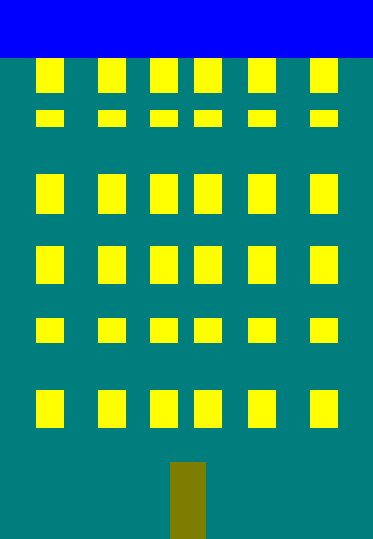}
	&
		\includegraphics[height=2.5cm,width=1.6cm]{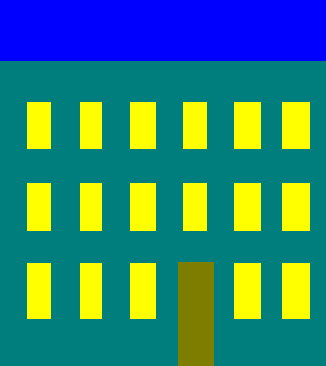}
	&
		\includegraphics[height=2.5cm,width=1.6cm]{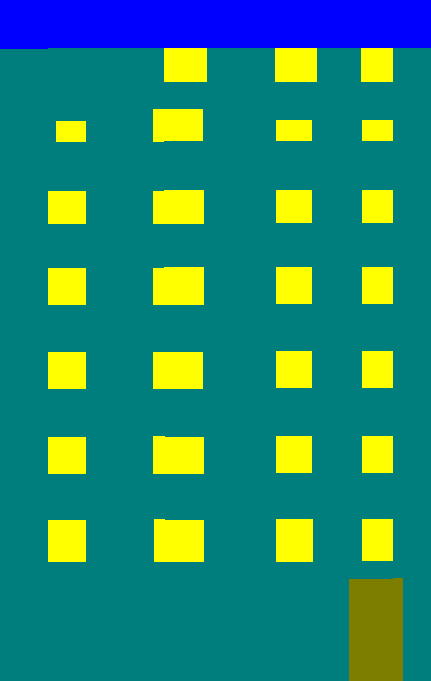}
	&
		\includegraphics[height=2.5cm,width=1.6cm]{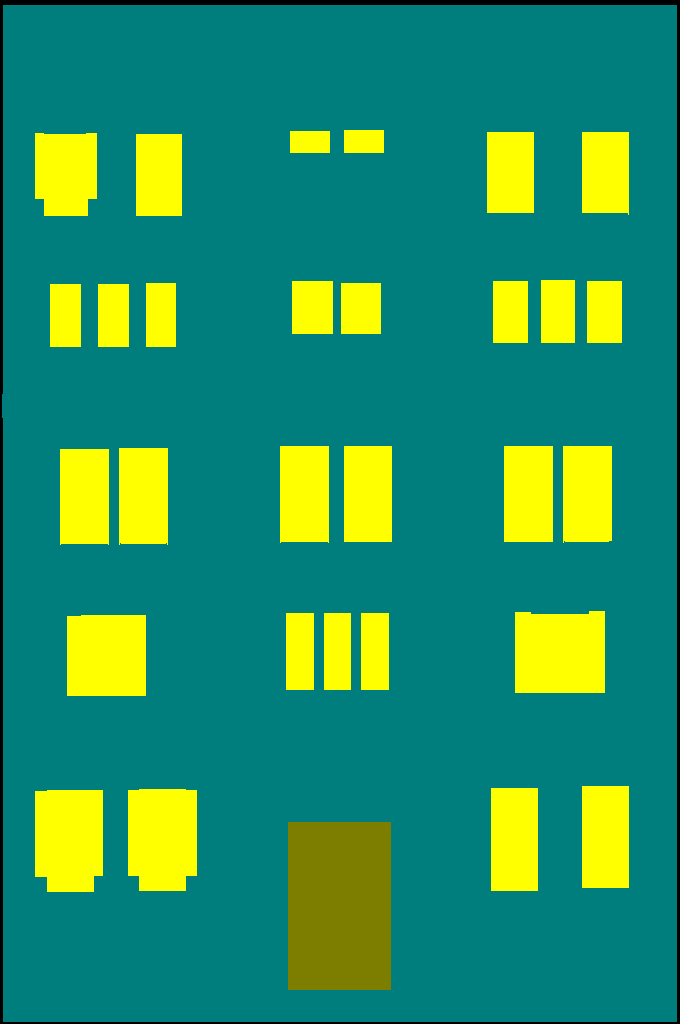}
	&
		\includegraphics[height=2.5cm,width=1.6cm]{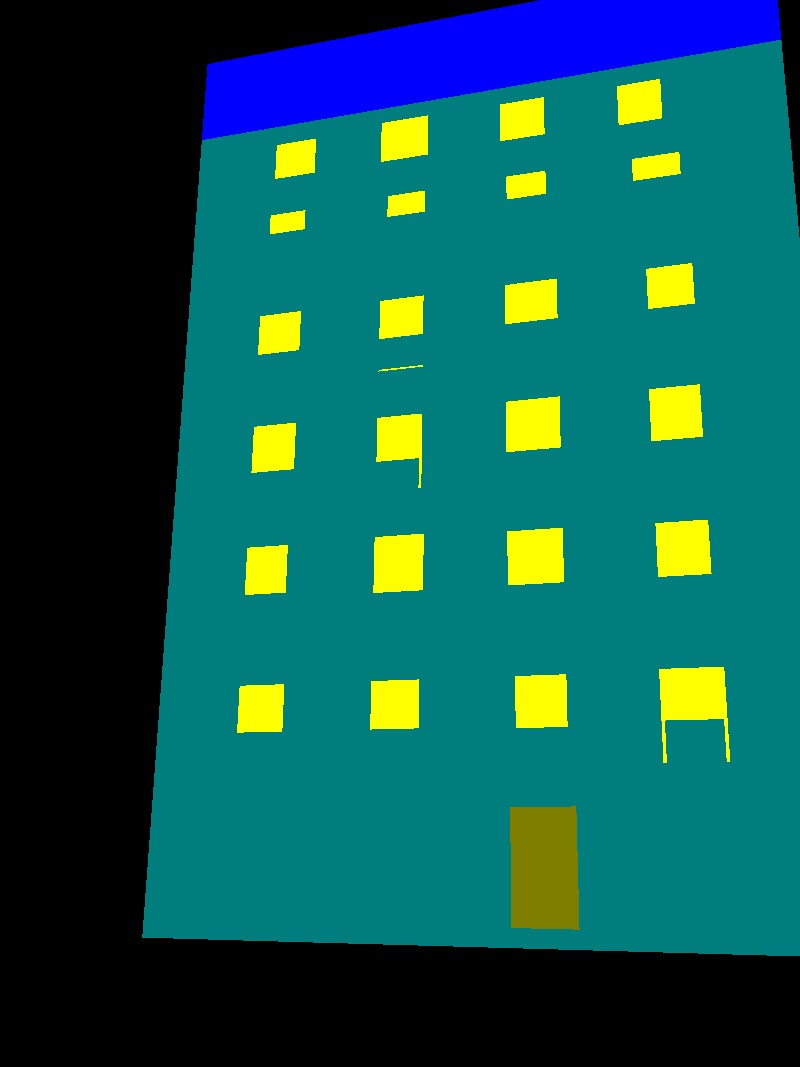}

\end{tabular}
\caption{Data sets class correlations. Rows: Original image, Original labels, transition to TMBuD labels; Columns: eTRIMS, LabelMeFacade, ECP, ICG Graz5, INRIA, CPM, VarCity}
\label{fig:dataset_labels}
\end{figure}%

\textbf{eTRIMS Image Database} \cite{eTRIMS} is comprised of 60 annotated images and offers two distinct labels: the 4-Class eTRIMS Dataset with 4 annotated object classes and the 8-Class eTRIMS Dataset with 8 annotated object classes. In the 8–class dataset are the following eight object classes: sky, building, window, door, vegetation, car, road, pavement.

\textbf{LabelMeFacade Database} \cite{LabelMe1} \cite{LabelMe2} contains 945  images with labeled polygons that describe the different classes. The classes provided are: buildings, windows, sky, and a limited number of unlabeled regions (maximum 20\% of the image). The pixelwise labeled images are created by utilizing the eTRIMS categories and a simple depth order heuristic.

\textbf{Ecole Centrale Paris Facades Database} \cite{ECP_dataset} \cite{ECP_dataset_2} contains 109 images of Paris facades with annotations that have been manually rectified. Classes used for annotation are: window, wall, door, roof, sky, shop.

\textbf{ICG Graz50 Facade Database} \cite{ICGGraz50} is a dataset of rectified facade images and semantic labels that was created with the goal of studying facades. It is comprised of 50 images of various architectural styles (Classicism, Biedermeier, Historicism, Modern and so on).

\textbf{The Paris Art Deco Facades dataset} \cite{ParisArtDecoFacadesDataset} consists of 80 images of rectified facades of the Art Deco style. The dataset offers 79 RGB images with 6 annotated labels. Occlusions of the facade are ignored but the occlusion reasoning is offered by the dataset.

\textbf{The CMP Facade Database} \cite{CPM} consists of facade images assembled at the Center for Machine Perception. The dataset includes 606 rectified images of facades from various cities of the world, which have been manually annotated. Annotation is defined as a set of rectangles scope with assigned class labels that can overlap if needed.

\textbf{VarCity 3D Dataset} \cite{VarCity} \cite{VarCity2} consists of 700 images along a street annotated with pixel-level labels for facade details. Classes provided are: windows, doors, balconies, roof, etc. The dataset provides images, labels and indexes to the 3D surface together with evaluation source code for comparing different tasks.

In Figure \ref{fig:dataset_labels} we can observe examples of images and the associated labels that are offered in the dataset presented in this section. As we can see the perspective of each dataset is different. ECP, ICG Graz5, CMP focus more on facade details offering several classes to better understand the facade features. In the VarCity dataset we see that the focus is on the main building discarding the rest of the buildings in the image.

\section{Our proposed TMBuD dataset}
\label{Sec:Proposed}

We intend with our dataset, TMBuD, to unify several ground truth evaluation in the same framework. Of course doing so in a global fashion is close to impossible so we wish to limit the use case to detection, feature extraction and localization of buildings in urban scenarios, based on image understanding. 

Building detection is the process of obtaining the approximate position and shape of a building, while building extraction can be defined as the problem of precisely determining the building outlines, which is one of the critical problems in digital photogrammetry \cite{elshehaby2009new}.

\begin{figure}[h!]
\setlength{\tabcolsep}{2pt}
\centering
\begin{tabular}{cccccccc} 
		\includegraphics[height=0.13\textheight]{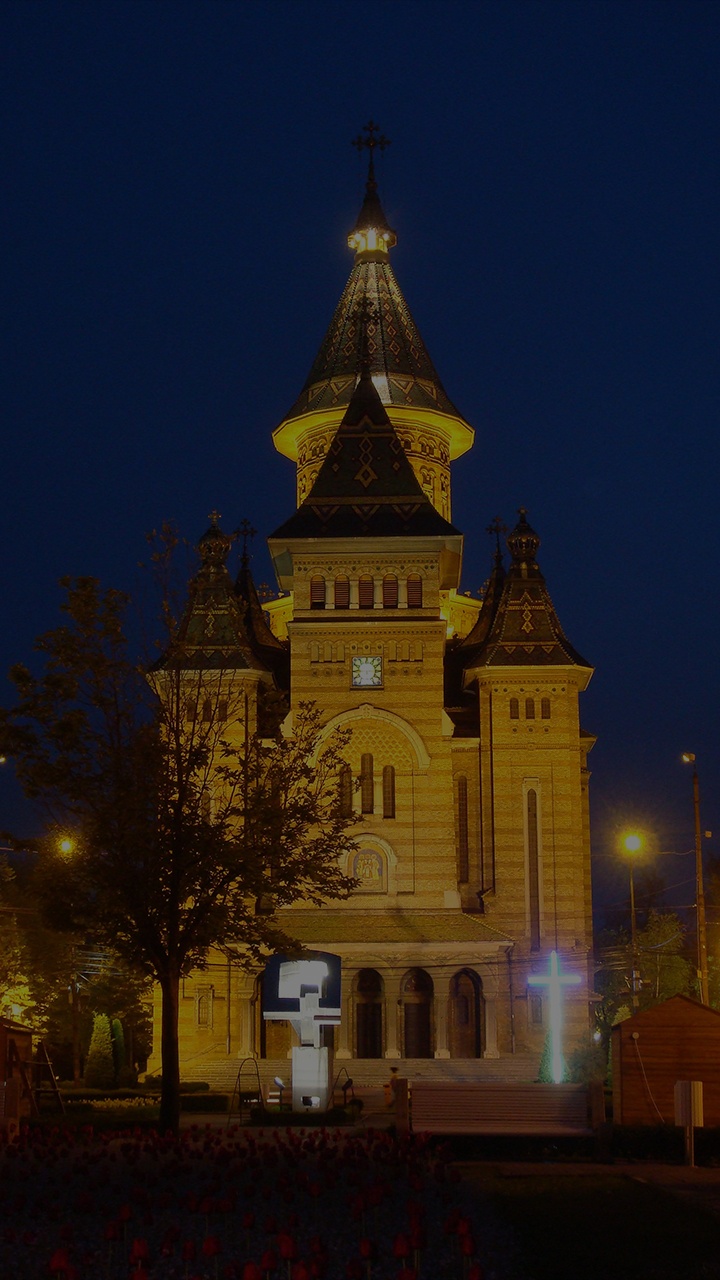}
	&
		\includegraphics[height=0.13\textheight]{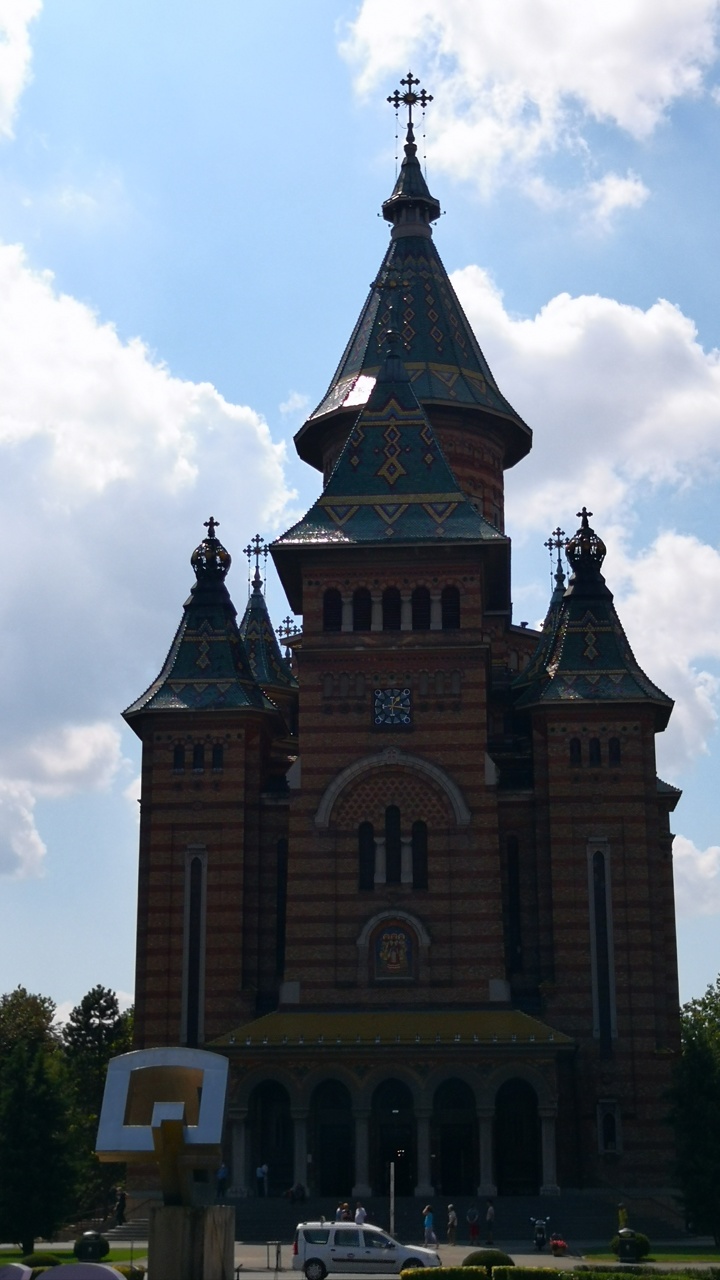}
	&
	    \includegraphics[height=0.13\textheight]{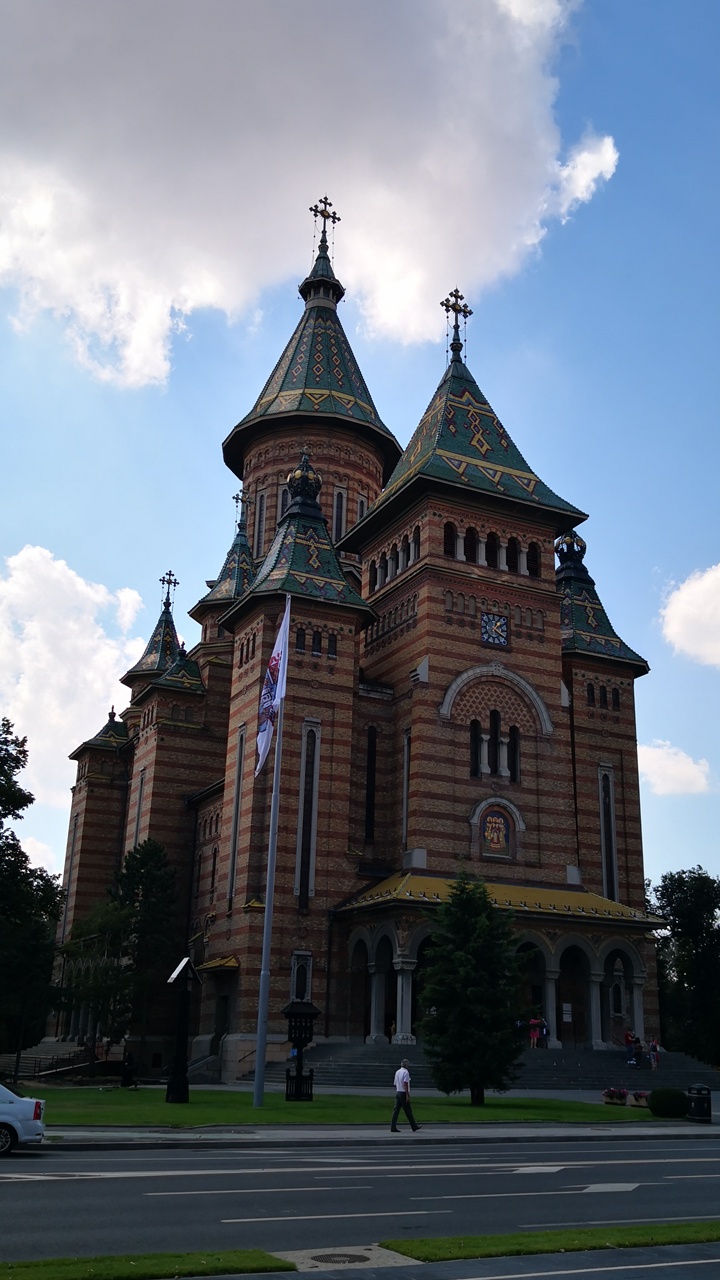}
	&
		\includegraphics[height=0.13\textheight]{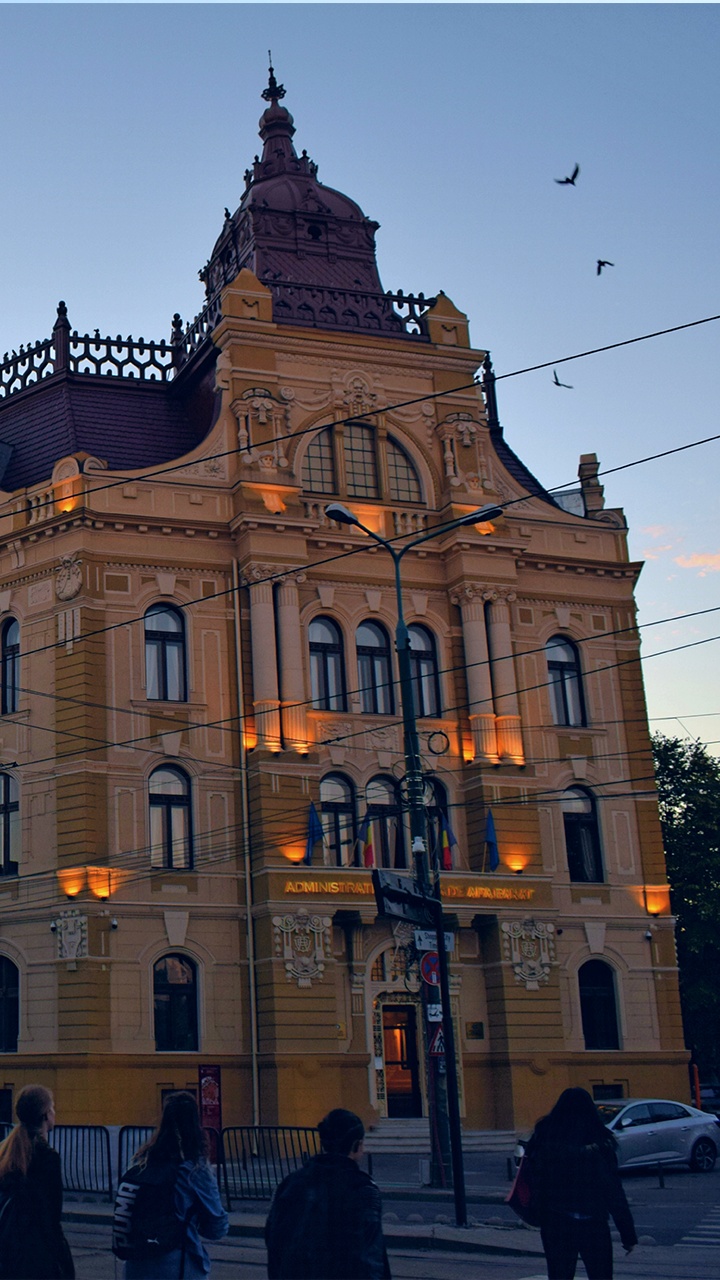}
	&
		\includegraphics[height=0.13\textheight]{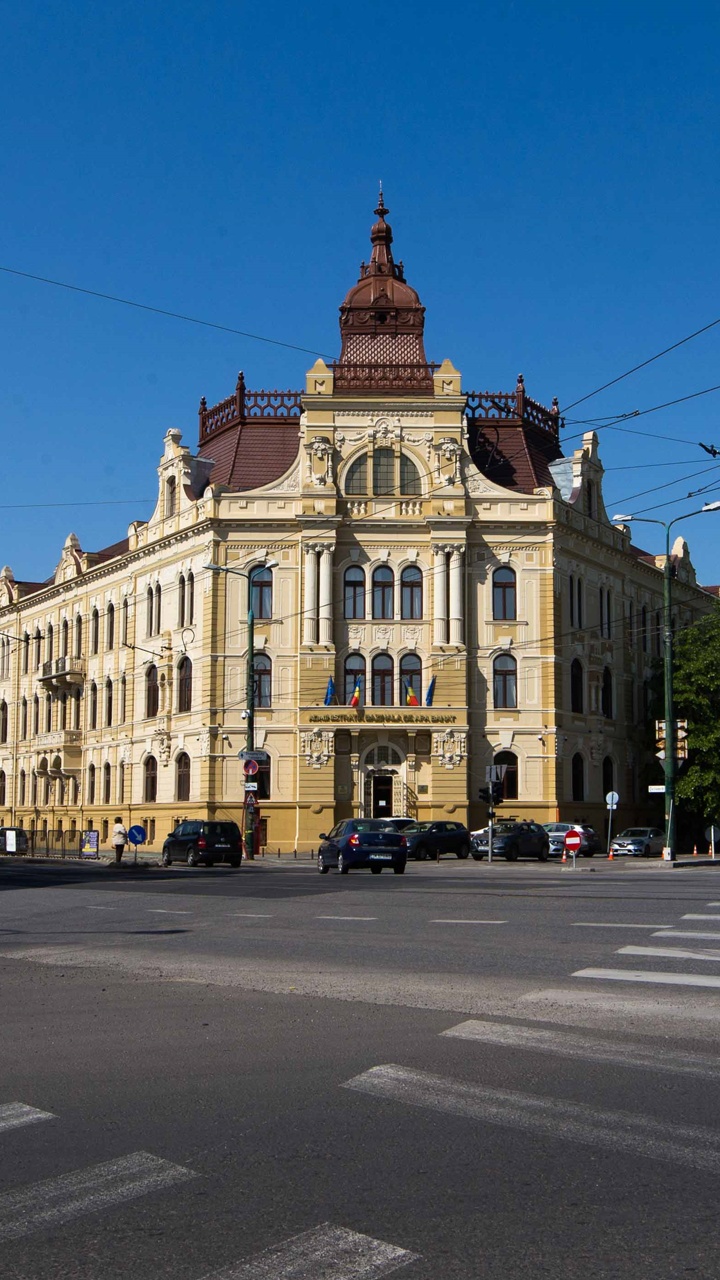}
	&
		\includegraphics[height=0.13\textheight]{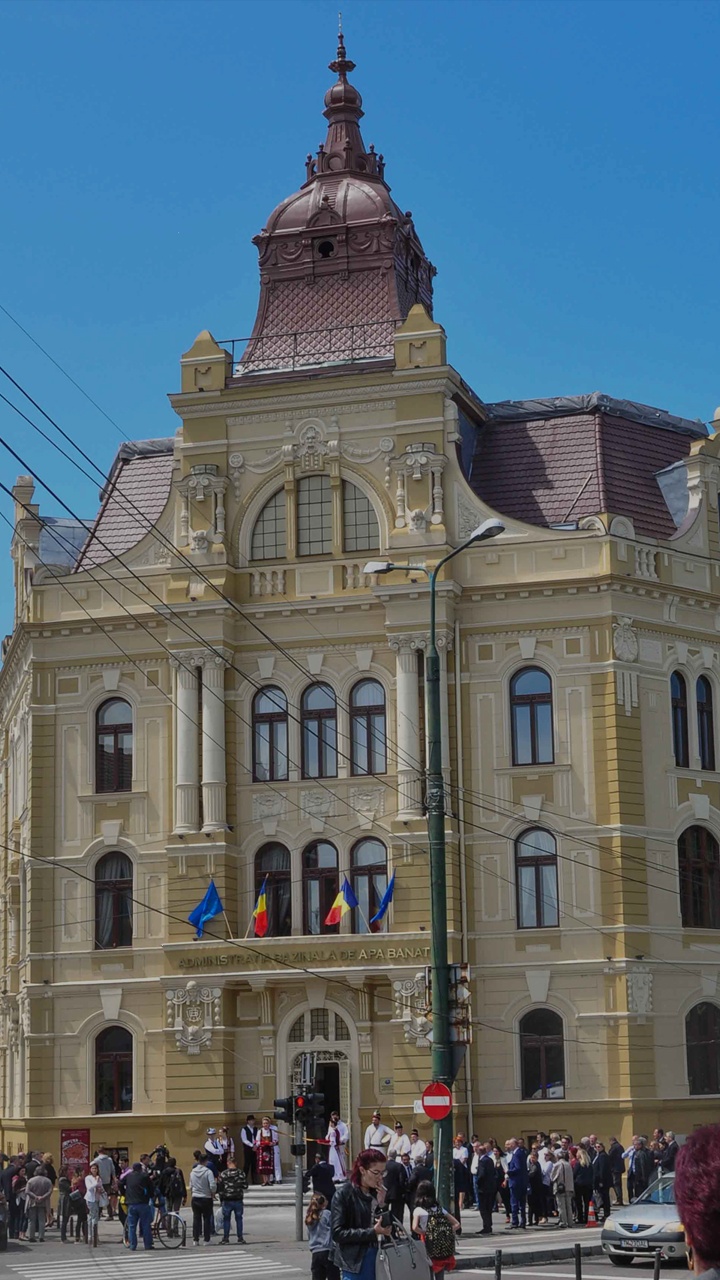}
	&
		\includegraphics[height=0.13\textheight]{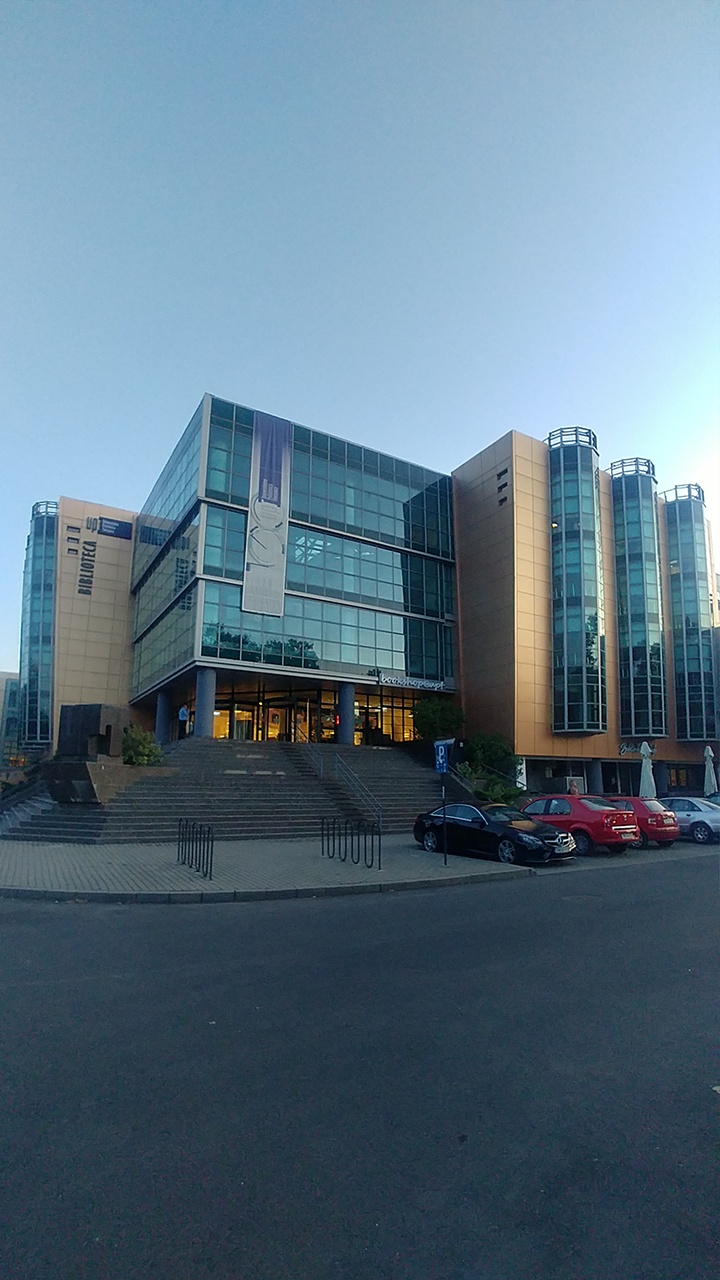}
	&
		\includegraphics[height=0.13\textheight]{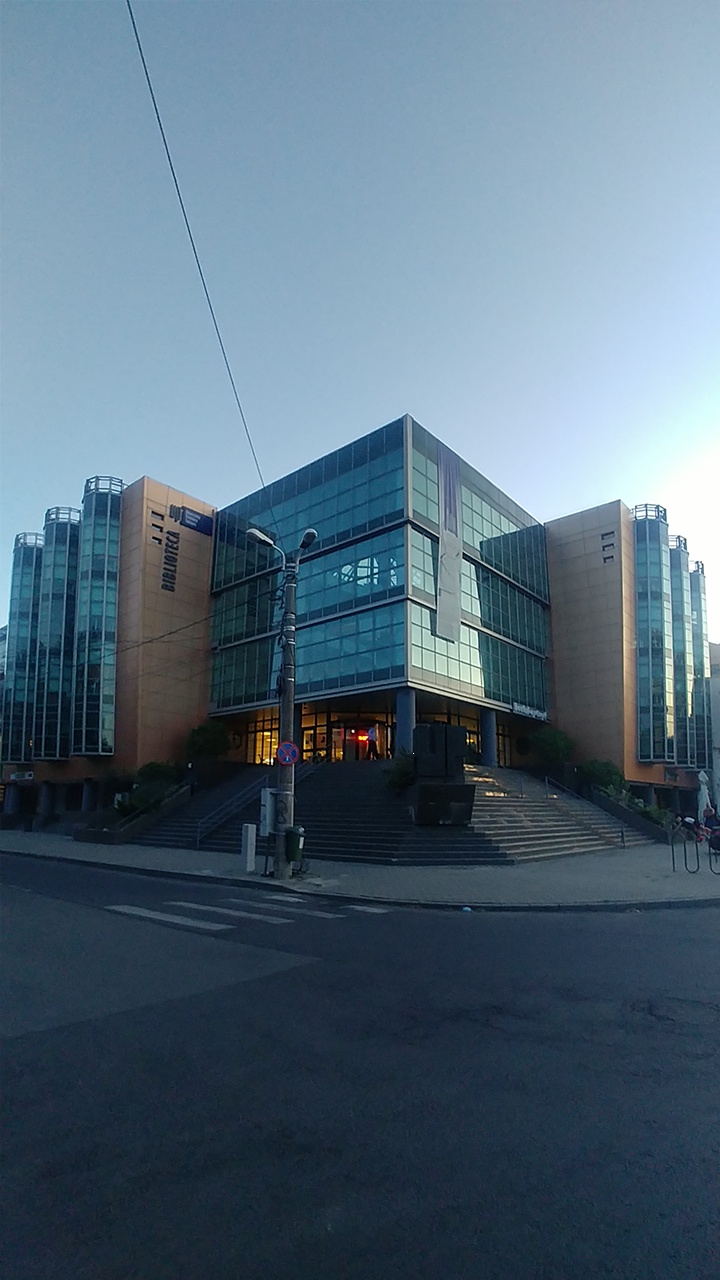}
\\
		\includegraphics[height=0.13\textheight]{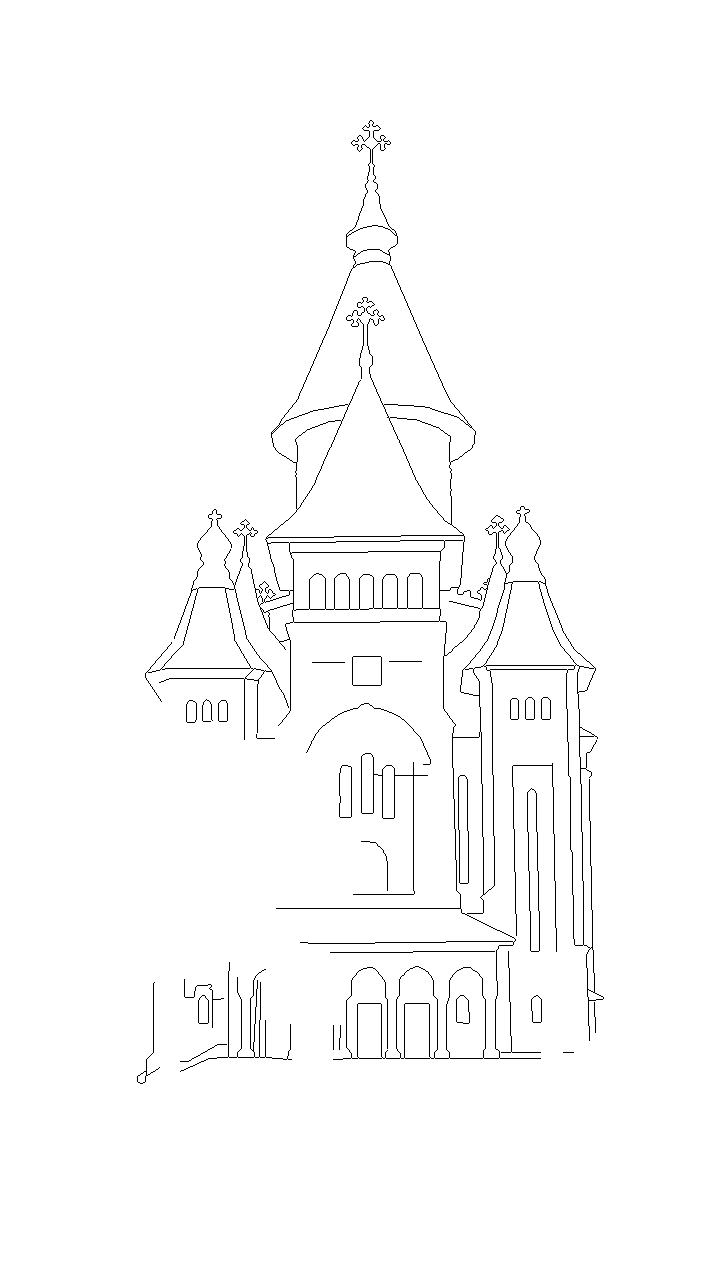}
	&
		\includegraphics[height=0.13\textheight]{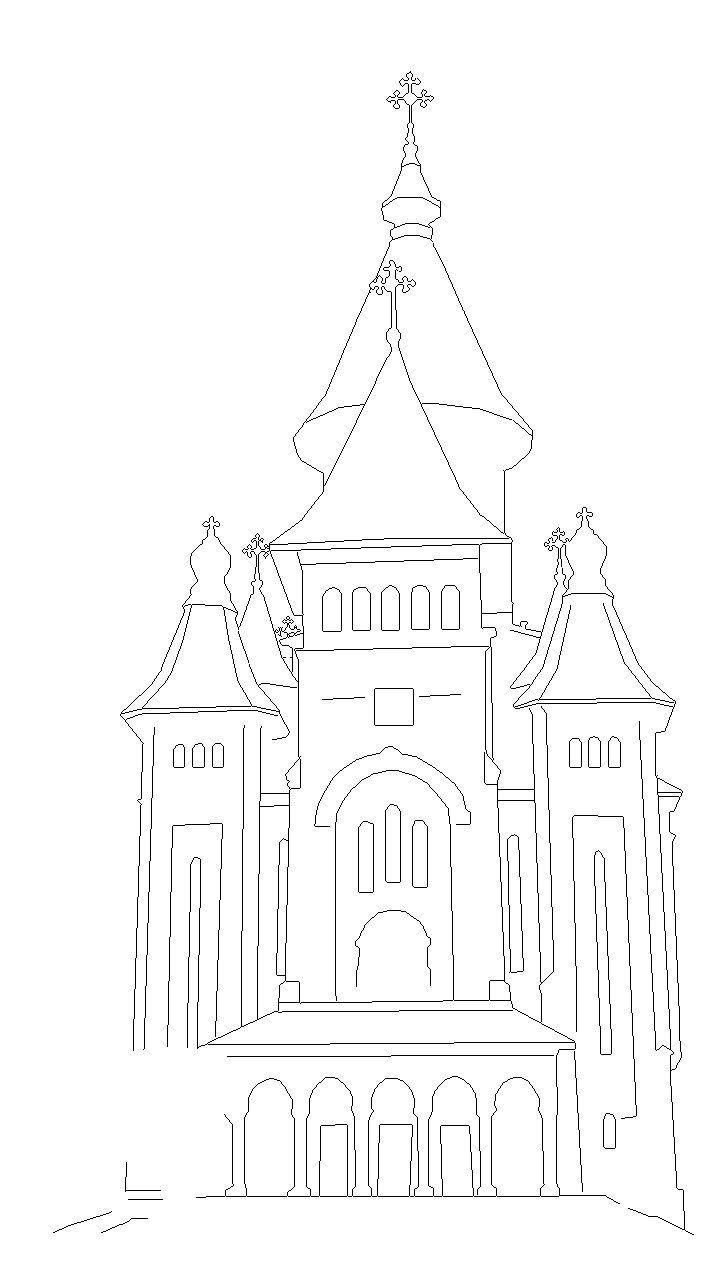}
	&
	    \includegraphics[height=0.13\textheight]{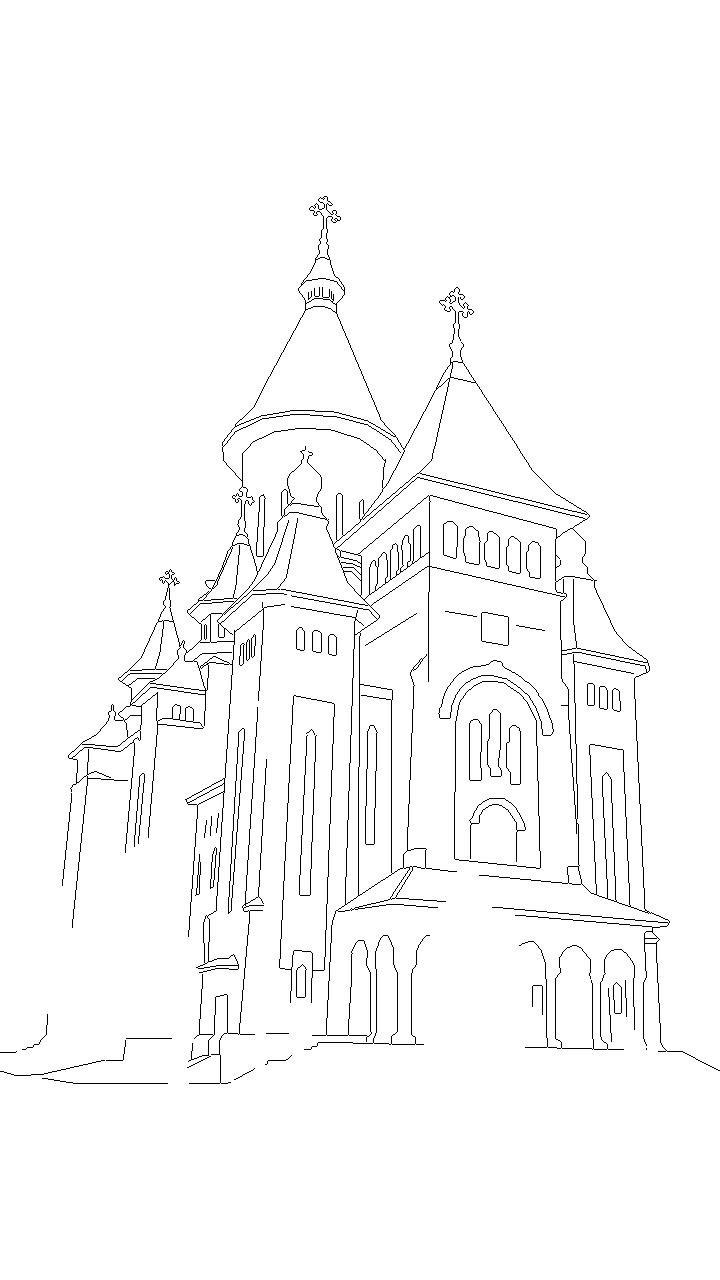}
	&
		\includegraphics[height=0.13\textheight]{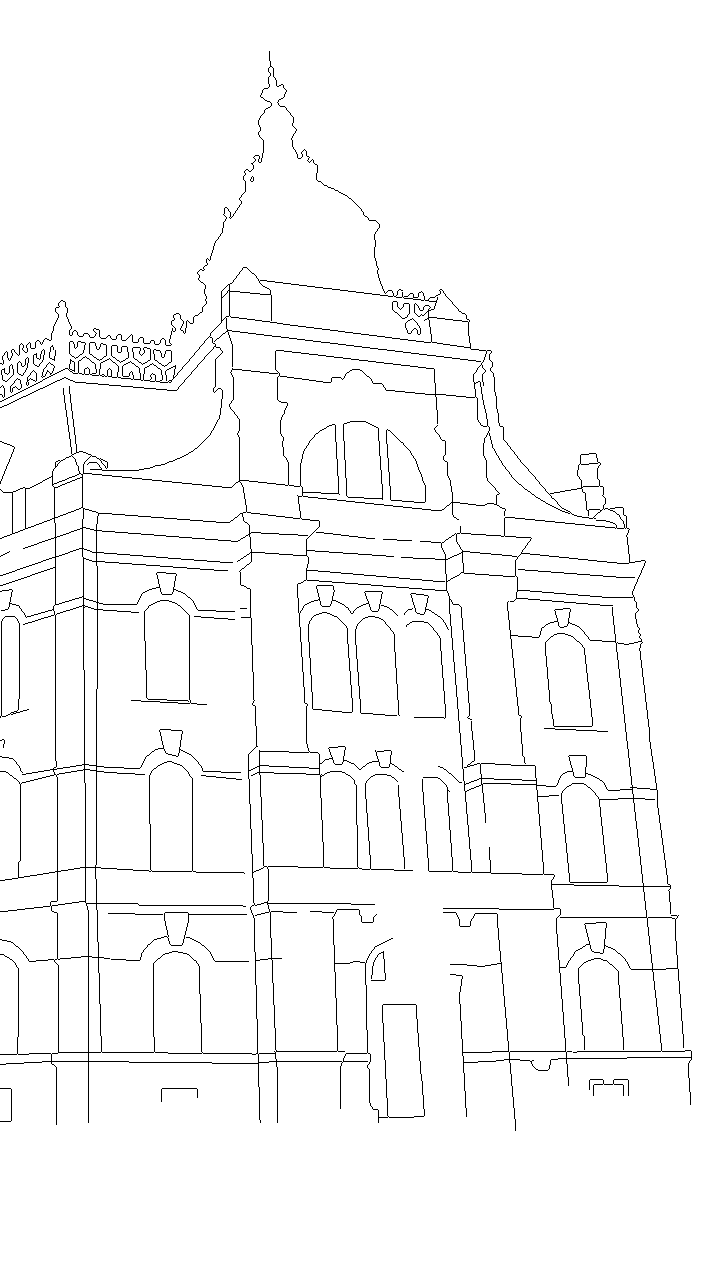}
	&
		\includegraphics[height=0.13\textheight]{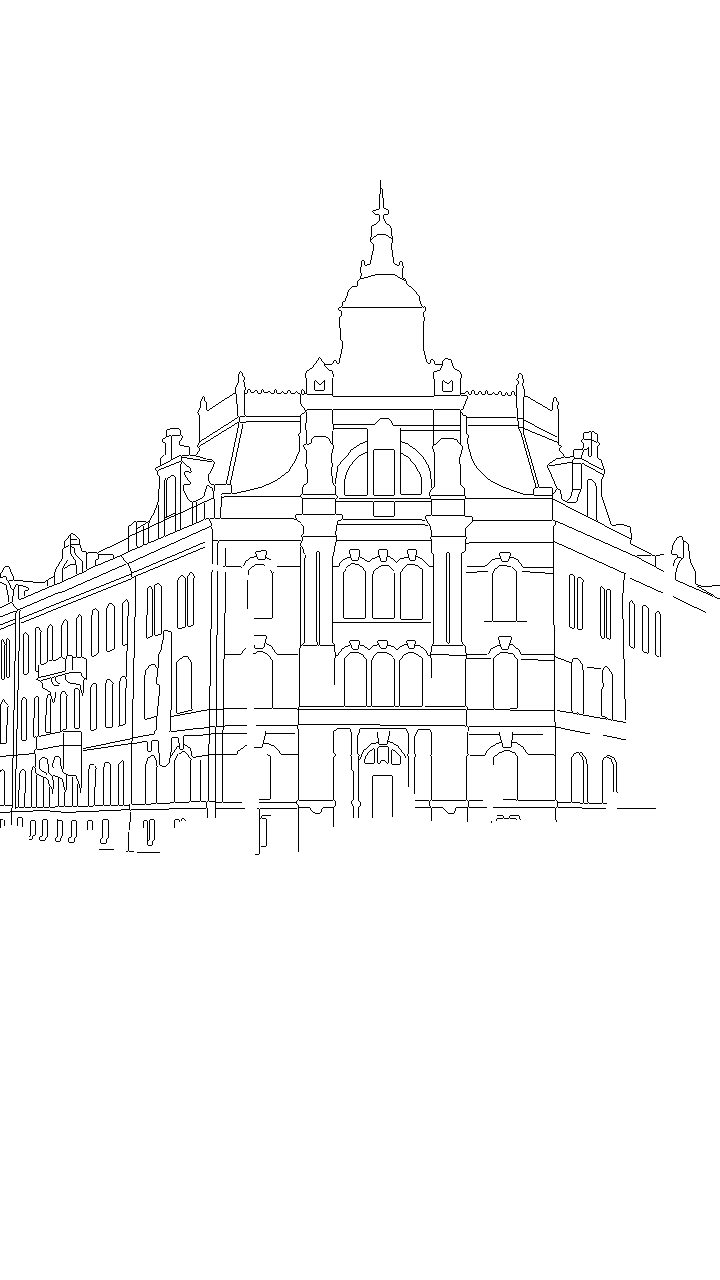}
	&
		\includegraphics[height=0.13\textheight]{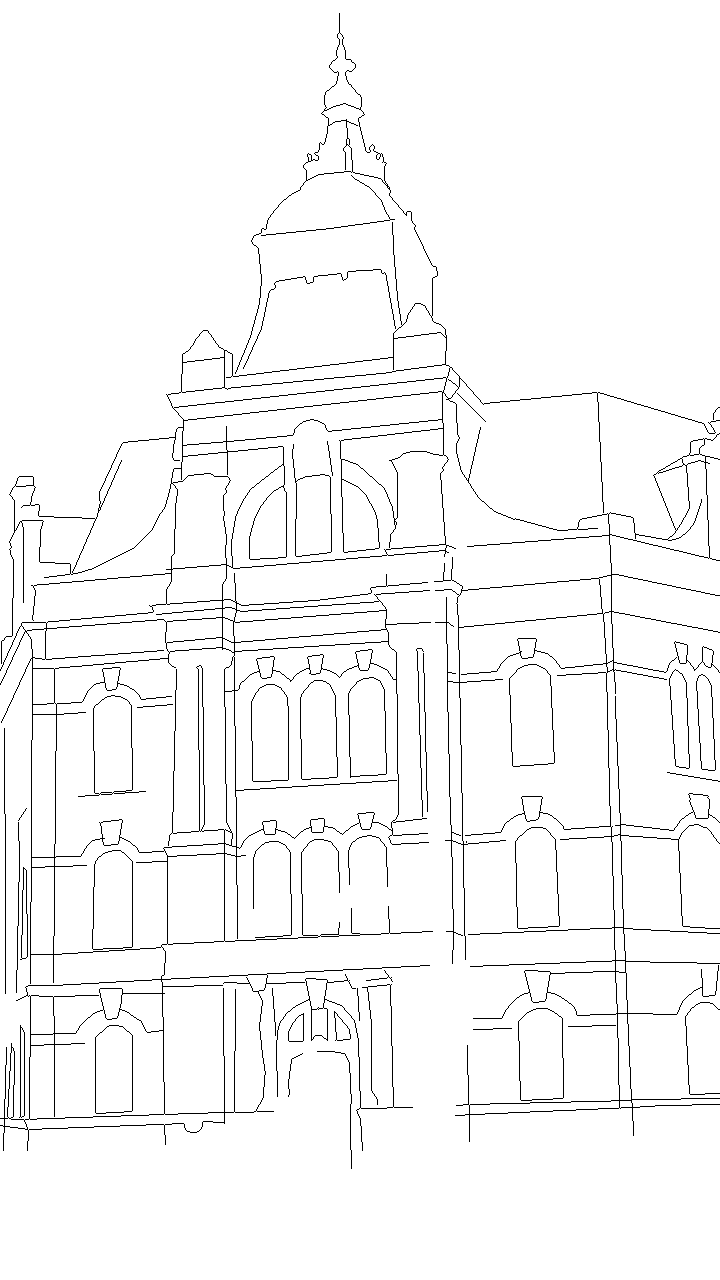}
	&
		\includegraphics[height=0.13\textheight]{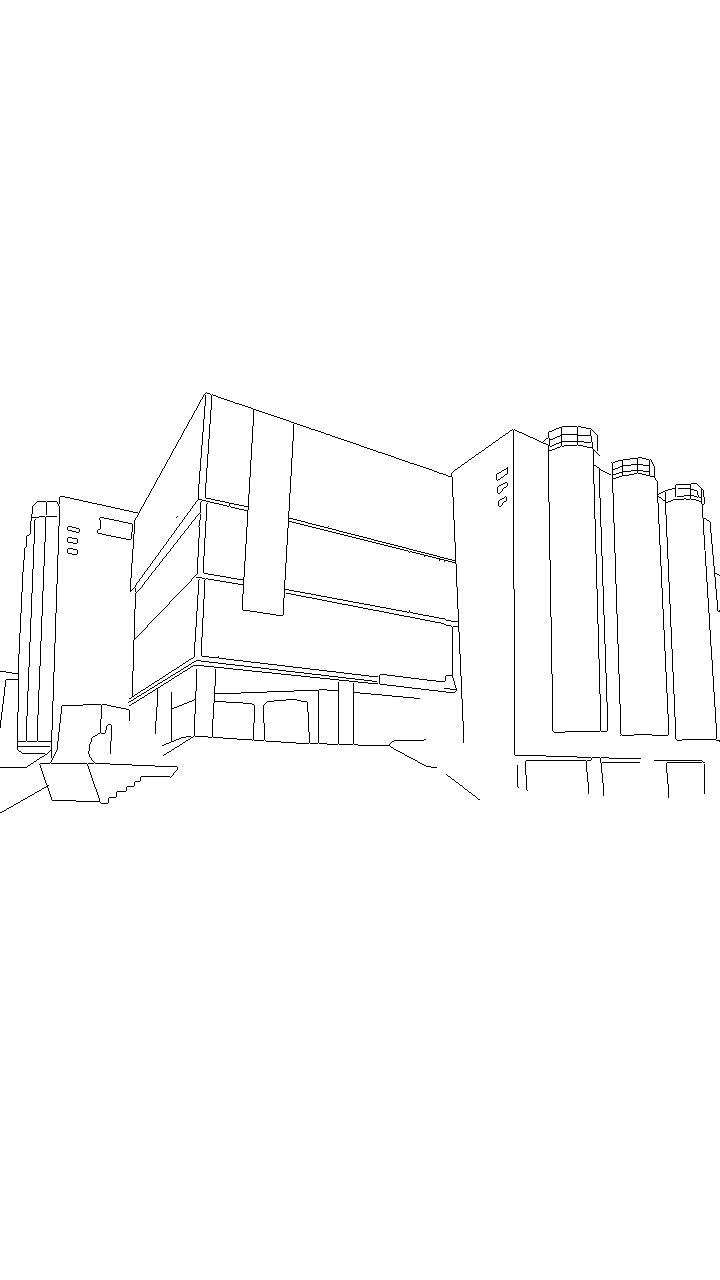}
	&
		\includegraphics[height=0.13\textheight]{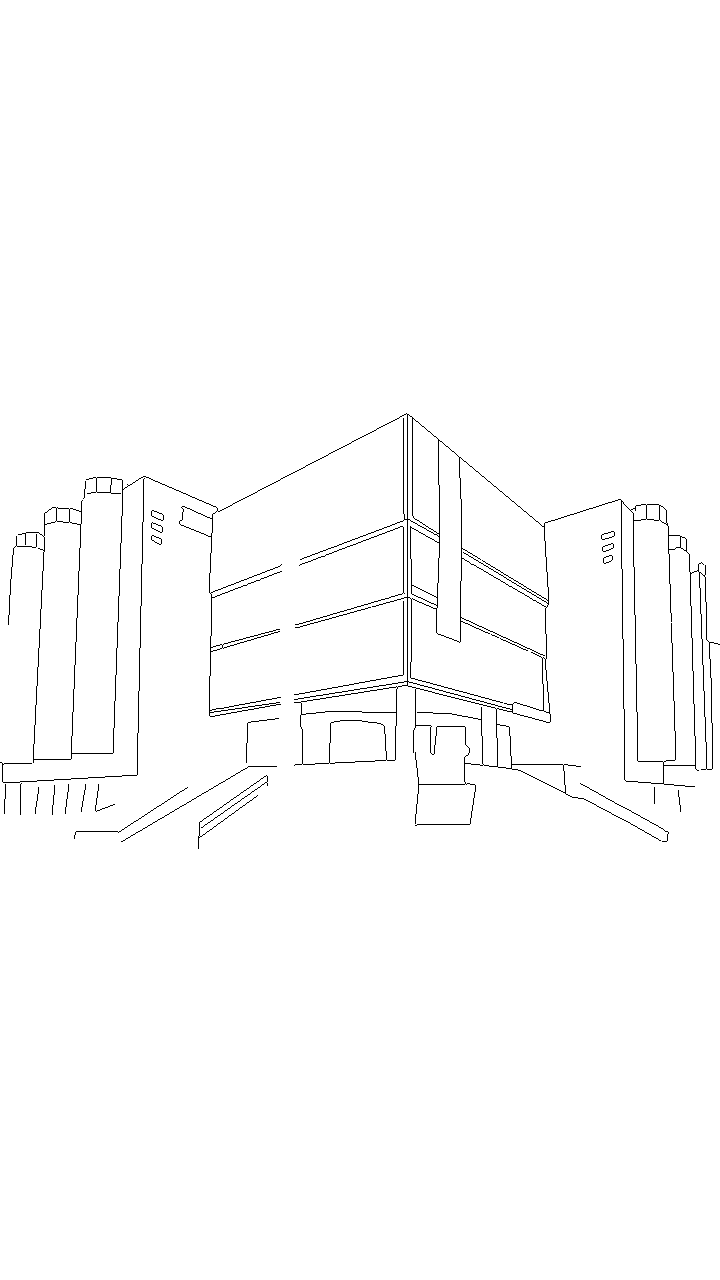}
\\
		\includegraphics[height=0.13\textheight]{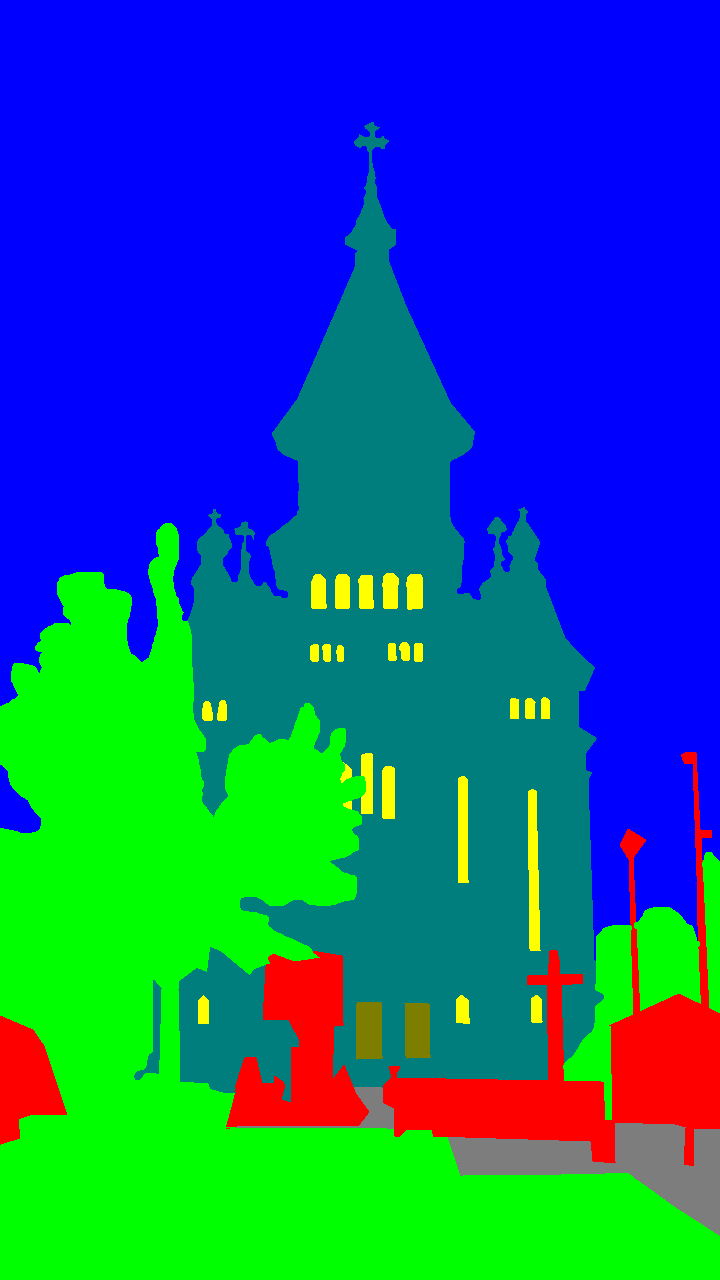}
	&
		\includegraphics[height=0.13\textheight]{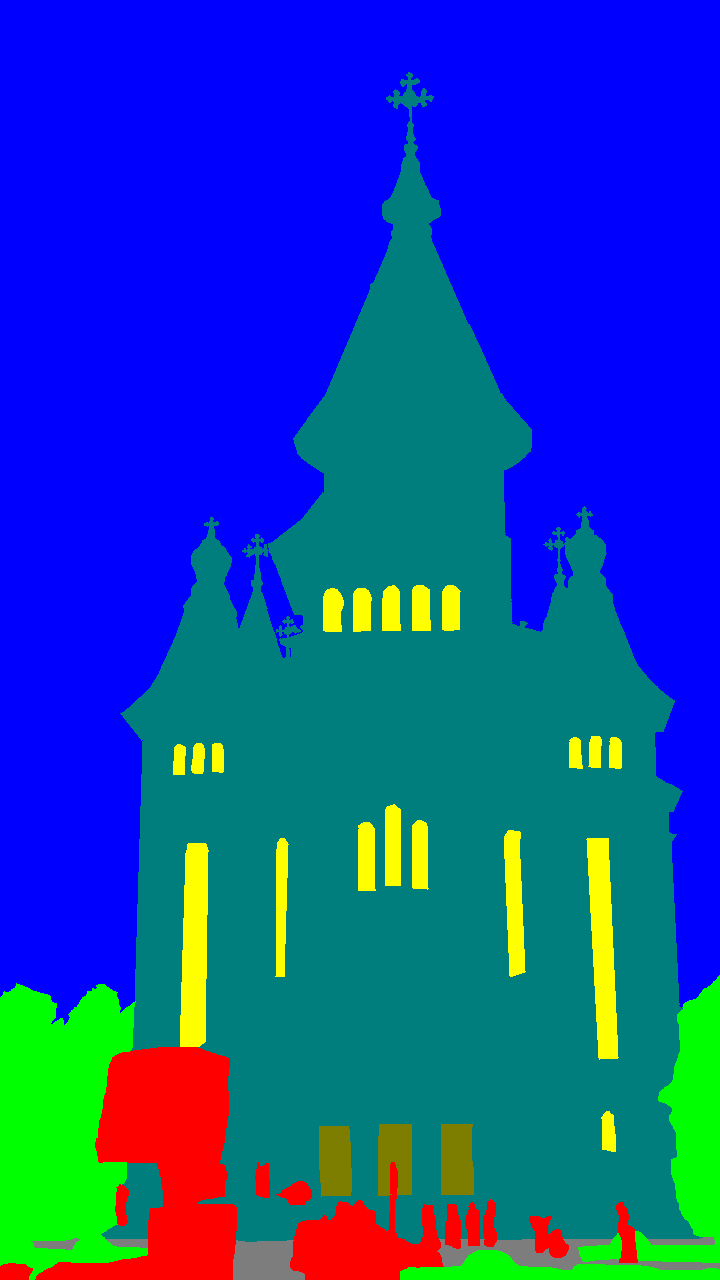}
	&
	    \includegraphics[height=0.13\textheight]{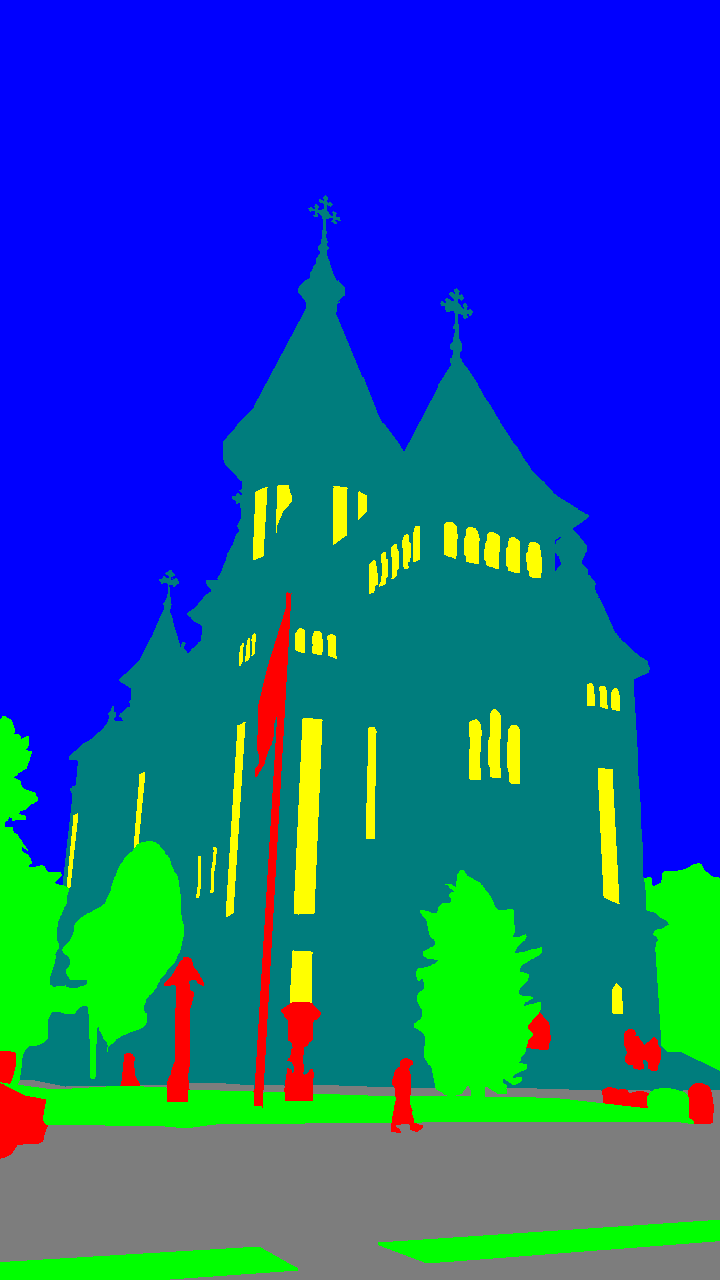}
	&
		\includegraphics[height=0.13\textheight]{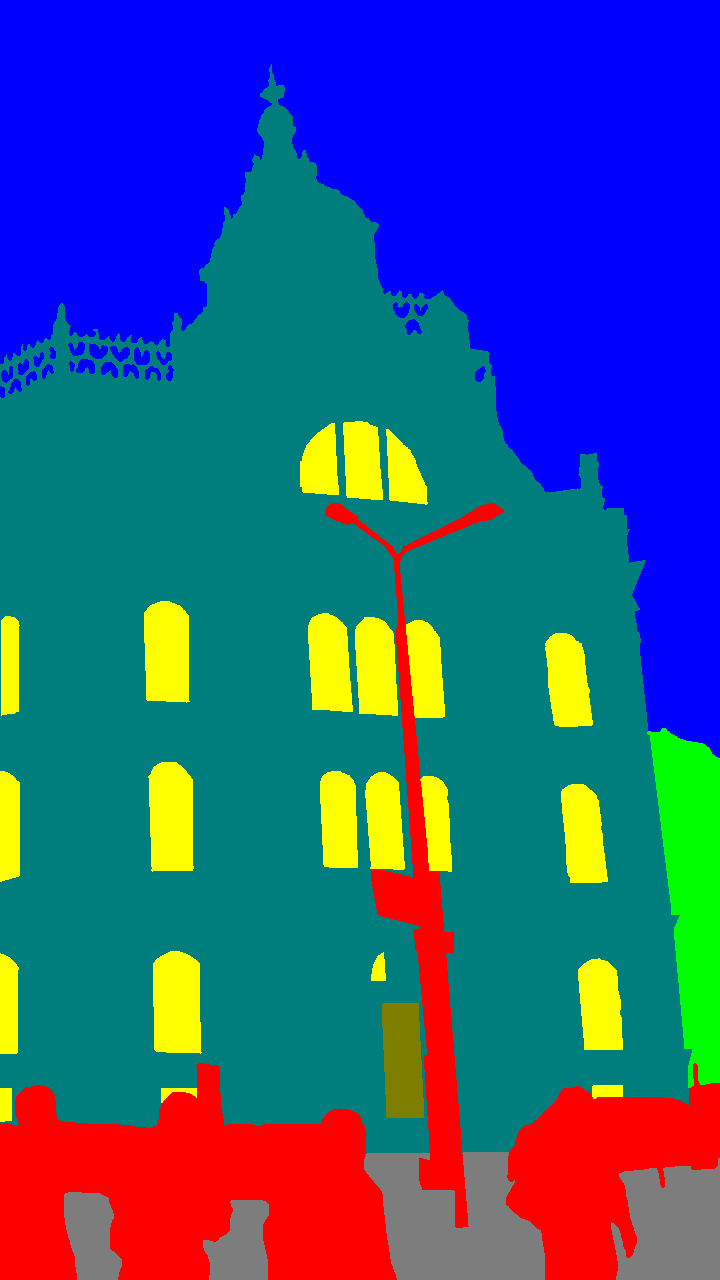}
	&
		\includegraphics[height=0.13\textheight]{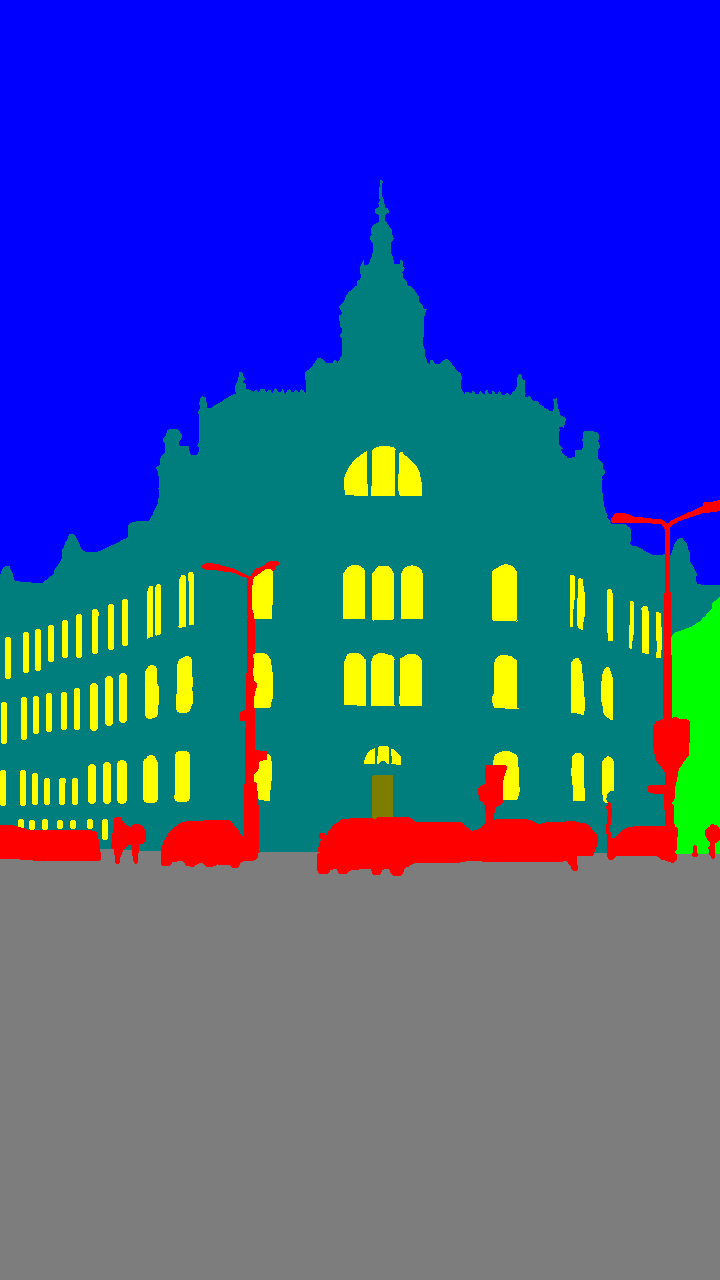}
	&
		\includegraphics[height=0.13\textheight]{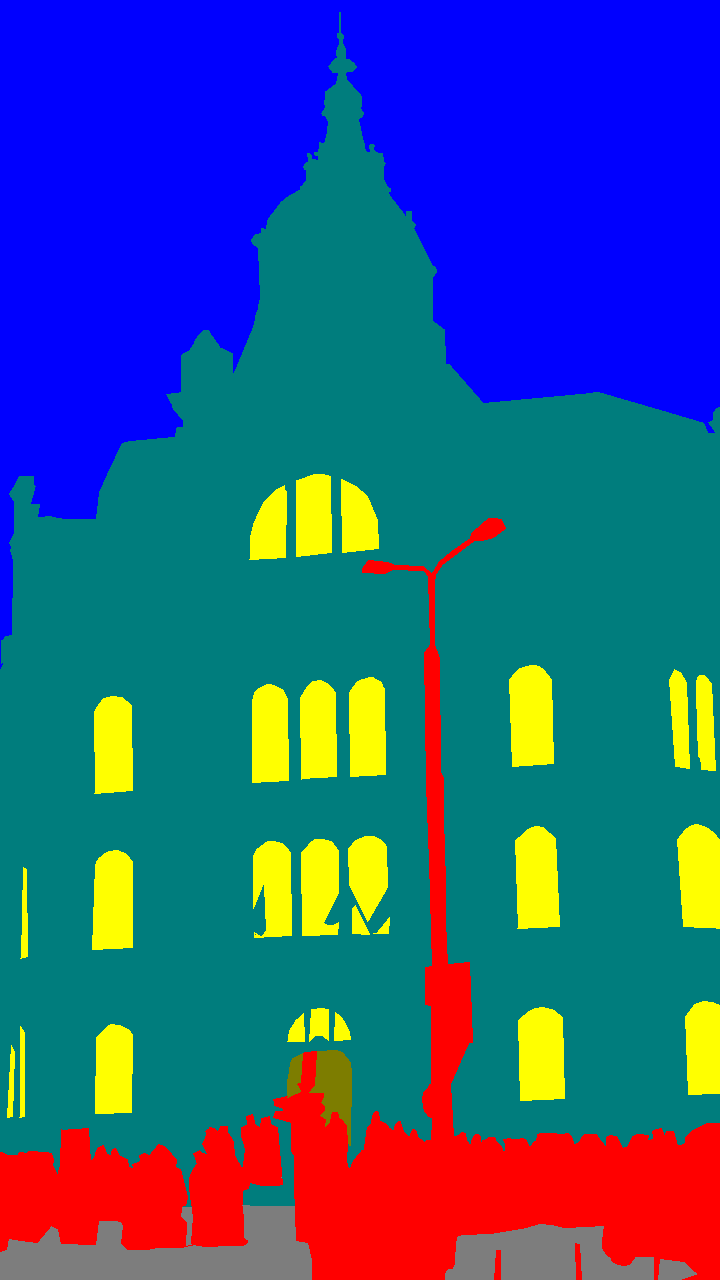}
	&
		\includegraphics[height=0.13\textheight]{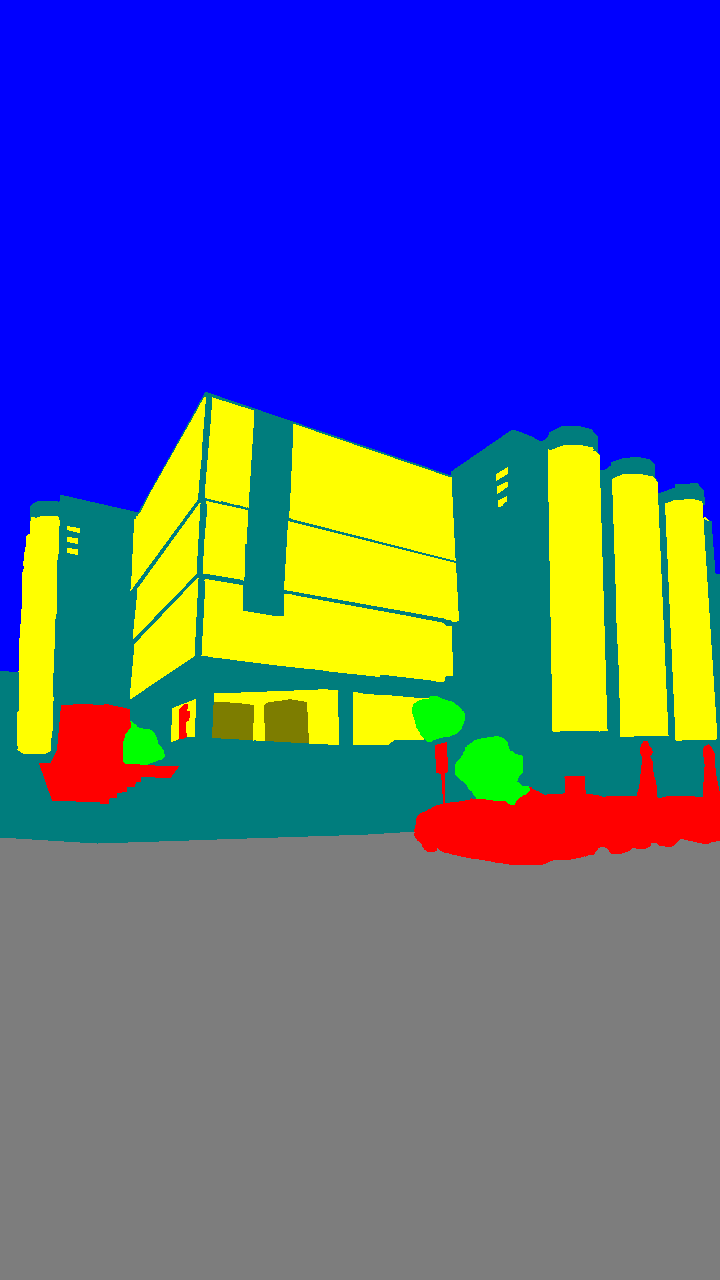}
	&
		\includegraphics[height=0.13\textheight]{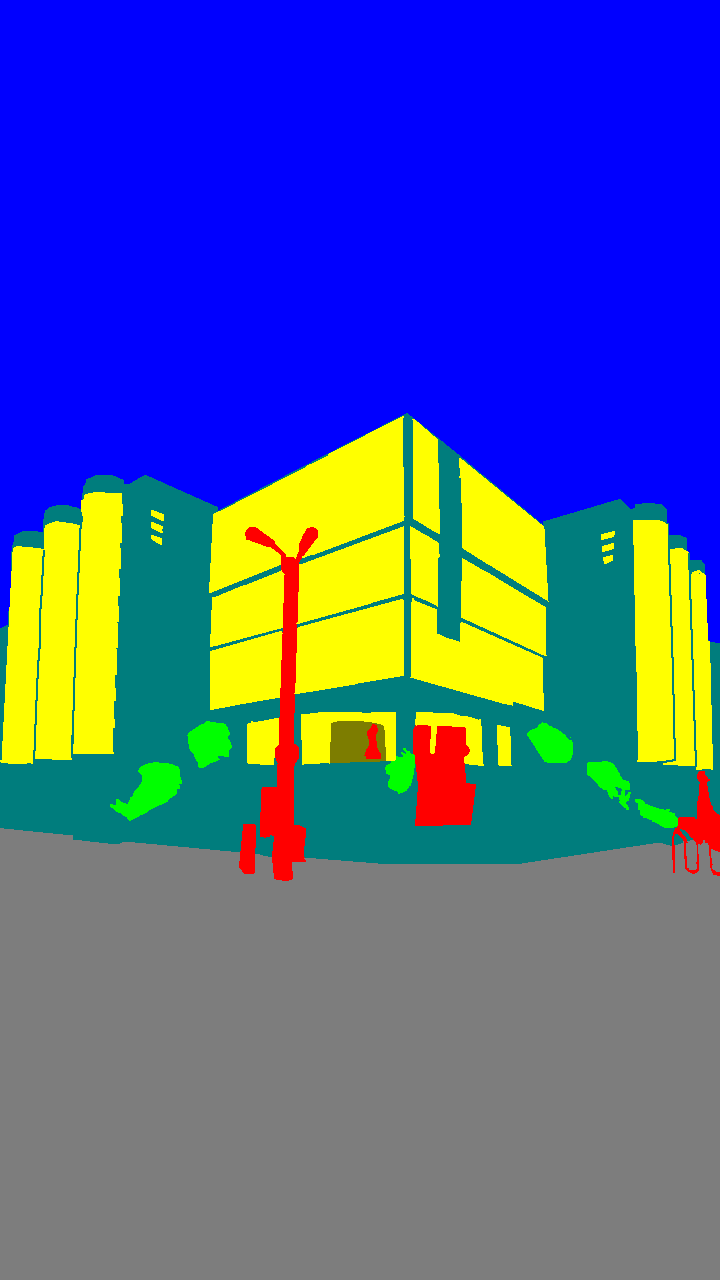}

\end{tabular}
\caption{Images from proposed dataset. Rows: original image, edge ground-truth, label ground-truth }
\label{fig:dataset_labels_proposed}
\end{figure}

TMBuD is created from images of buildings in Timisoara. Each building is presented from several perspectives, so this dataset can be used for evaluating a building detection algorithm too. The dataset contains ground-truth images for salient edges, for semantic segmentation and the GPS coordinates of the buildings. The dataset contains 160 images grouped in the following sets: 100 consist of the training dataset, 25 consist of the validation data and 35 consist of the test data. We can see examples of images from the dataset in Figure \ref{fig:dataset_labels_proposed}.

As we can observe in Figure \ref{fig:dataset_labels_proposed}, the database focuses on a view that will be available if the input sensor device is a mobile phone. We consider this to be a very important aspect because the main domain where the building detection algorithm is used is the AR domain. Even if the edge features are focused only in the building area we desire to offer a full understanding of the environment via the semantic segmentation label. 

The data was annotated using human subjects that were asked to label (draw) what they perceived as important edges of a building, like the boundaries of the building and differences between facades of the building, different buildings, windows, doors and so on. We asked them not to fill edges or lines that are occluded by other structures even if it's natural that they are present. Secondly they were asked to semantically label the image they created according to the label specification. After this step was finished, we proceeded to unify and correct the edges and labels created by the human subjects into one single ground-truth image. The correction was mainly to eliminate as much as possible the false salient edges that can occur when the data is labeled by a human subject unaware by the inner works of line detection or line matching algorithm. 

We believe that a dataset is useful for evaluating algorithms for facade detection or building if it is focused on the building itself present in the image. Firstly, the images should be selected having in mind the street view perspective and uniqueness of features available on them. Secondly, the ground truth images should offer a basis for evaluating boundaries - an important aspect indeed -, but to offer a solution for evaluating facade edges and boundaries.

Boundary detection and edge detection are similar but not identical. Edges represent discontinuities of brightness which are usually found using low-level CV processes. The process of feature extraction of edges works under the assumption of ideal edge models. Boundary detection is viewed as a mid-level process of finding margins of objects in scenes. This task has close ties both with grouping/segmentation and object shape detection \cite{ren2008multi}.

If we analyse the available datasets and benchmarks for edge detection, as we did in Section \ref{fig:edge_example}, we can observe that they focus on edges or boundaries generated from all structures from the image. Of course as we can see in Figure \ref{fig:edge_example} they clearly focus on evaluating and training edge algorithm for natural scenes and do not consider a certain use case of the resulting features. 

\begin{figure}[h!]
\centering
\begin{tabular}{cccc}
& \includegraphics[width=0.2\columnwidth]{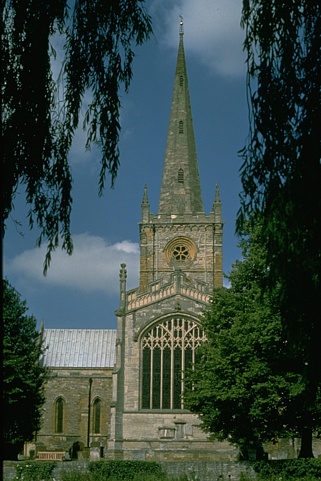}
& \includegraphics[width=0.2\columnwidth]{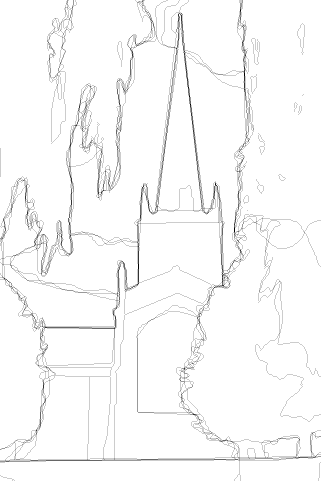}
& \includegraphics[width=0.2\columnwidth]{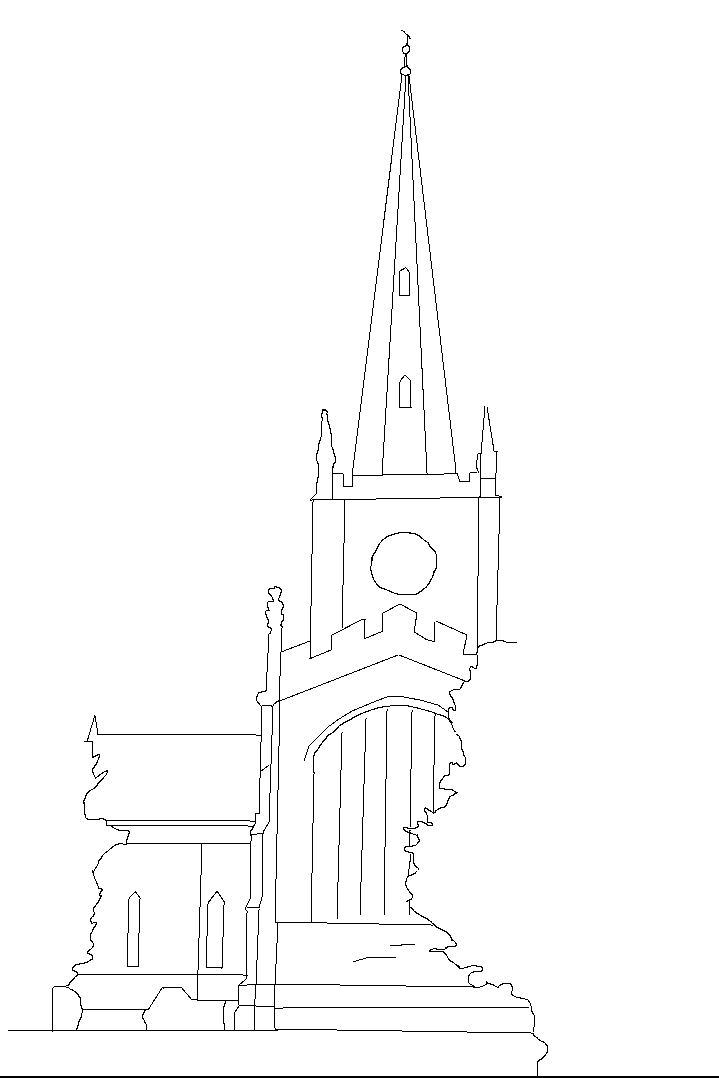}
\\
&BSDS500 image
&BSDS500 gt image
&Our gt image
\end{tabular}
\caption{Images and ground truth}
\label{fig:diff_dataset_example}
\end{figure}

Modern urban building detection techniques like \cite{yi2019semantic}, \cite{wang2020reconstruction} use line segment matching for performing this task. In parallel, the domain of line feature matching \cite{line_match_ex1}, \cite{li2016joint} for finding relevant features is growing, bringing to the table new solutions for this complex problem. In this scope we consider that our proposed approach for annotating the edges for a dataset becomes more relevant. 

As we can observe in Figure \ref{fig:diff_dataset_example}, we concern ourselves with salient edges produced by the details or shape of the building and ignore the edges produced by adjacent structures in the image, such as persons, cars, sky, ground and so on. We consider that this will help better fine tune the line features extraction algorithms that concern with building detection. 

Semantic segmentation is an important aspect in the field of computer vision. The importance of scene understanding is highlighted by the fact that an increasing number of applications emerge from inferring knowledge from imagery. This step in the pipeline has become more popular in object detection applications, even if we talk about building detection.

The proposed dataset focuses more on the scene understanding of the environment rather than on semantically understanding the structures of the building. As we can see in Figure \ref{fig:dataset_labels_proposed}, the existing datasets offer solid grounds for training and evaluating semantic segmentation solutions but lack a certain capability to be used to fine tune a semantic segmentation for urban building scenario (as we can see from the last column where we made a transition from there label scheme to ours).

In Table \ref{table:dataset_correlation} we can observe the existing classes offered by TMBuD, by value and RGB code, and the corresponding classes from the datasets presented in Section \ref{Sec:SemSegDataset}. Most of the classes are self explanatory but by correlating the classes from TMBuD with other dataset we aim to explain our view for segmenting the environment. We consider it essential to differentiate between BACKGROUND, that we consider unclassified data, and NOISE that we consider elements or objects that appear temporary in the field of view, such as cars, people, terraces, human made temporary structure and so on. 

\begin{table}[h!]
\begin{center}
\setlength{\tabcolsep}{2pt}
\scalebox{0.55}
{
\begin{tabular}{|l|c|c|c|c|c|c|c|c|c|}
\hline
TMBuD           &Label  &RGB            &eTRIMS         &LabelMeFacade  &ECP                   &ICG Graz5   &INRIA      &CPM            &VarCity\\
\hline
\hline
BACKGROUND      &0      &(0,0,0)        &Not labeled    &VARIOUS        &OUTLIER                &-          &VARIOUS    &BACKGROUND     &BACKGROUND\\
\hline
BUILDING        &1      &(125,125,0)    &BUILDING       &BUILDING       &WALL                   &WALL       &WALL       &FACADE         &WALL\\
~               &~      &~              &~              &~              &BALCONY                &~          &BALCONY    &CORNICE        &BALCONY\\ 
~               &~      &~              &~              &~              &ROOF                   &~          &ROOF       &SILL           &ROOF\\
~               &~      &~              &~              &~              &CHIMNEY                &~          &SHOP       &BALCONY        &SHOP\\
~               &~      &~              &~              &~              &SHOP                   &~          &~          &MOLDING        &~ \\
~               &~      &~              &~              &~              &~                      &~          &~          &DECO           &~ \\
~               &~      &~              &~              &~              &~                      &~          &~          &PILLAR         &~ \\
~               &~      &~              &~              &~              &~                      &~          &~          &SHOP           &~ \\
\hline
DOOR            &2      &(0,125,125)    &DOOR           &DOOR           &DOOR                   &DOOR       &DOOR       &DOOR           &DOOR\\
\hline
WINDOW          &3      &(0,255,255)    &WINDOW         &WINDOW         &WINDOW                 &WINDOW     &WINDOW     &WINDOW         &WINDOW\\
~               &~              &~              &~                      &~          &~          &BLIND          &~\\
\hline
SKY             &4      &(255,0,0)      &SKY            &SKY            &SKY                    &SKY        &SKY        &BACKGROUND     &SKY\\
\hline
VEGETATION      &5      &(0,255,0)      &VEGETATION     &VEGETATION     &-                      &-          &-          &-              &~\\
\hline
GROUND          &6      &(125,125,125)  &PAVEMENT       &PAVEMENT       &-                      &-          &-          &-              &~\\
~               &~      &~              &ROAD           &ROAD           &-                      &-          &-          &-              &~\\
\hline
NOISE           &7      &(0,0,255)      &CAR            &CAR            &-                      &-          &-          &-              &~\\
\hline

\end{tabular}}
\vspace{1.5pt}
\caption{Dataset's corelations} 
\label{table:dataset_correlation}
\end{center}
\end{table}

The TMBuD does not offer a build-in benchmarking capability of edge detection or semantic segmentation but it is part of EECVF \cite{eecvf2020} \cite{eecvf_jurnal}, our Python-based End-To-End CV Framework, where a user can evaluate capabilities of algorithms. The dataset offers the possibility of extending or reorganizing the image in the train - validate - test groups by using a Python module that exists in the repository. 

\section{Conclusion}

In this paper we presented a review o existing boundaries and edges dataset and a review of existing semantic segmentation dataset with the scope of highlighting current evaluation solutions. Afterwards we proposed a dataset that is better fitted to serve the tuning and evaluation of urban scenario building detection algorithms. 

We believe that the proposed TMBuD dataset can facilitate research in image processing when focusing on urban scenarios. TMBuD has two main benefits: the unified evaluating system for several linked problems from this area and the targeted dataset on human made structures in urban scenarios. Both aspects mentioned can become relevant aspects for future development and research work.

TMBuD has proven to be a useful dataset for evaluation when trying to determine the best fitted edge detection variant or the best fitted semantic segmentation model for urban scenarios \cite{eecvf_jurnal} \cite{orhei2021}. From the experience of our work we consider that the proposed dataset as an useful component in constructing a content based image retrieval urban building systems.

We want to expand in the near future the quantity of the dataset images and ground truth images respecting the same principles that we exposed in the paper: important human made structures in urban areas, from a street perspective and different angles.

Regarding the expansion of the dataset we are thinking of including a series of metadata information to be available for each landmark so we can serve algorithms focused on classifying buildings according to facts like: age of the building, architecture style.


\appendix
\newpage
\label{appendix:A}
\setlength{\tabcolsep}{2pt}
\section*{Appendix A}

\begin{figure}[h!]
\centering
\begin{tabular}{cccccccccc} 
		\includegraphics[height=0.1\textheight]{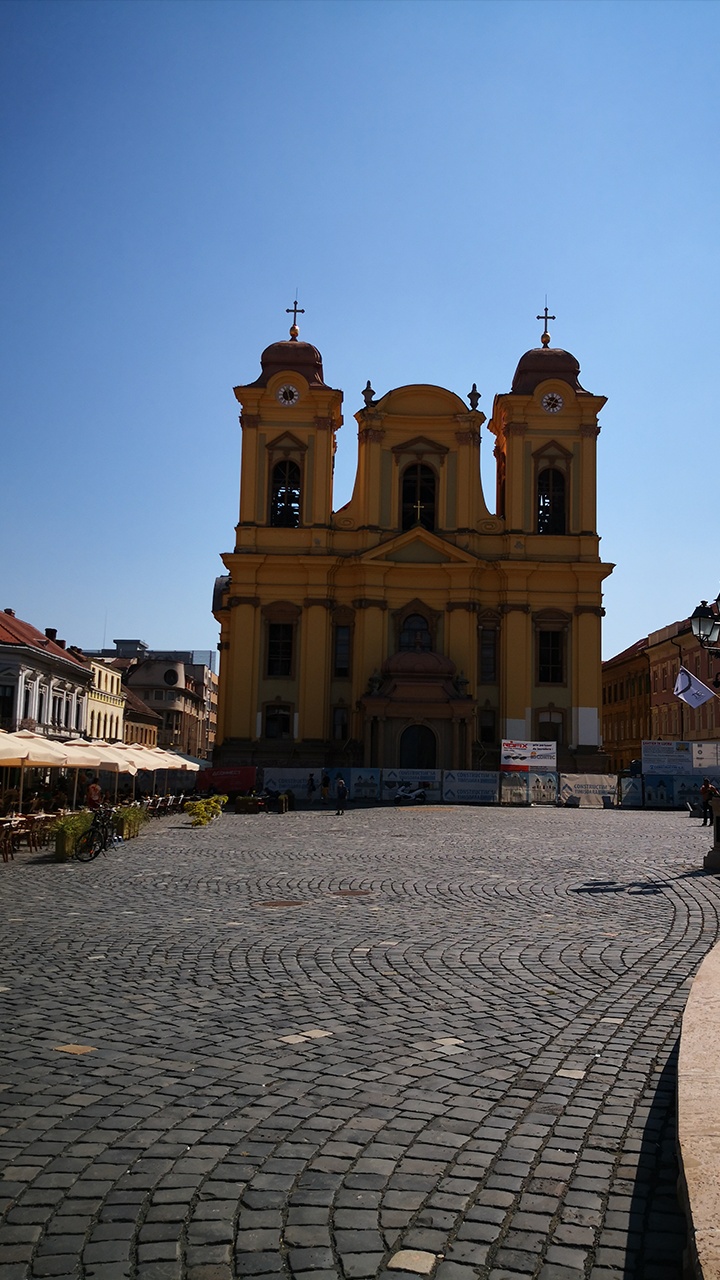}
	&
		\includegraphics[height=0.1\textheight]{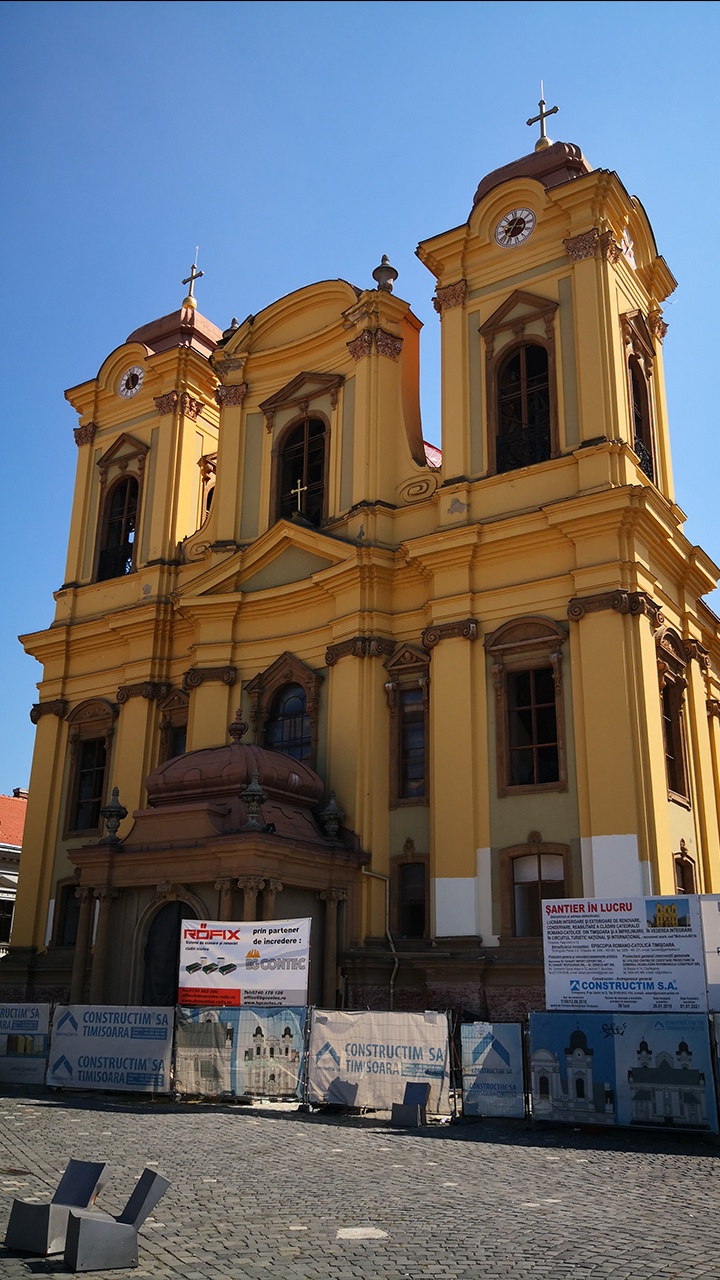}
	&
	    \includegraphics[height=0.1\textheight]{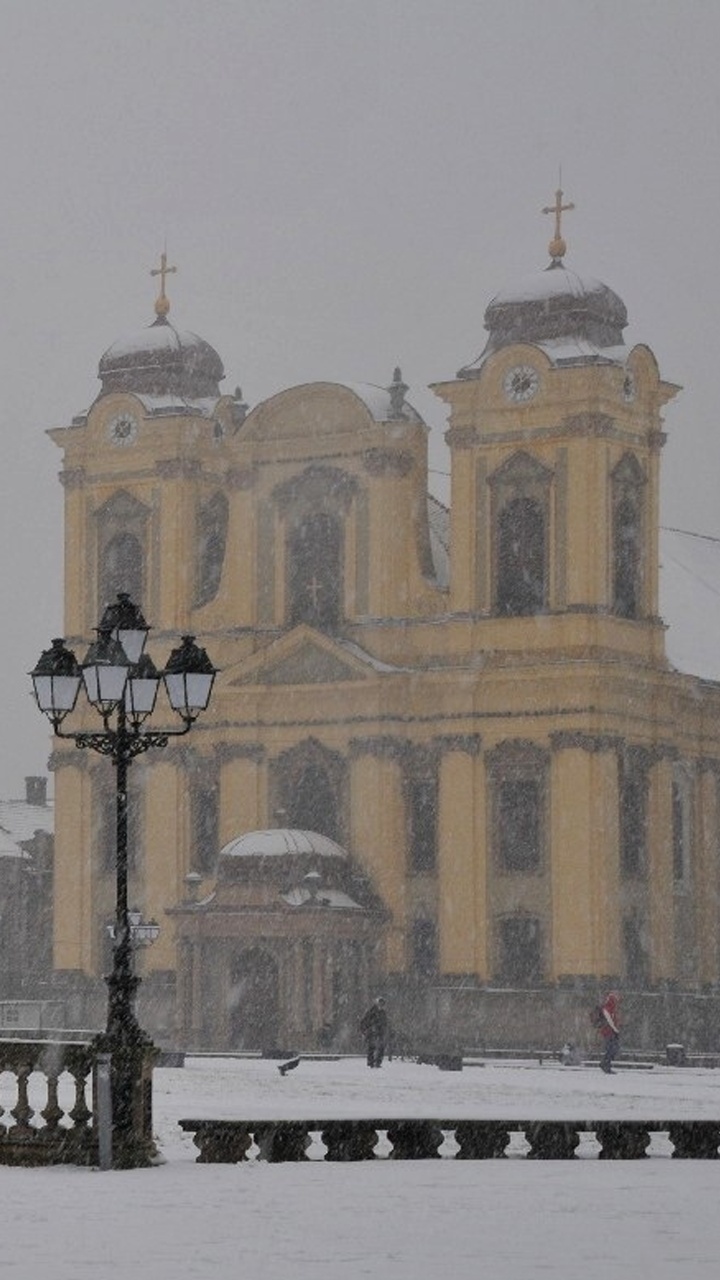}
	&
		\includegraphics[height=0.1\textheight]{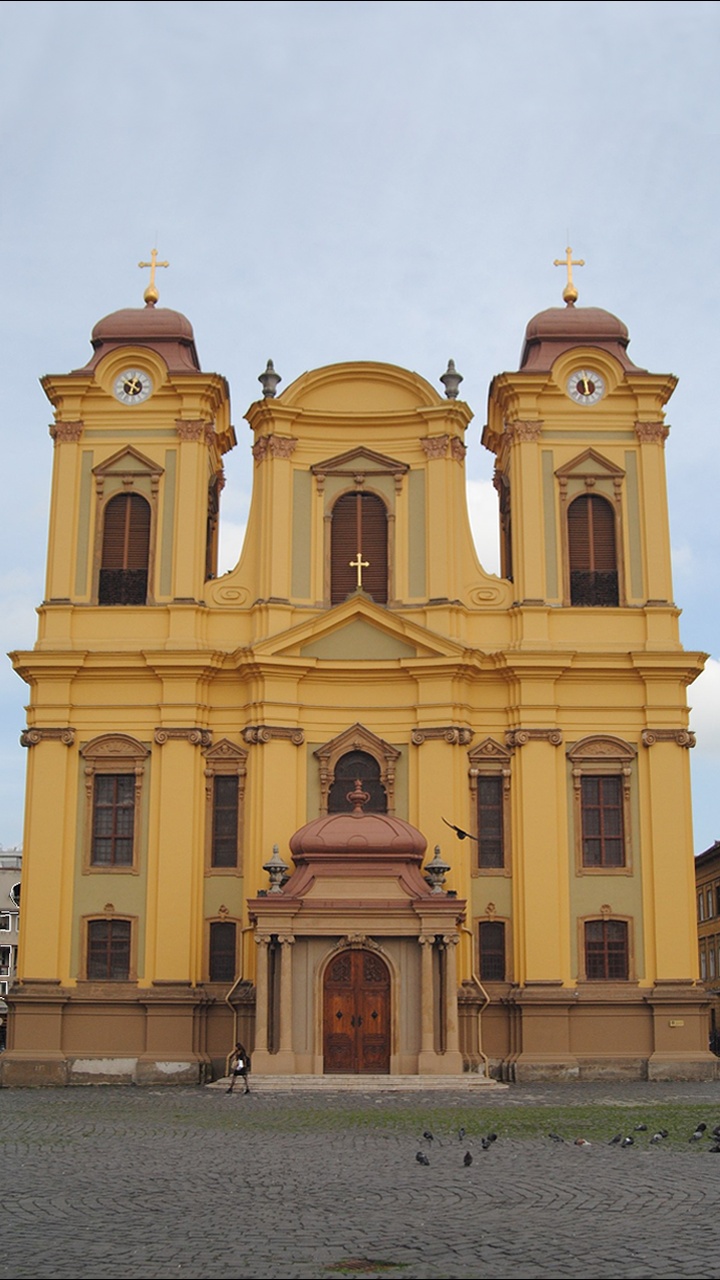}
	&
		\includegraphics[height=0.1\textheight]{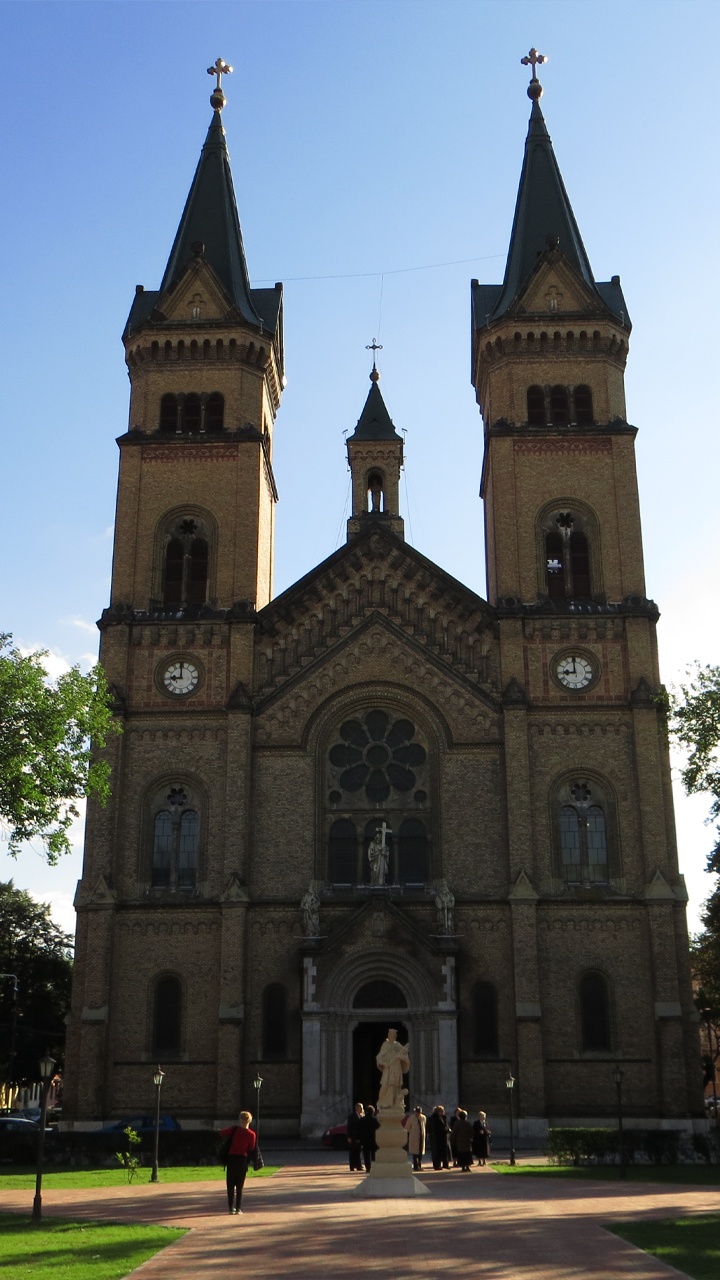}
	&
		\includegraphics[height=0.1\textheight]{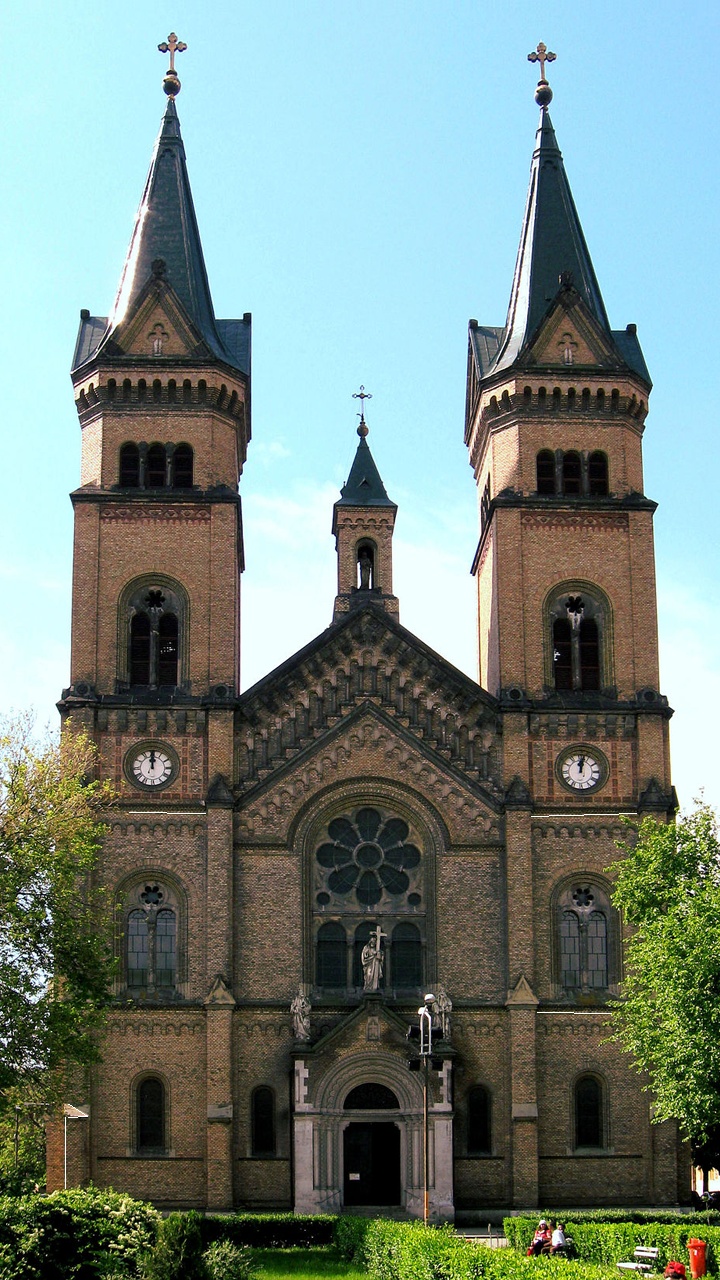}
	&
		\includegraphics[height=0.1\textheight]{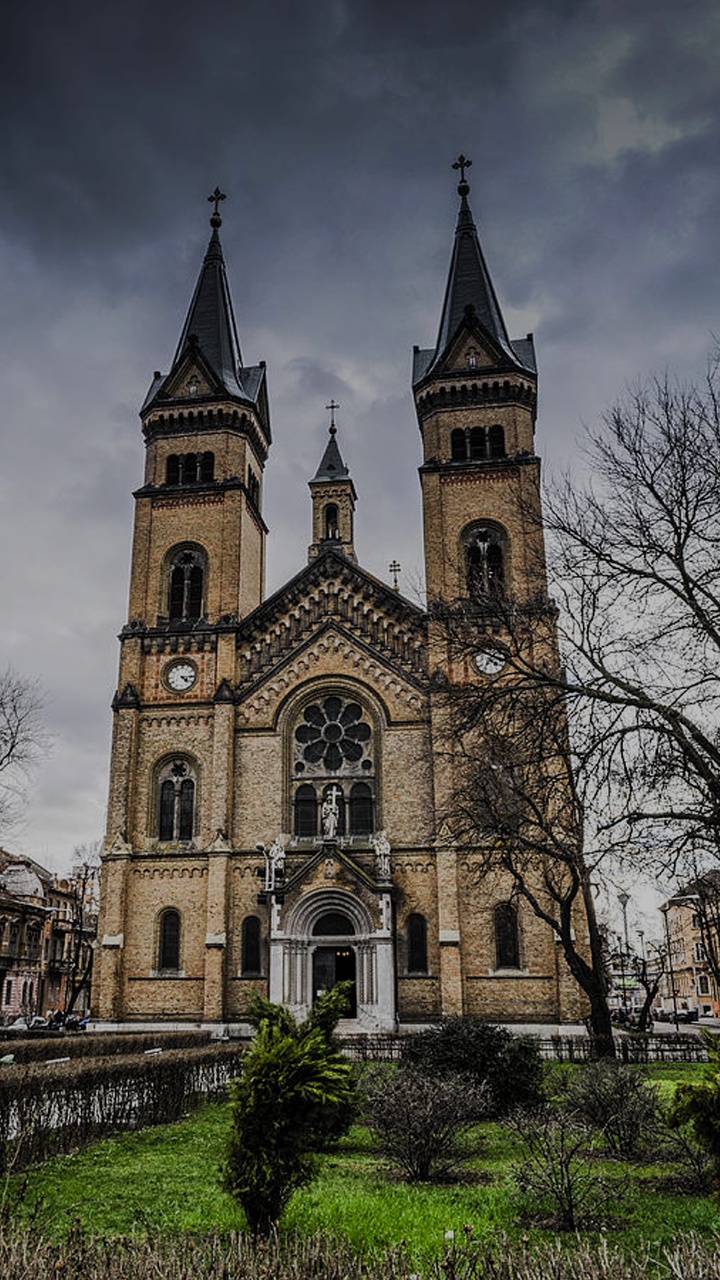}
	&
		\includegraphics[height=0.1\textheight]{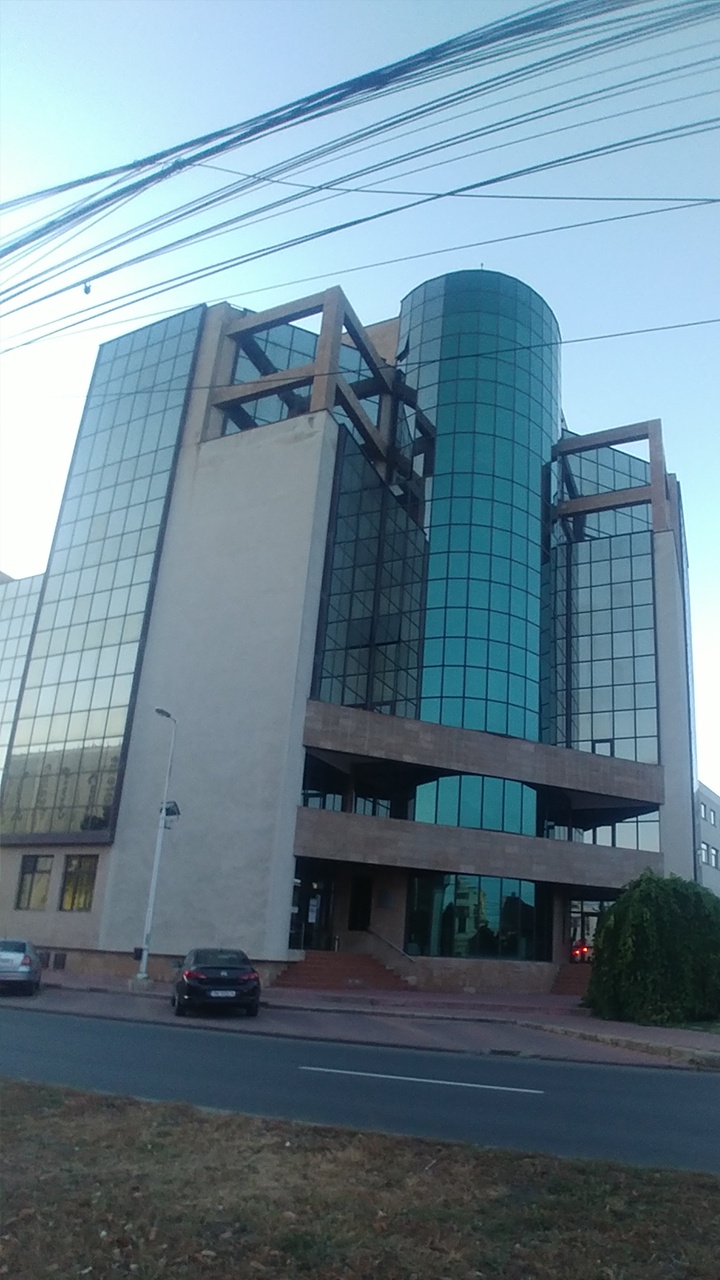}
	&
		\includegraphics[height=0.1\textheight]{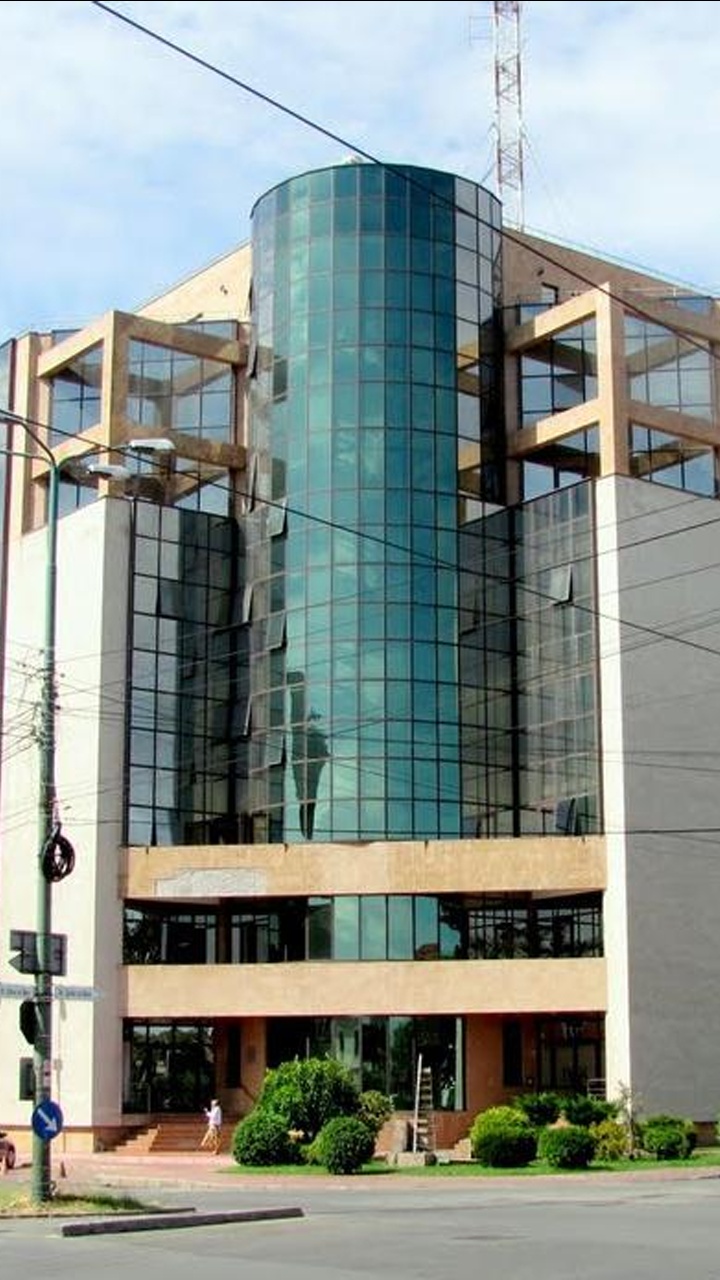}
	&
		\includegraphics[height=0.1\textheight]{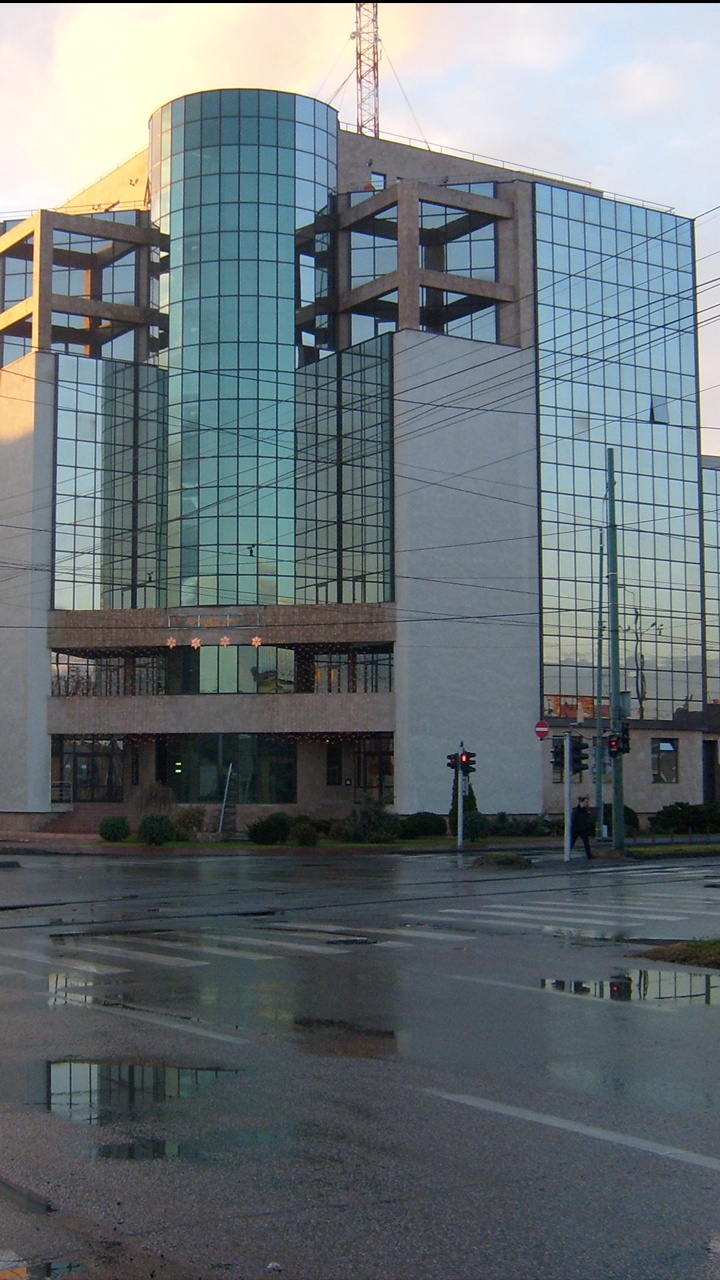}
\\
		\includegraphics[height=0.1\textheight]{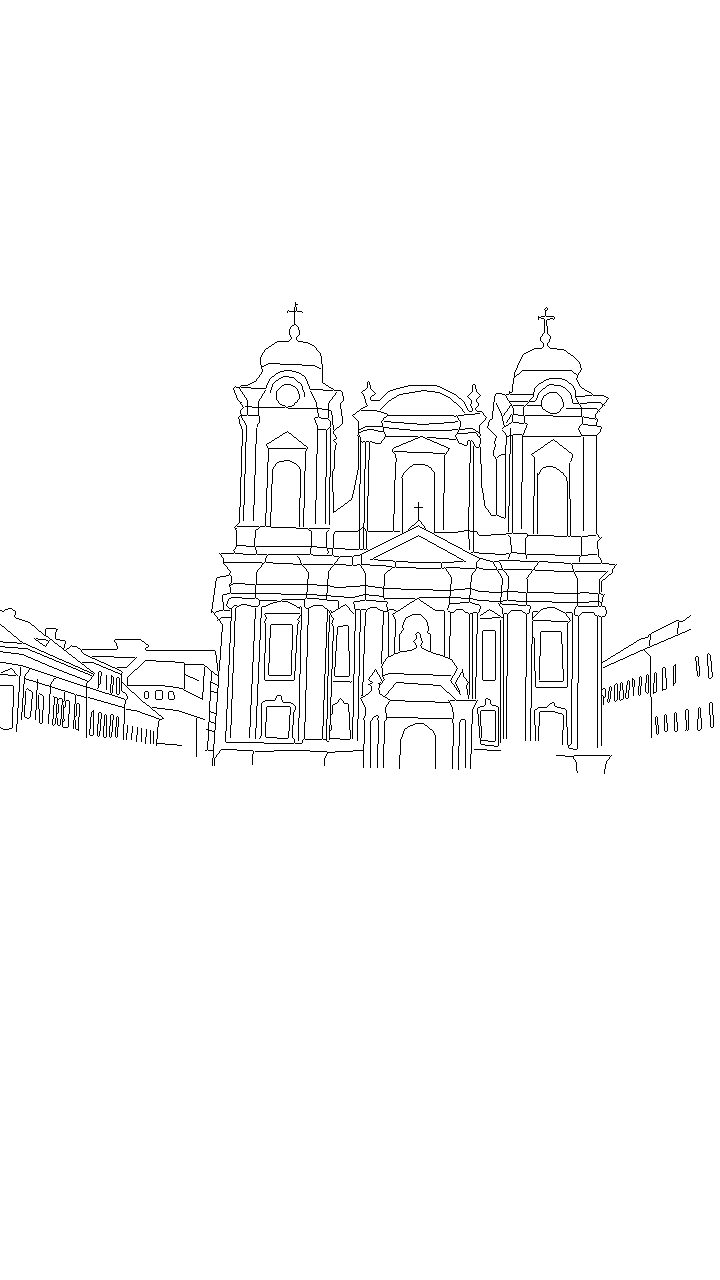}
	&
		\includegraphics[height=0.1\textheight]{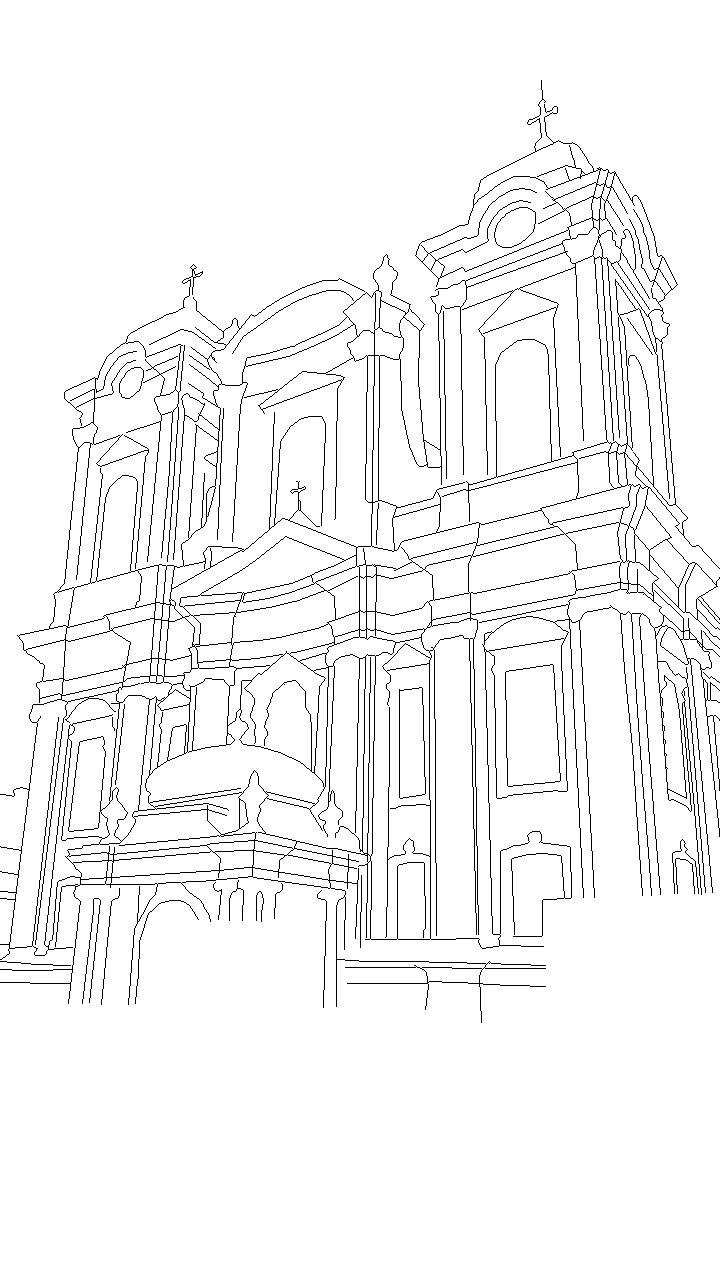}
	&
	    \includegraphics[height=0.1\textheight]{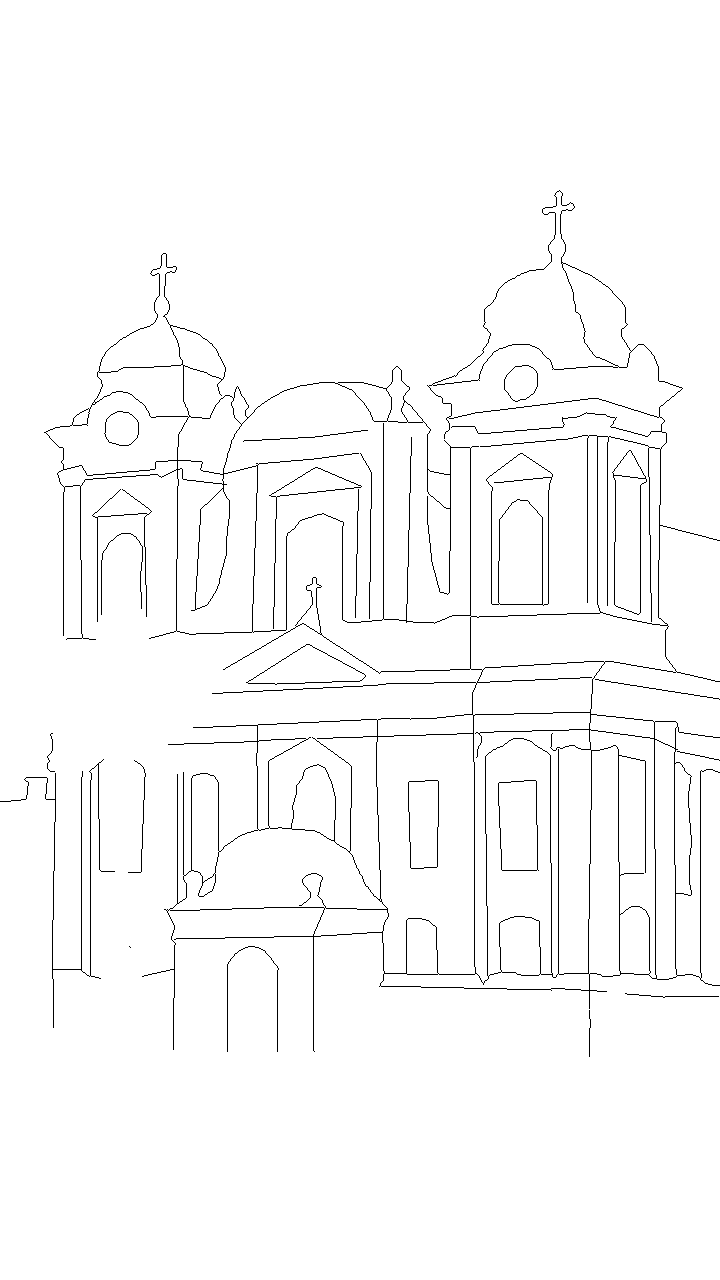}
	&
		\includegraphics[height=0.1\textheight]{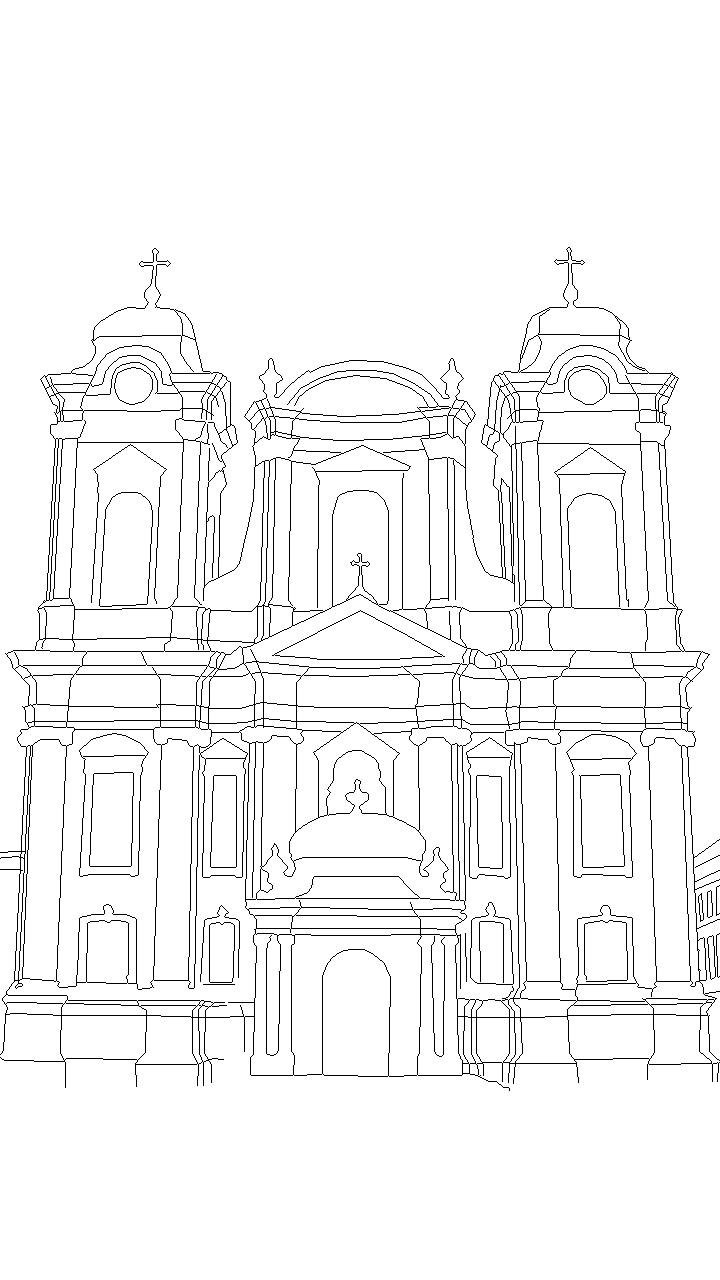}
	&
		\includegraphics[height=0.1\textheight]{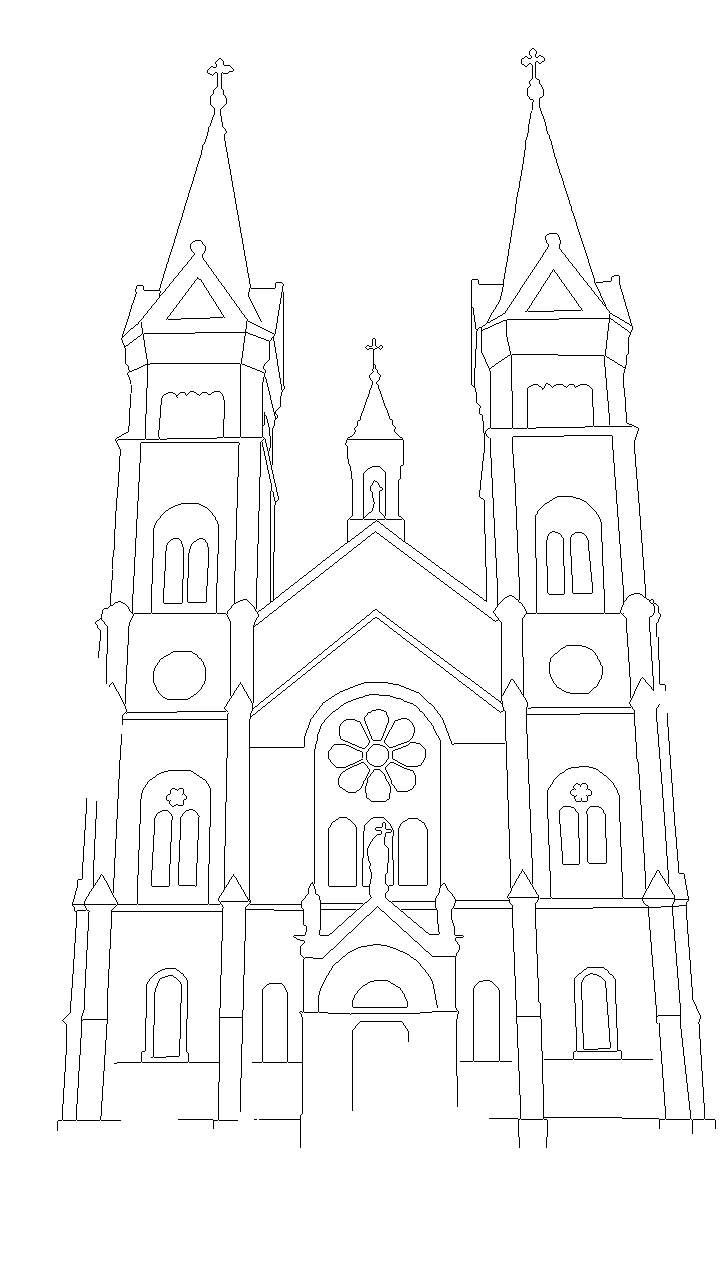}
	&
		\includegraphics[height=0.1\textheight]{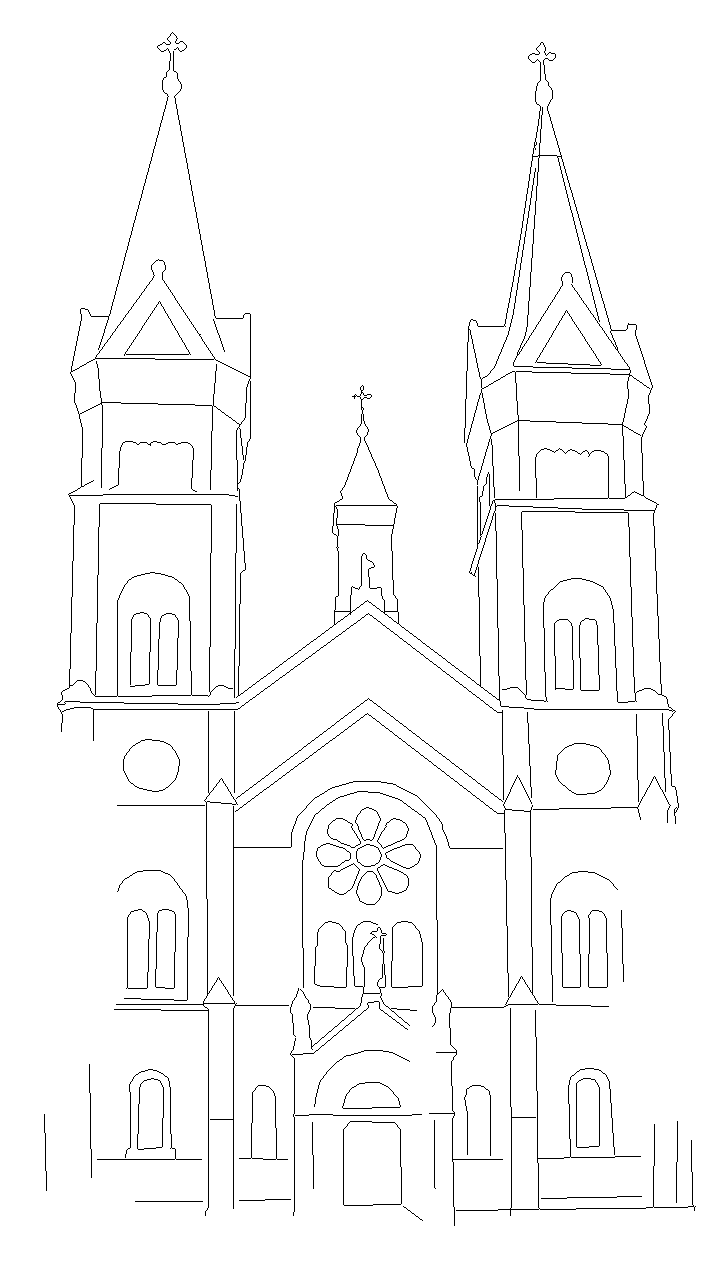}
	&
		\includegraphics[height=0.1\textheight]{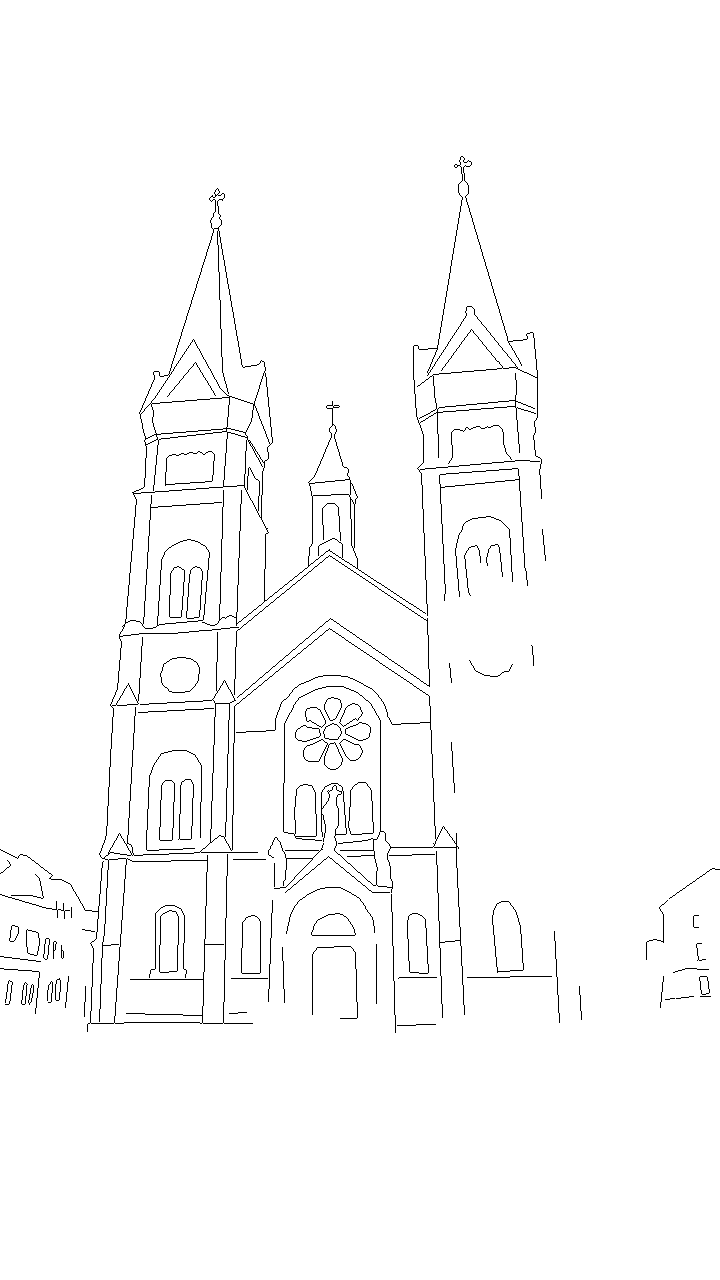}
	&
		\includegraphics[height=0.1\textheight]{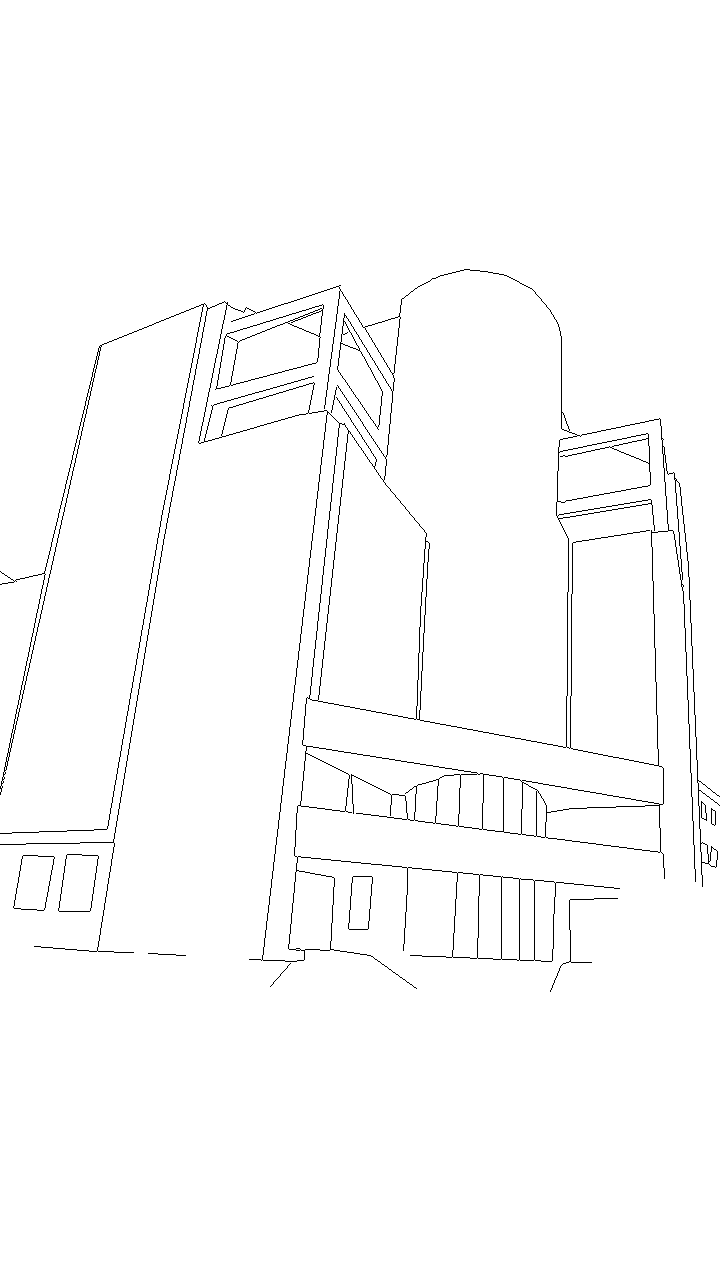}
	&
		\includegraphics[height=0.1\textheight]{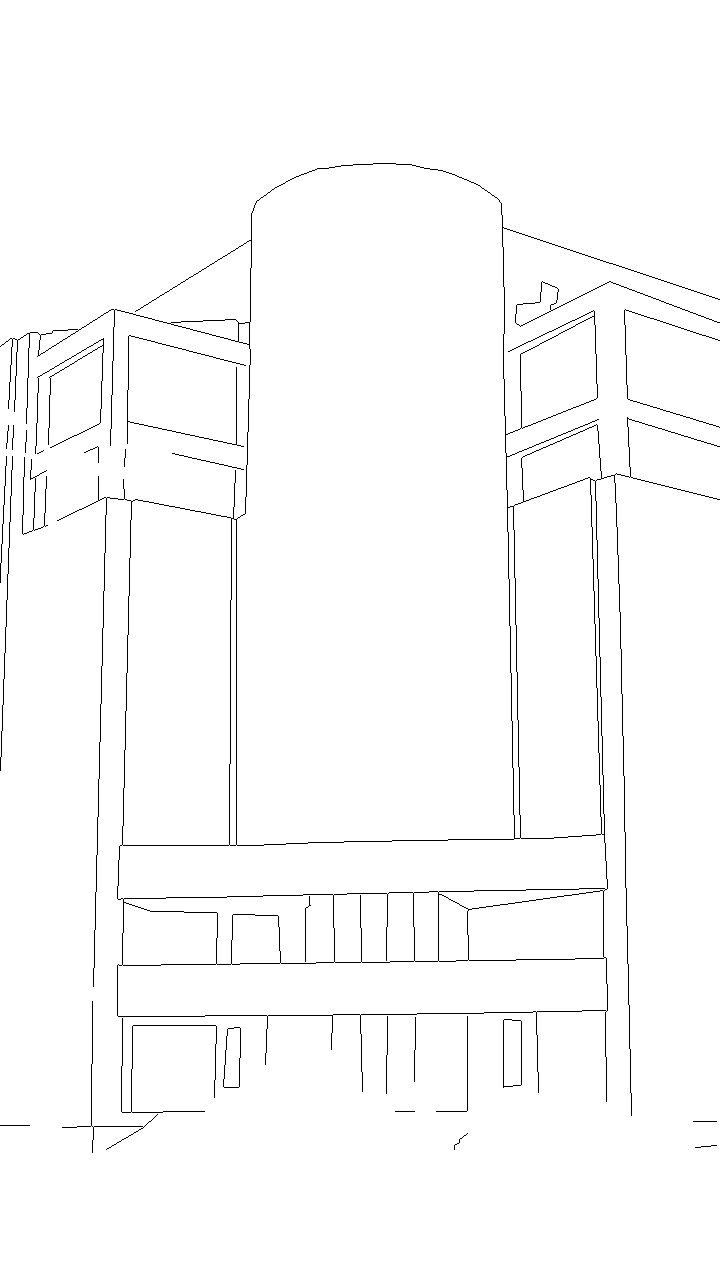}
	&
		\includegraphics[height=0.1\textheight]{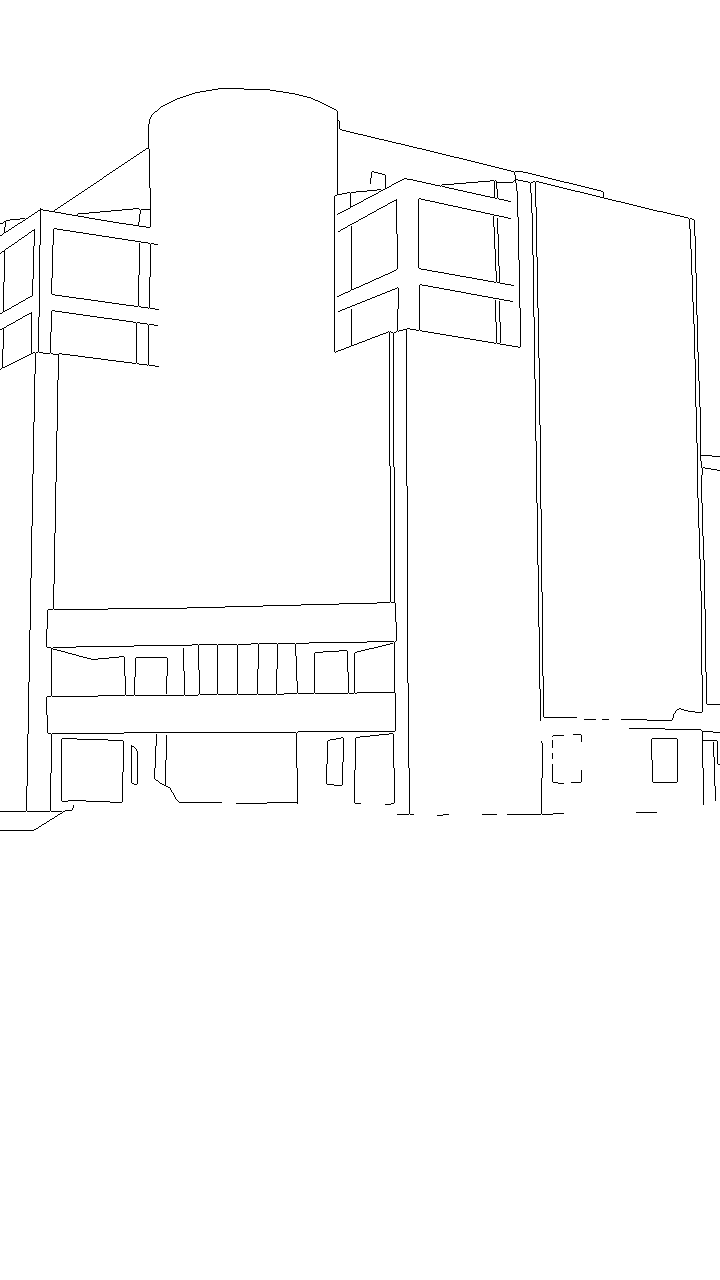}
\\
		\includegraphics[height=0.1\textheight]{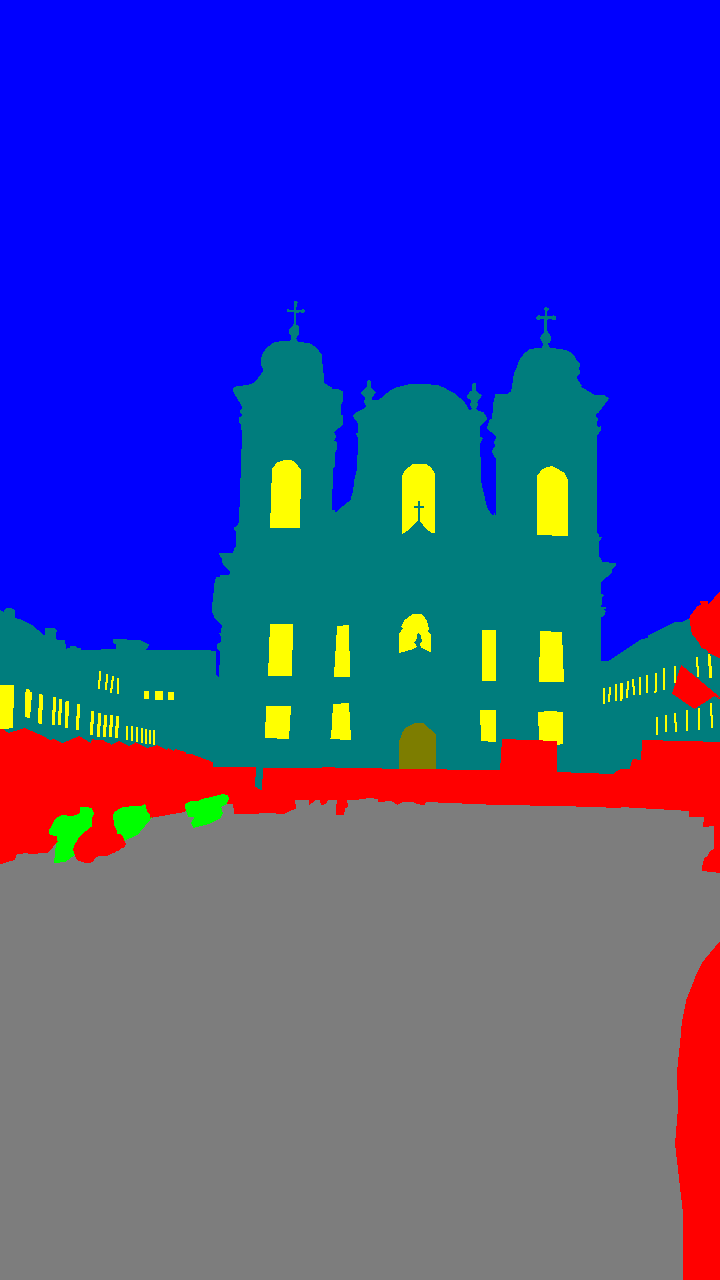}
	&
		\includegraphics[height=0.1\textheight]{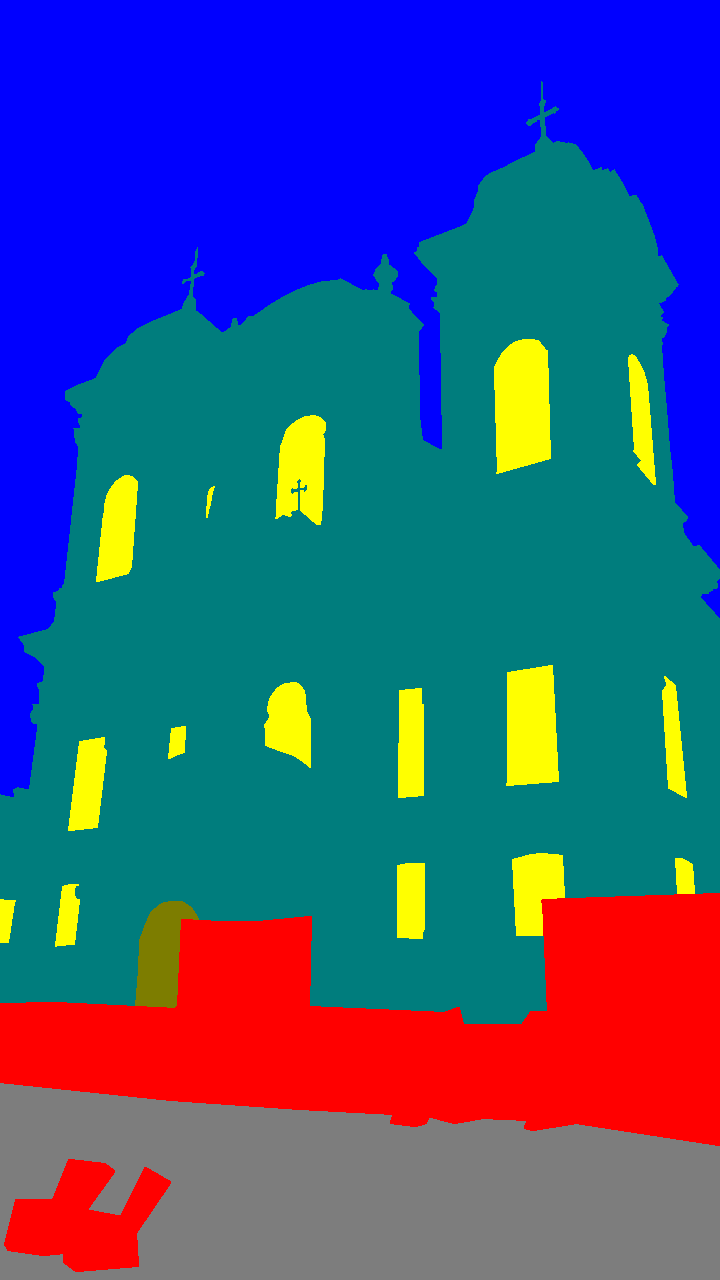}
	&
	    \includegraphics[height=0.1\textheight]{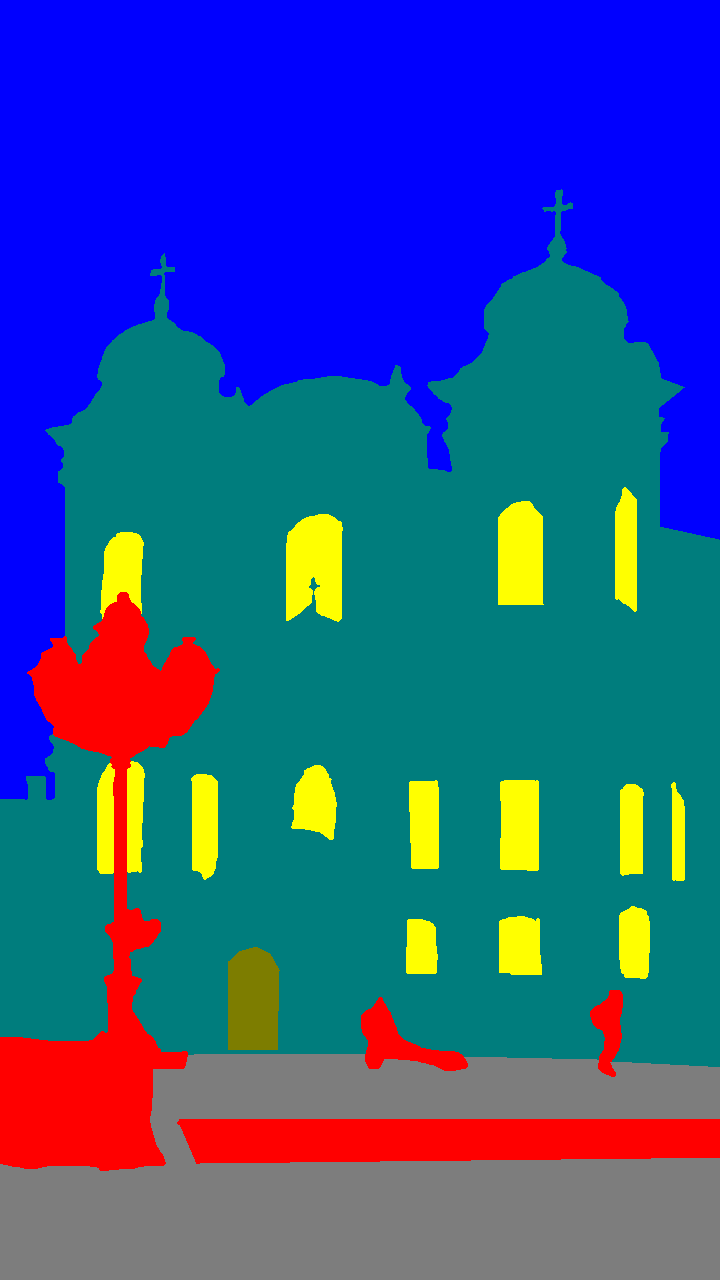}
	&
		\includegraphics[height=0.1\textheight]{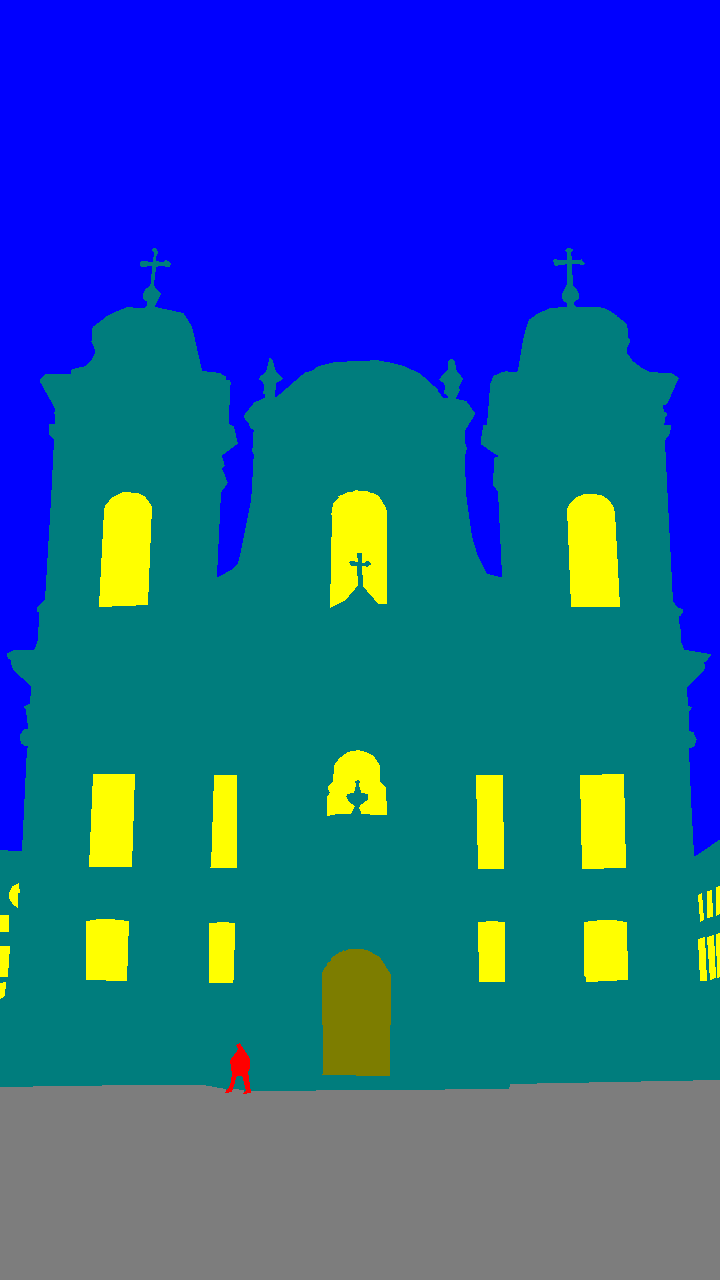}
	&
		\includegraphics[height=0.1\textheight]{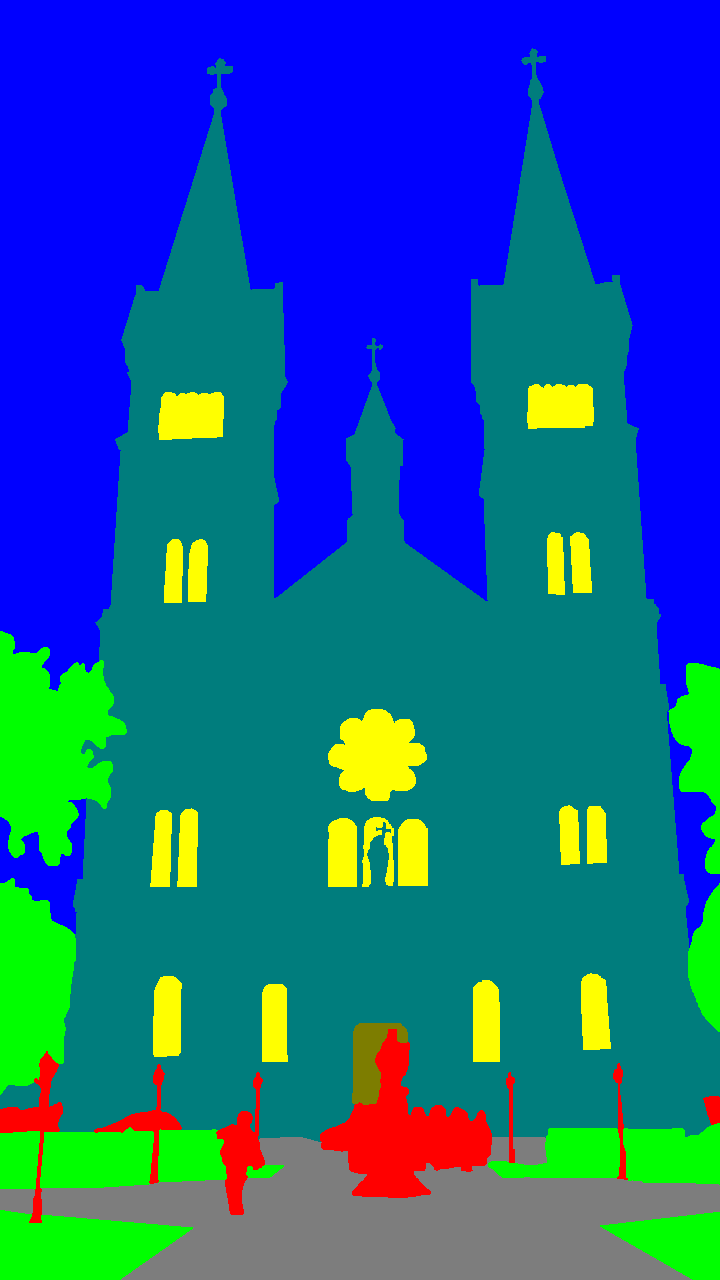}
	&
		\includegraphics[height=0.1\textheight]{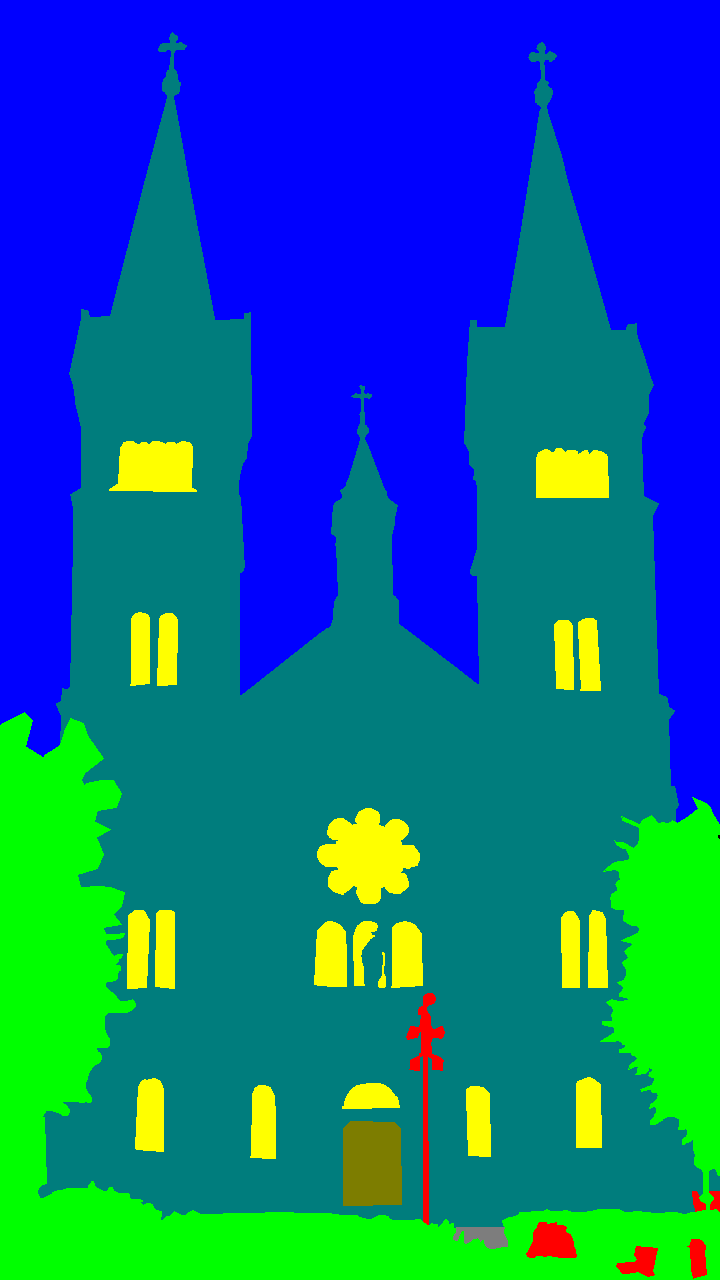}
	&
		\includegraphics[height=0.1\textheight]{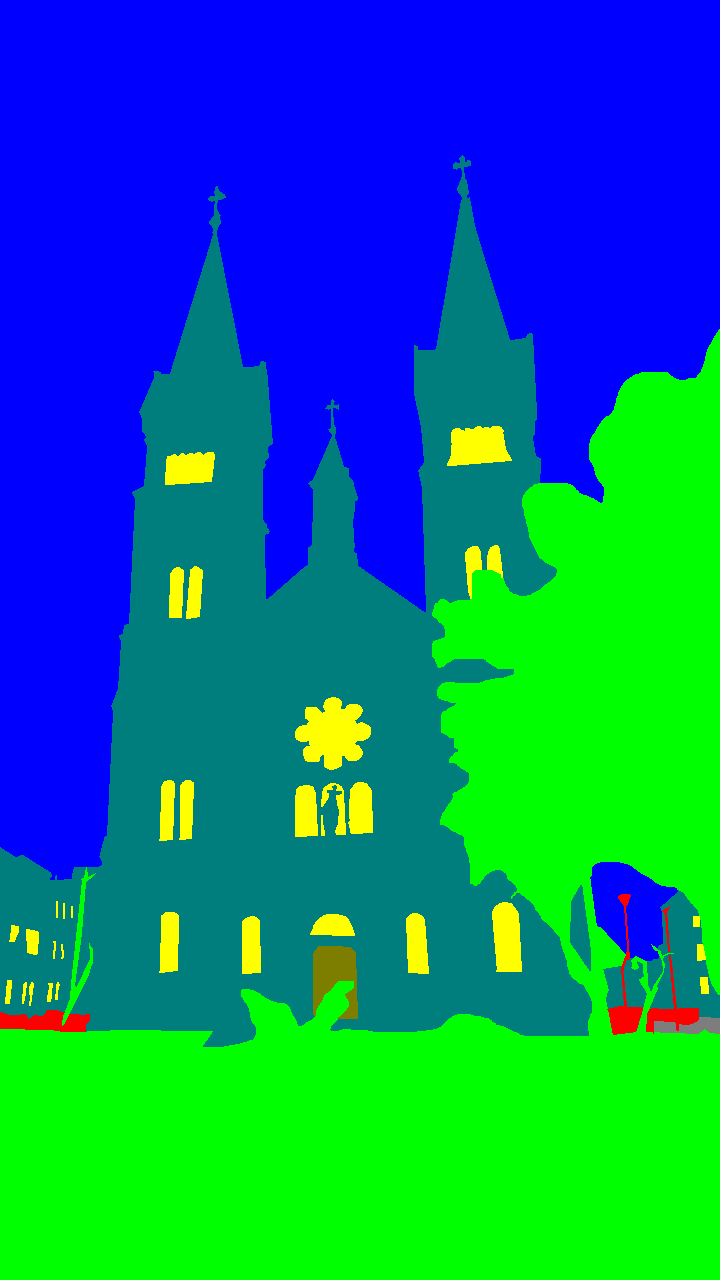}
	&
		\includegraphics[height=0.1\textheight]{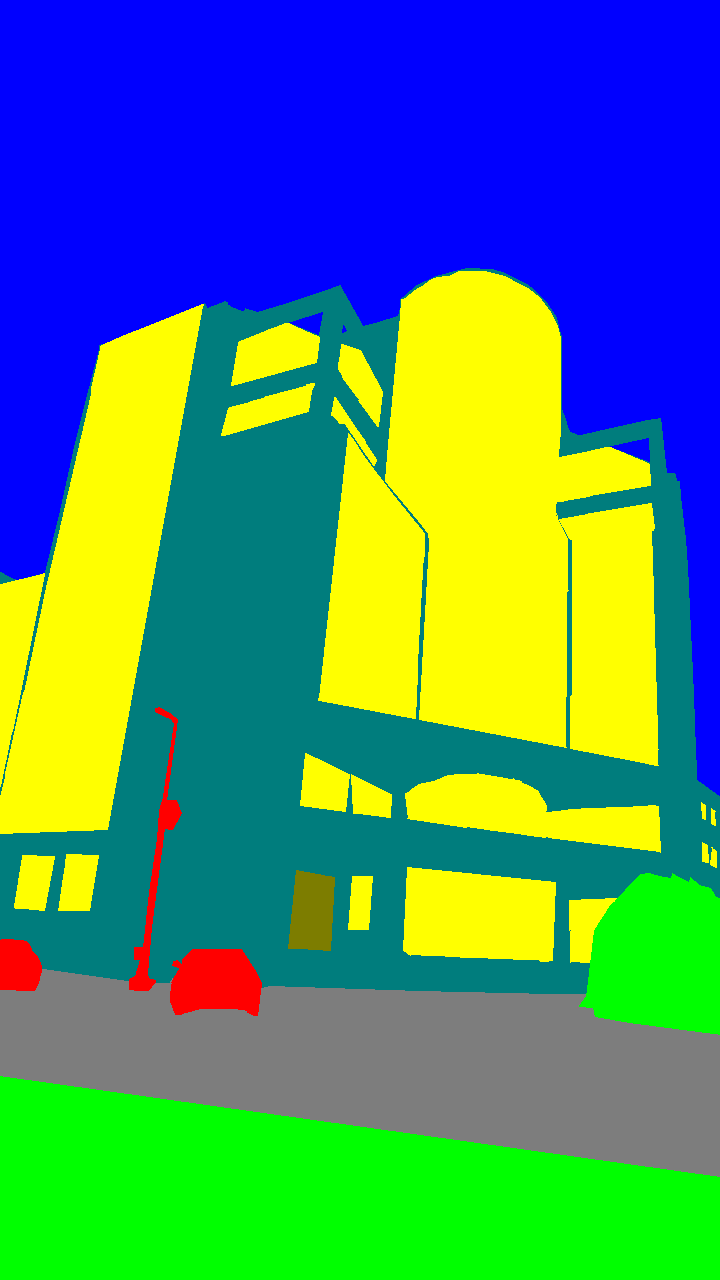}
	&
		\includegraphics[height=0.1\textheight]{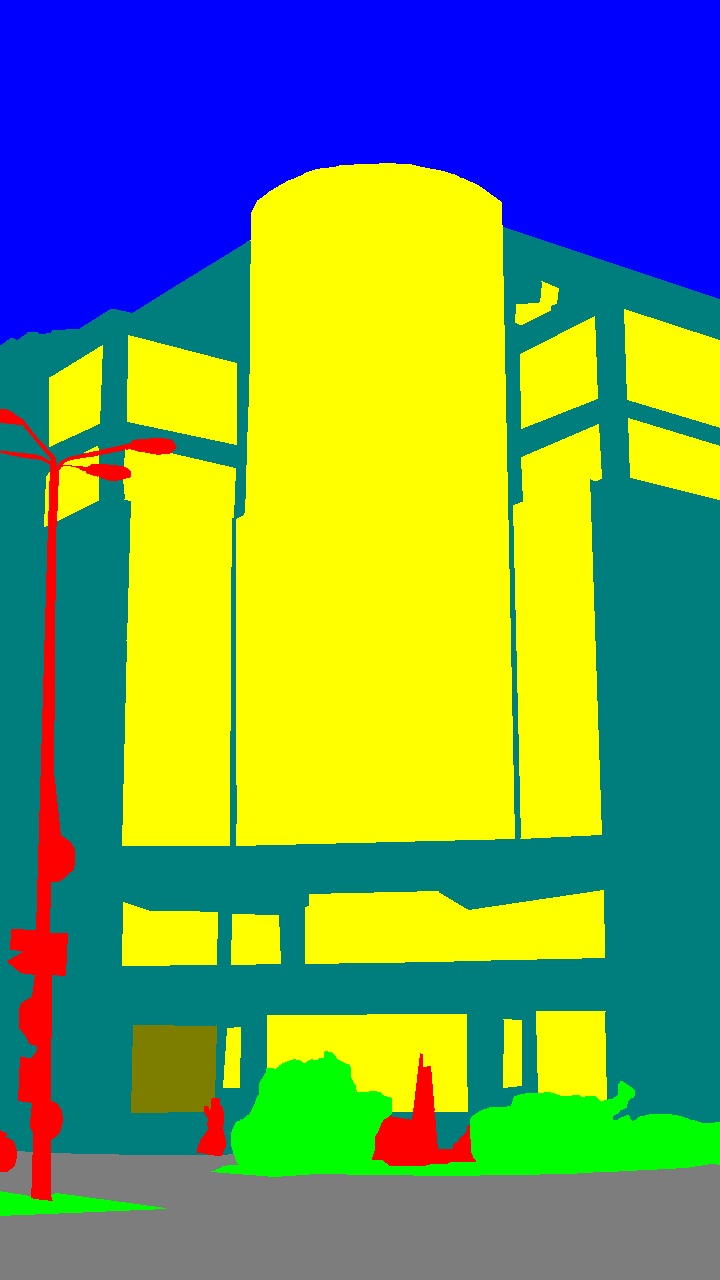}
	&
		\includegraphics[height=0.1\textheight]{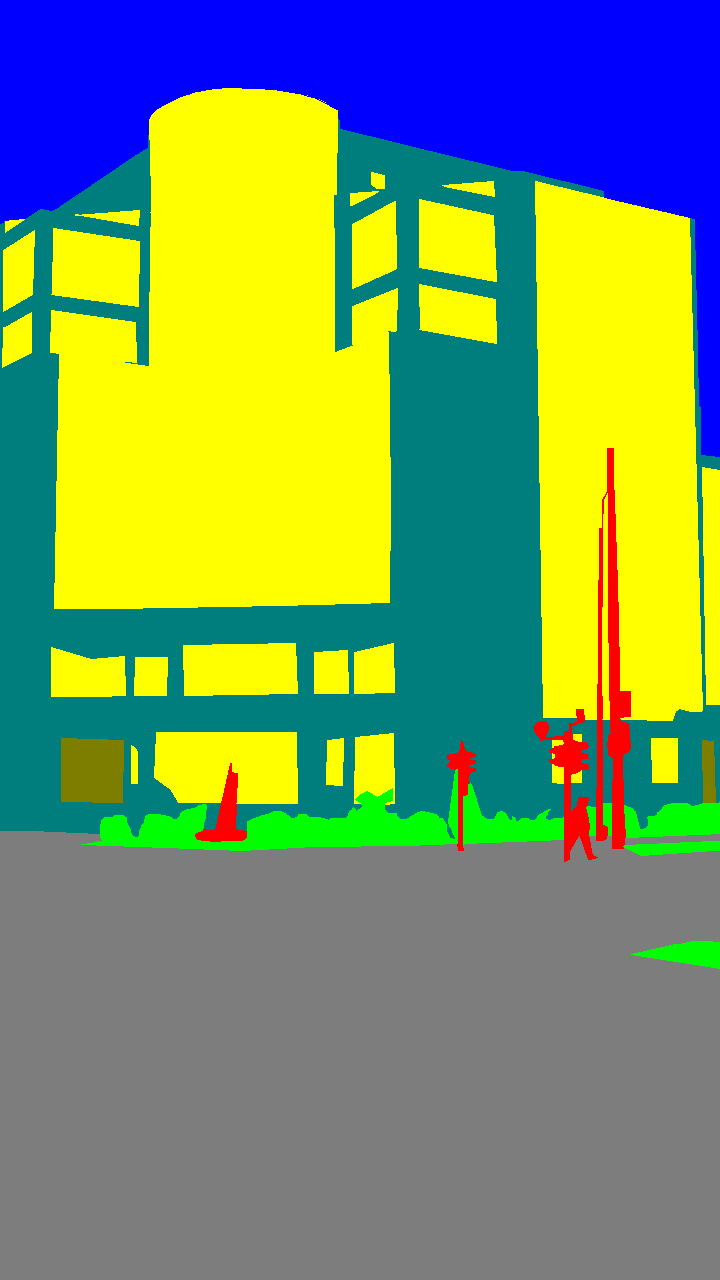}
\\
		\includegraphics[height=0.1\textheight]{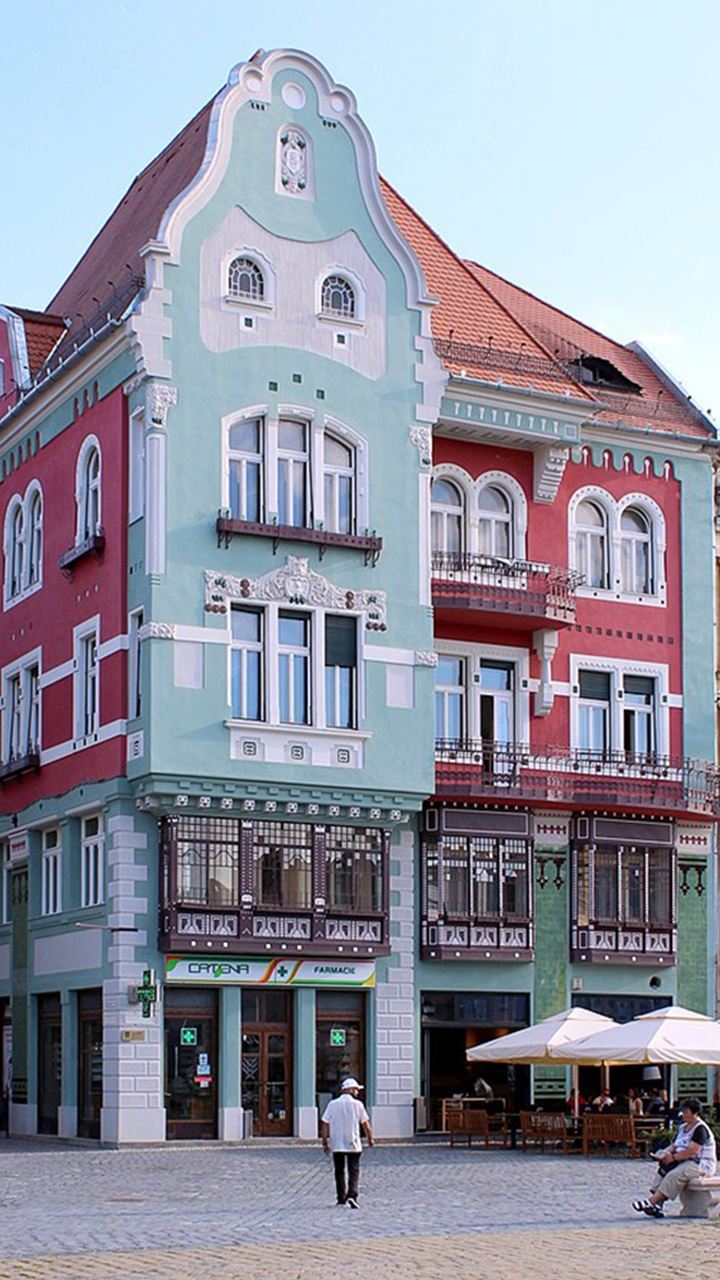}
	&
		\includegraphics[height=0.1\textheight]{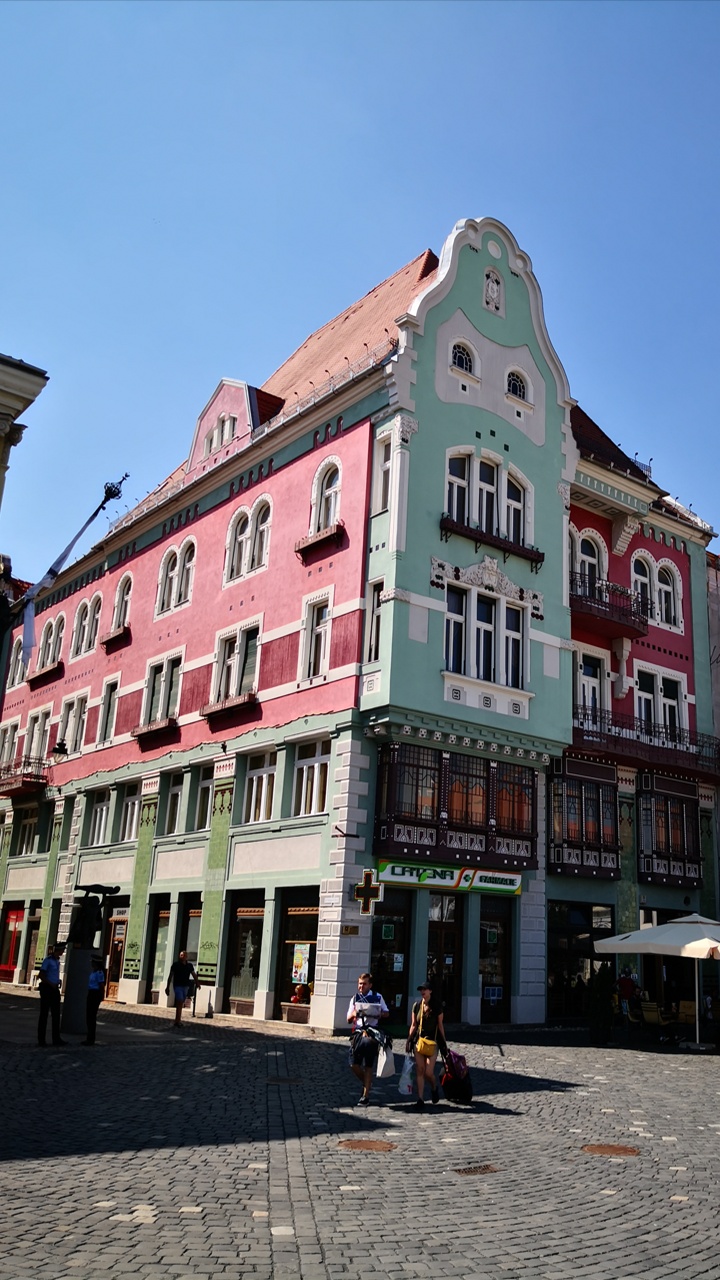}
	&
	    \includegraphics[height=0.1\textheight]{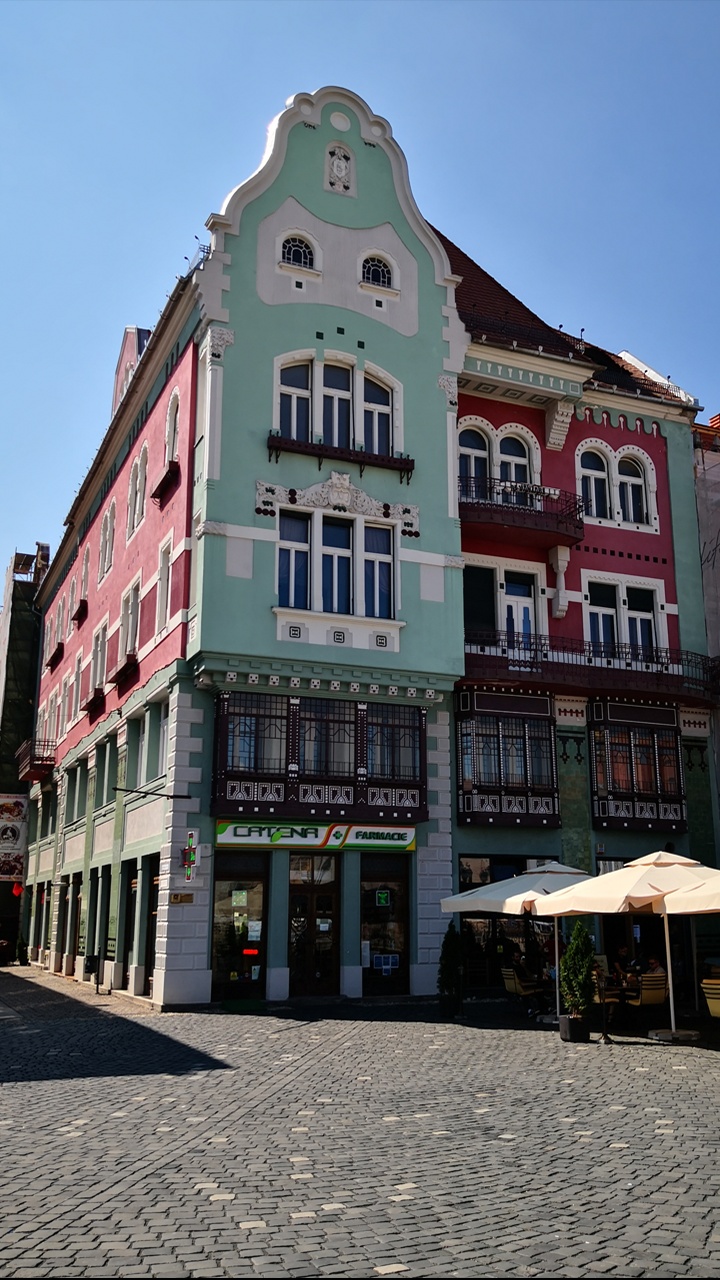}
	&
		\includegraphics[height=0.1\textheight]{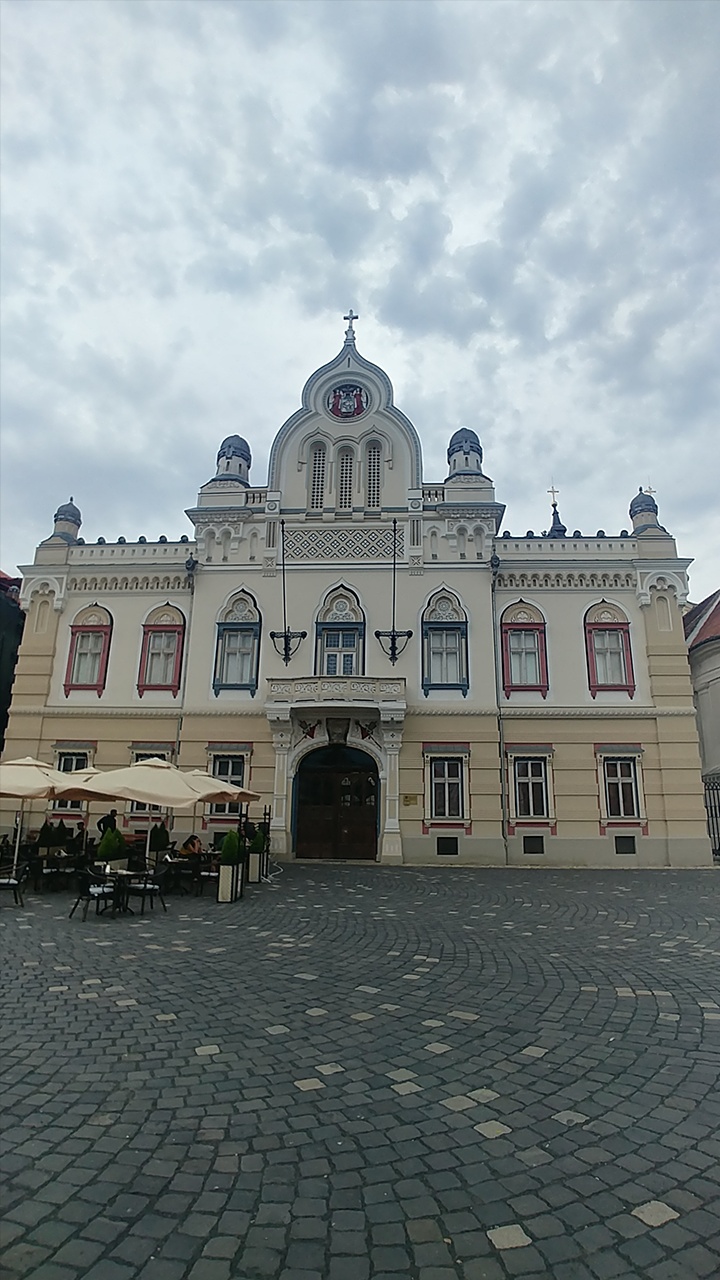}
	&
		\includegraphics[height=0.1\textheight]{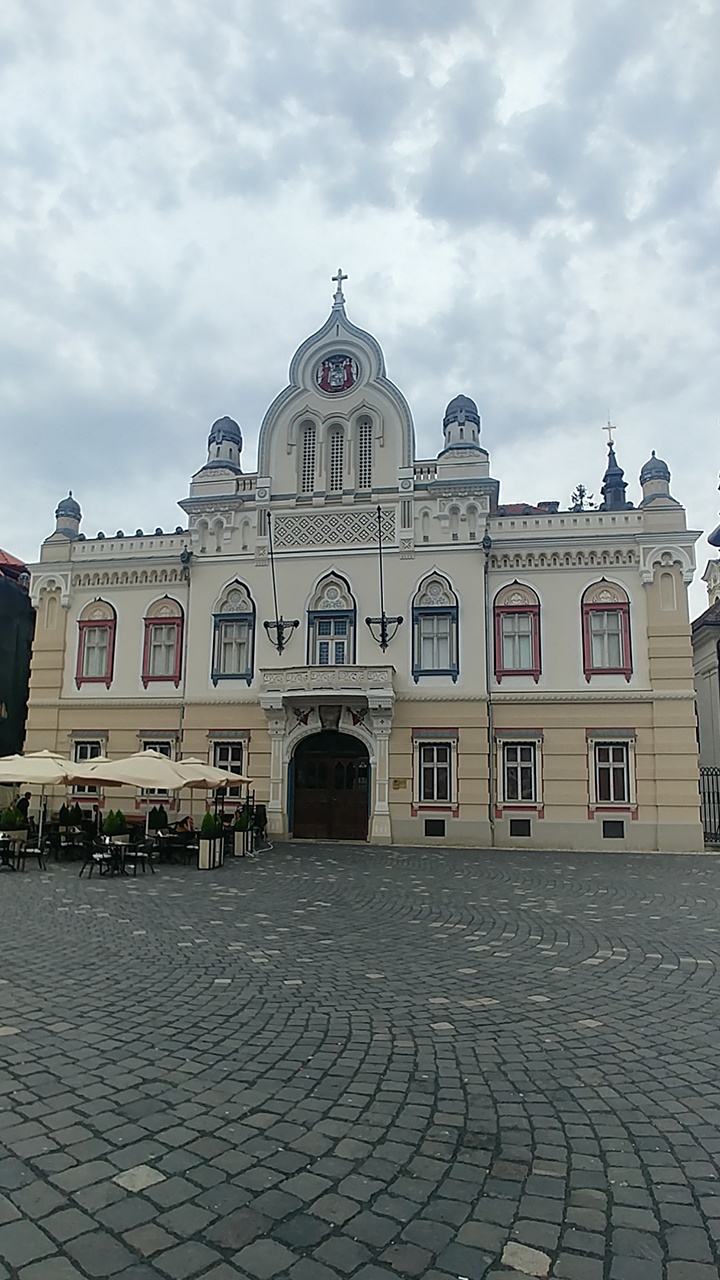}
	&
		\includegraphics[height=0.1\textheight]{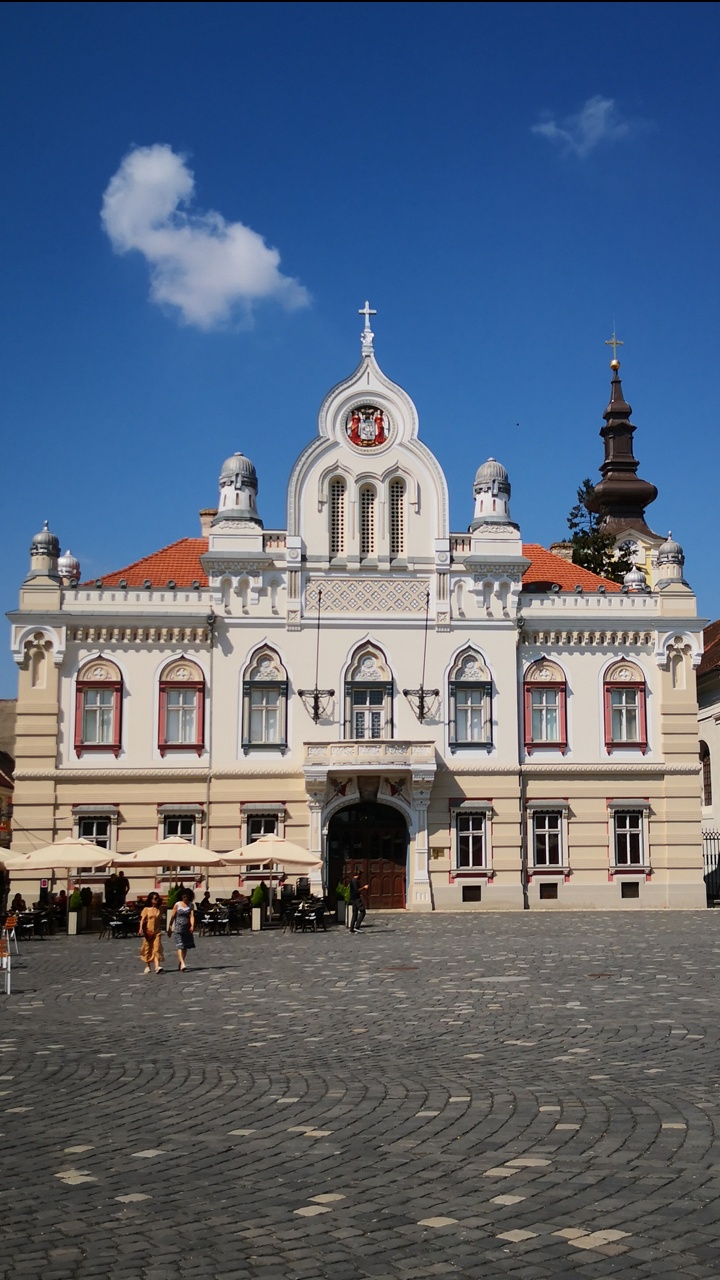}
	&
		\includegraphics[height=0.1\textheight]{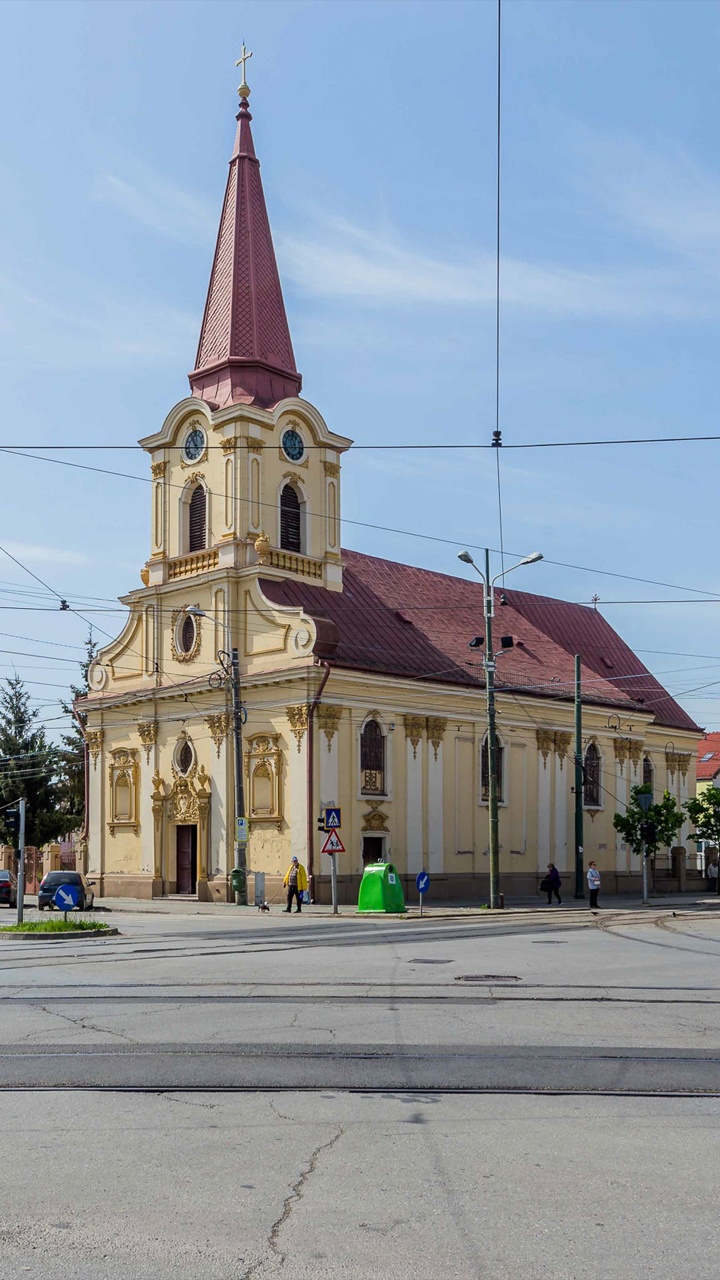}
	&
		\includegraphics[height=0.1\textheight]{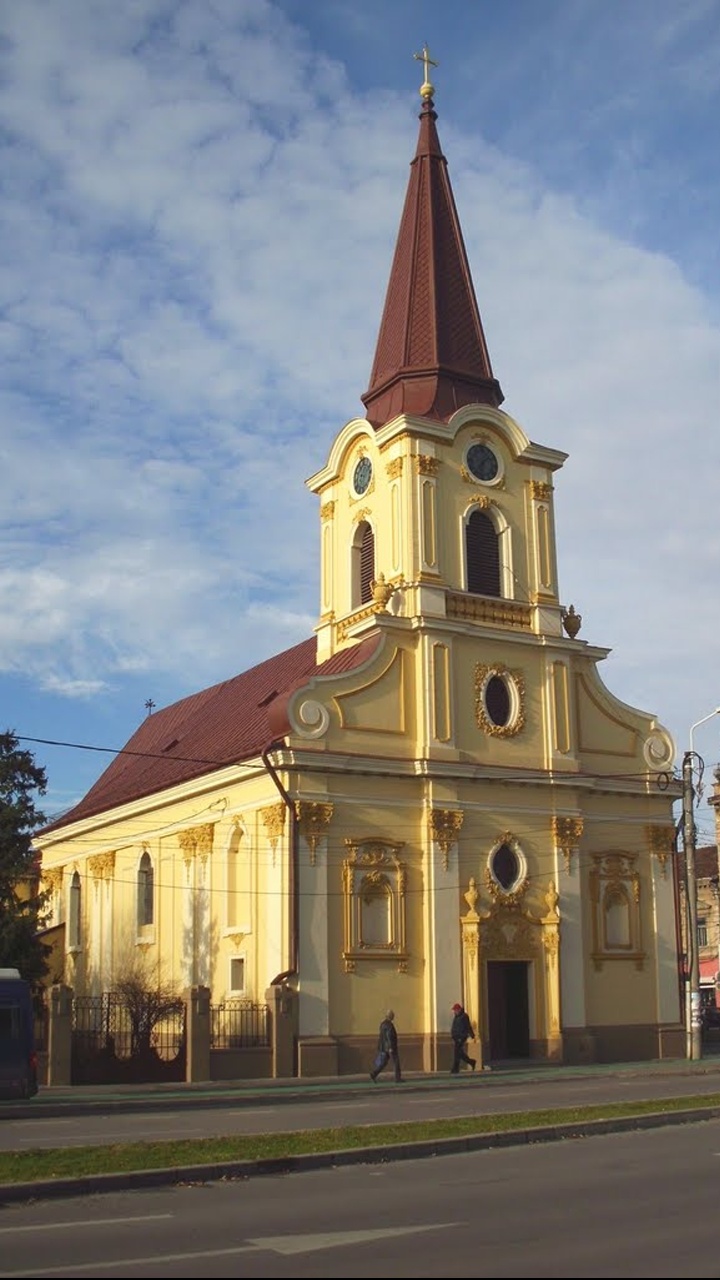}
	&
		\includegraphics[height=0.1\textheight]{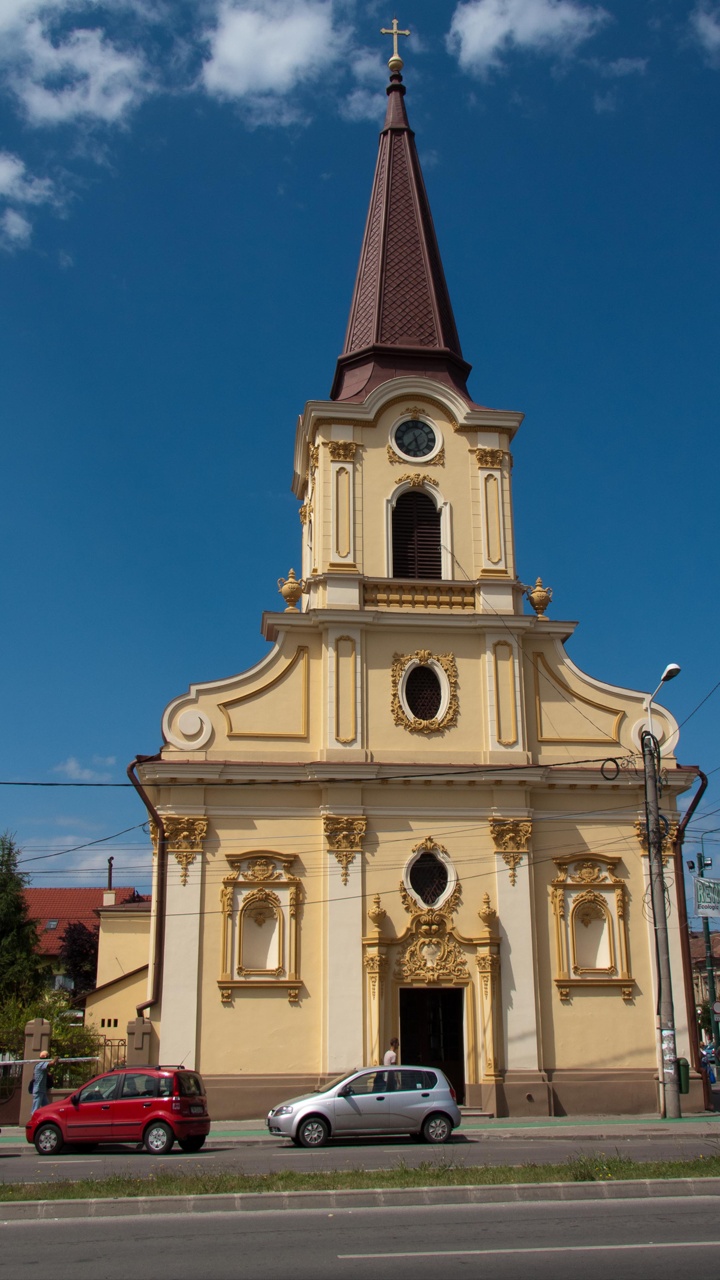}
	&
		\includegraphics[height=0.1\textheight]{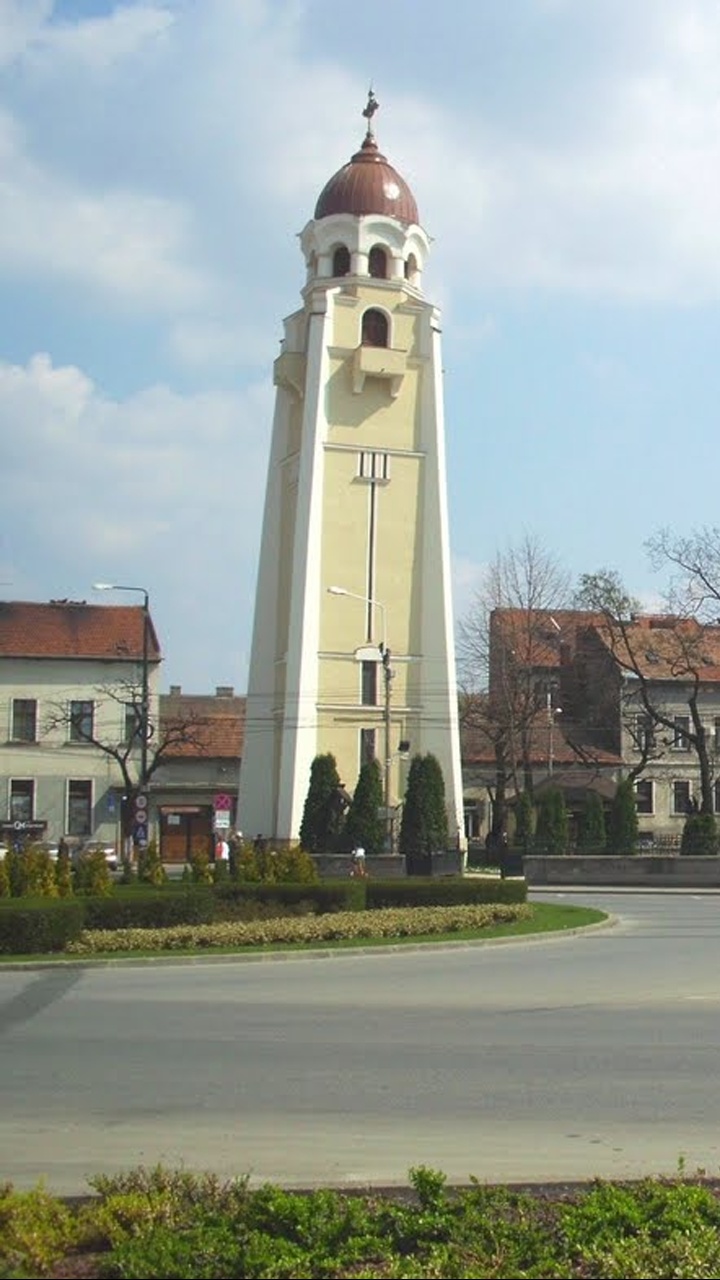}
\\
		\includegraphics[height=0.1\textheight]{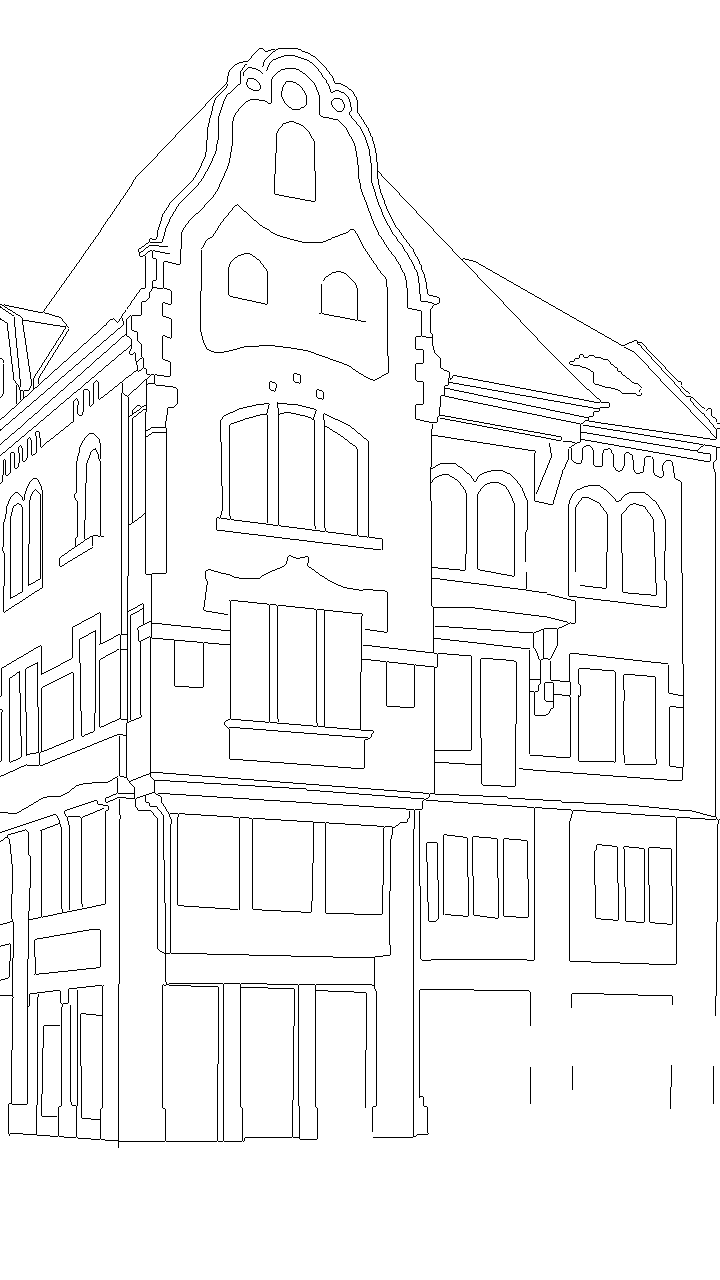}
	&
		\includegraphics[height=0.1\textheight]{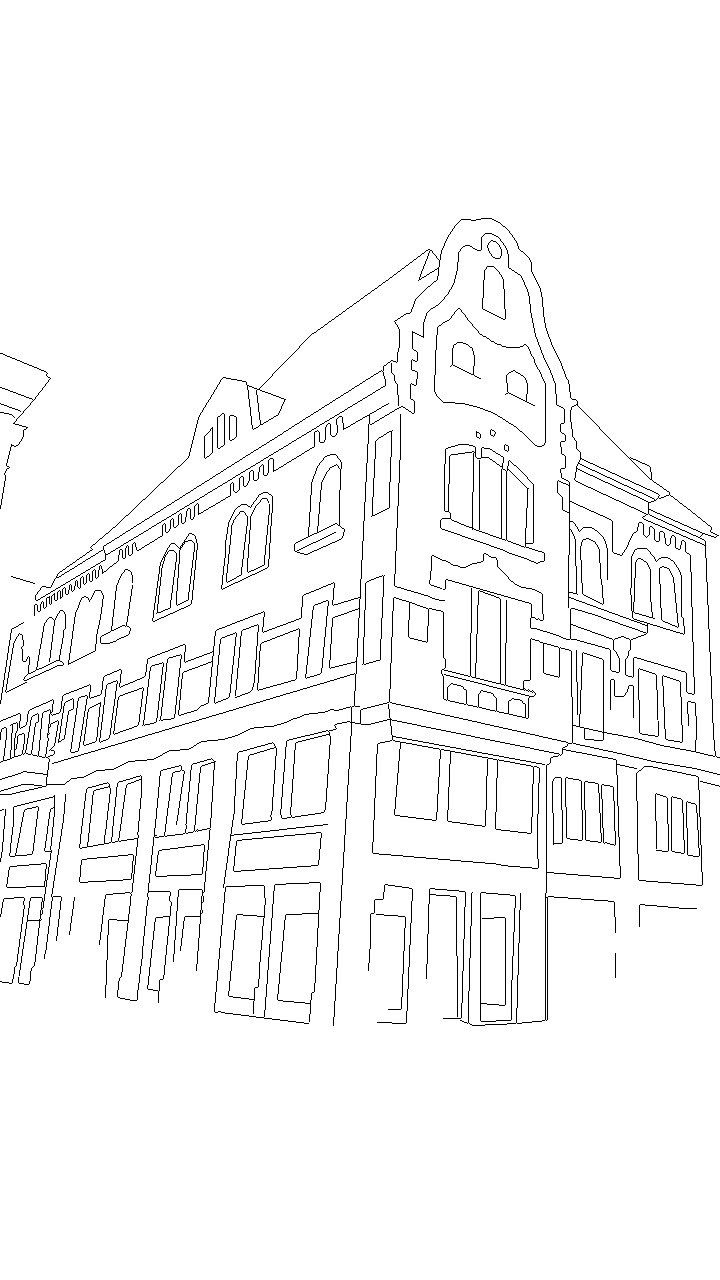}
	&
	    \includegraphics[height=0.1\textheight]{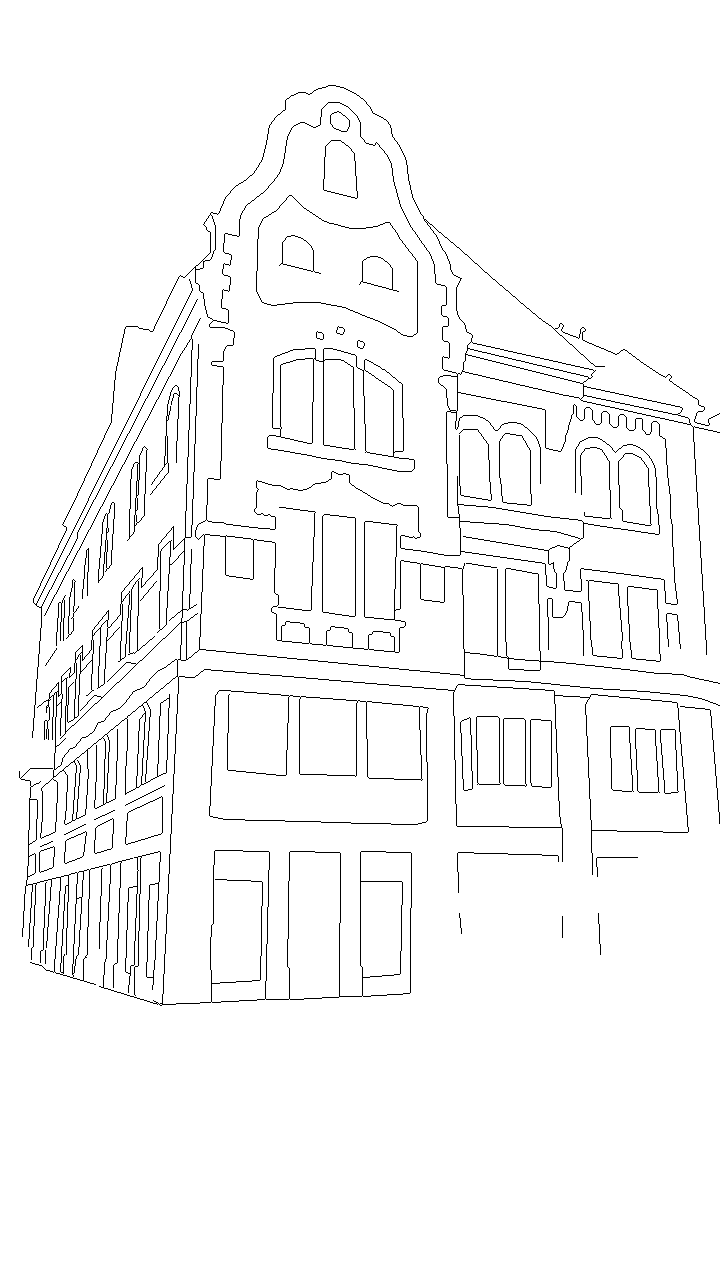}
	&
		\includegraphics[height=0.1\textheight]{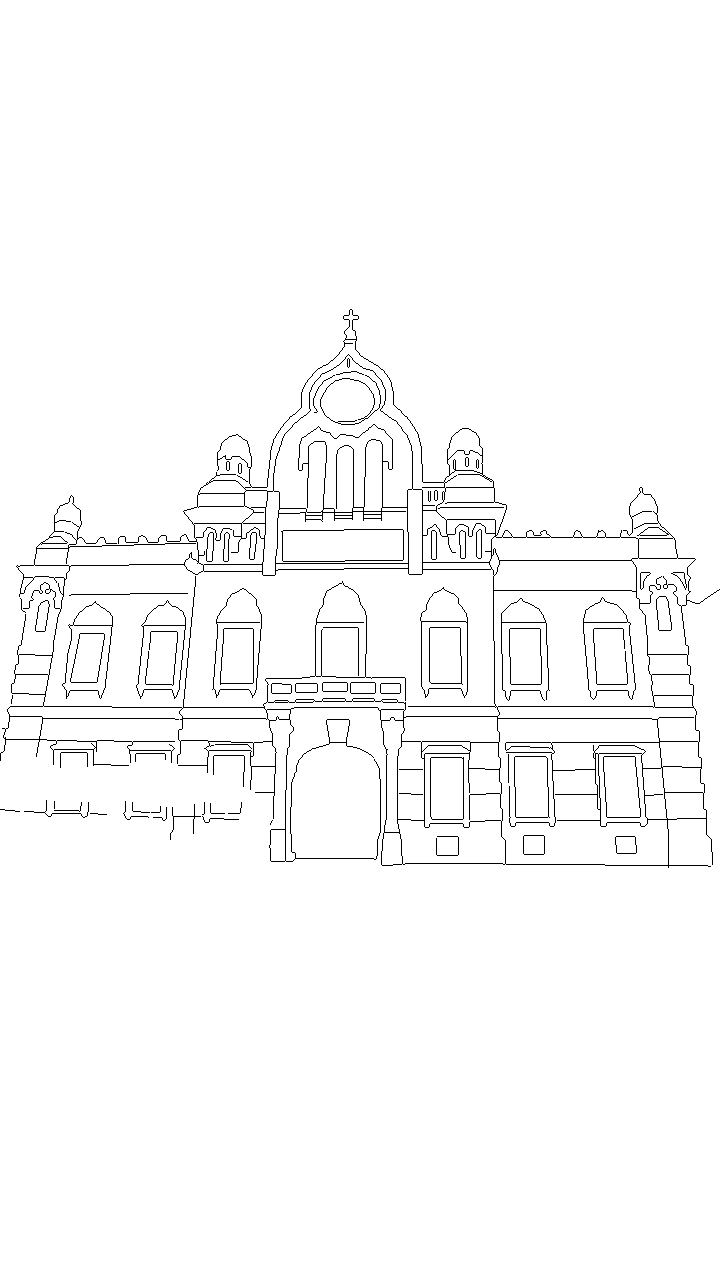}
	&
		\includegraphics[height=0.1\textheight]{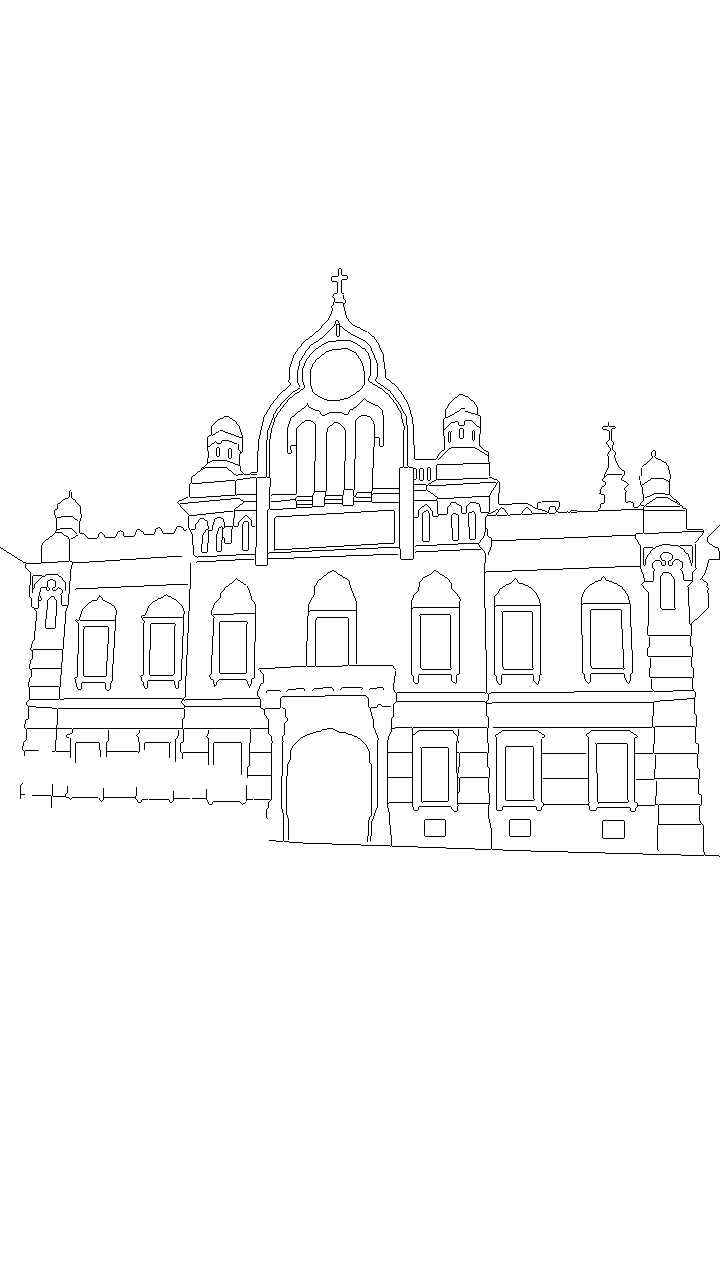}
	&
		\includegraphics[height=0.1\textheight]{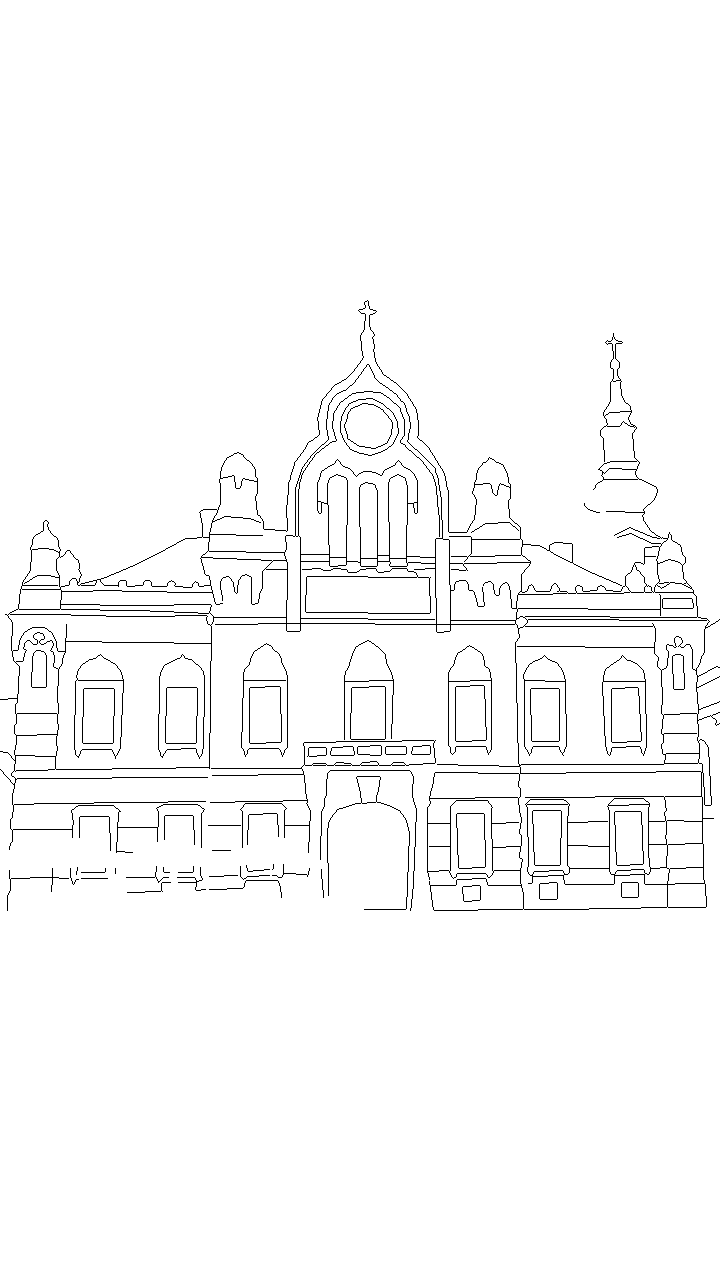}
	&
		\includegraphics[height=0.1\textheight]{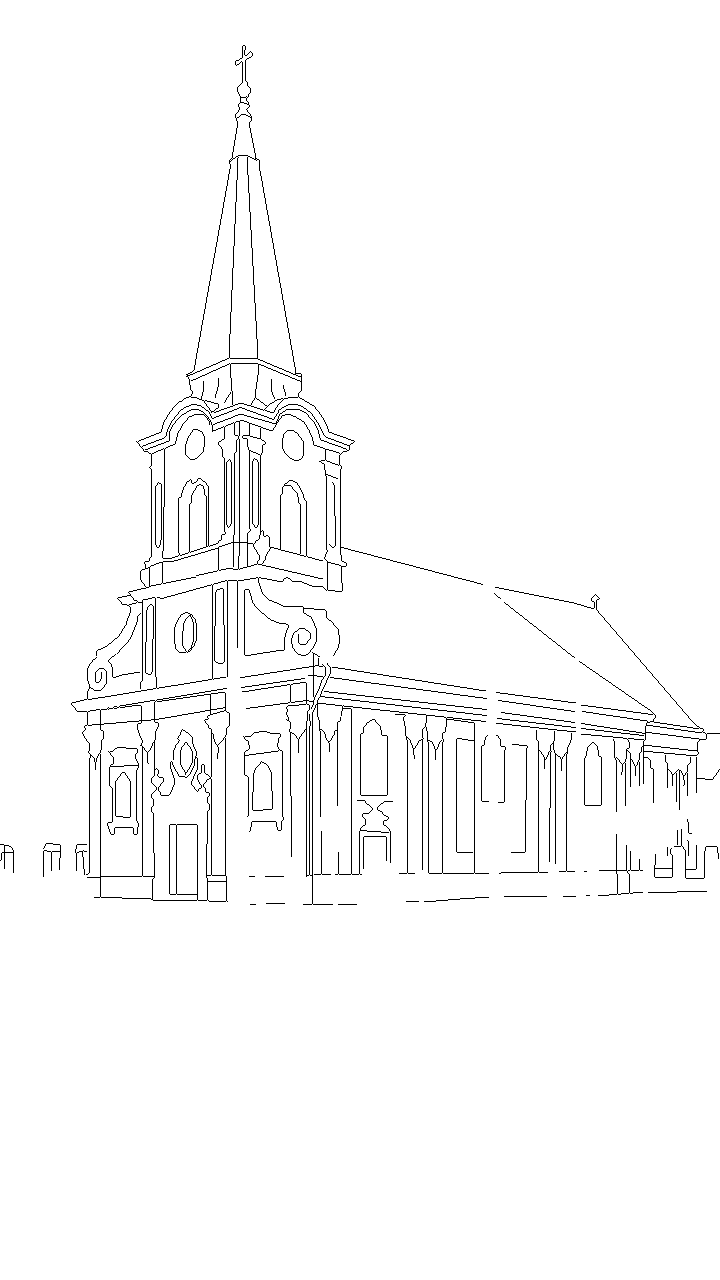}
	&
		\includegraphics[height=0.1\textheight]{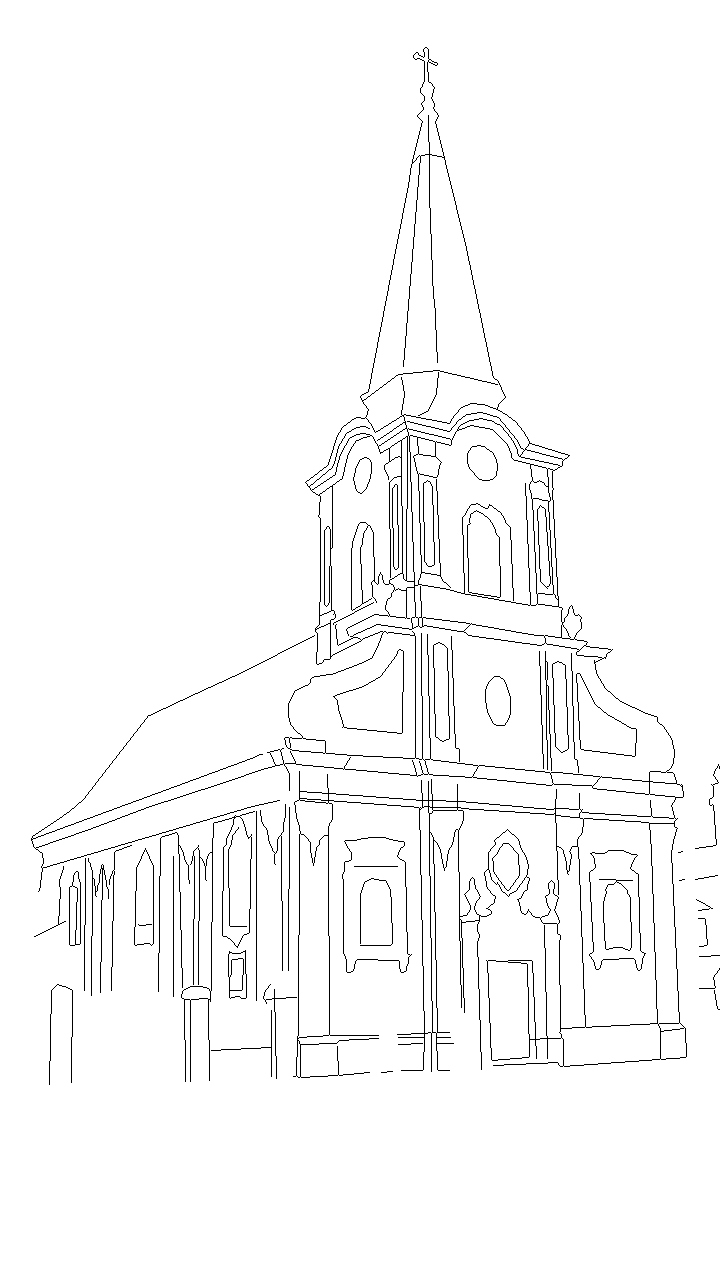}
	&
		\includegraphics[height=0.1\textheight]{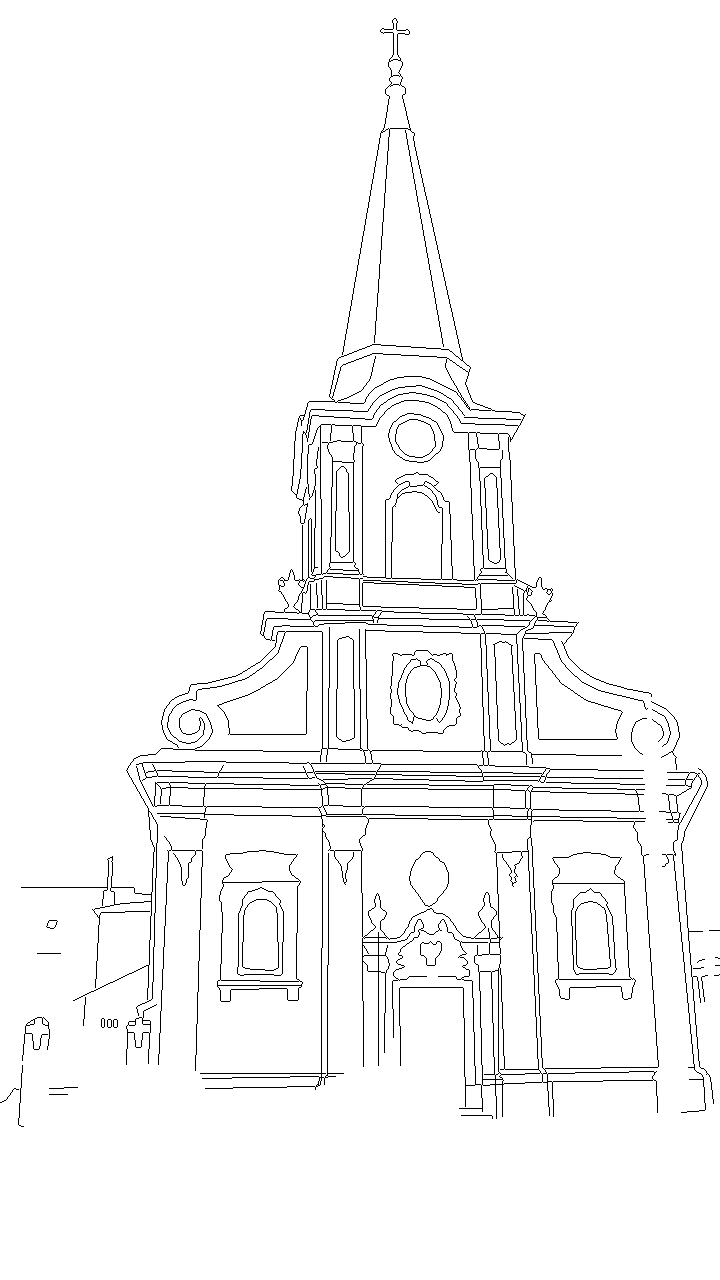}
	&
		\includegraphics[height=0.1\textheight]{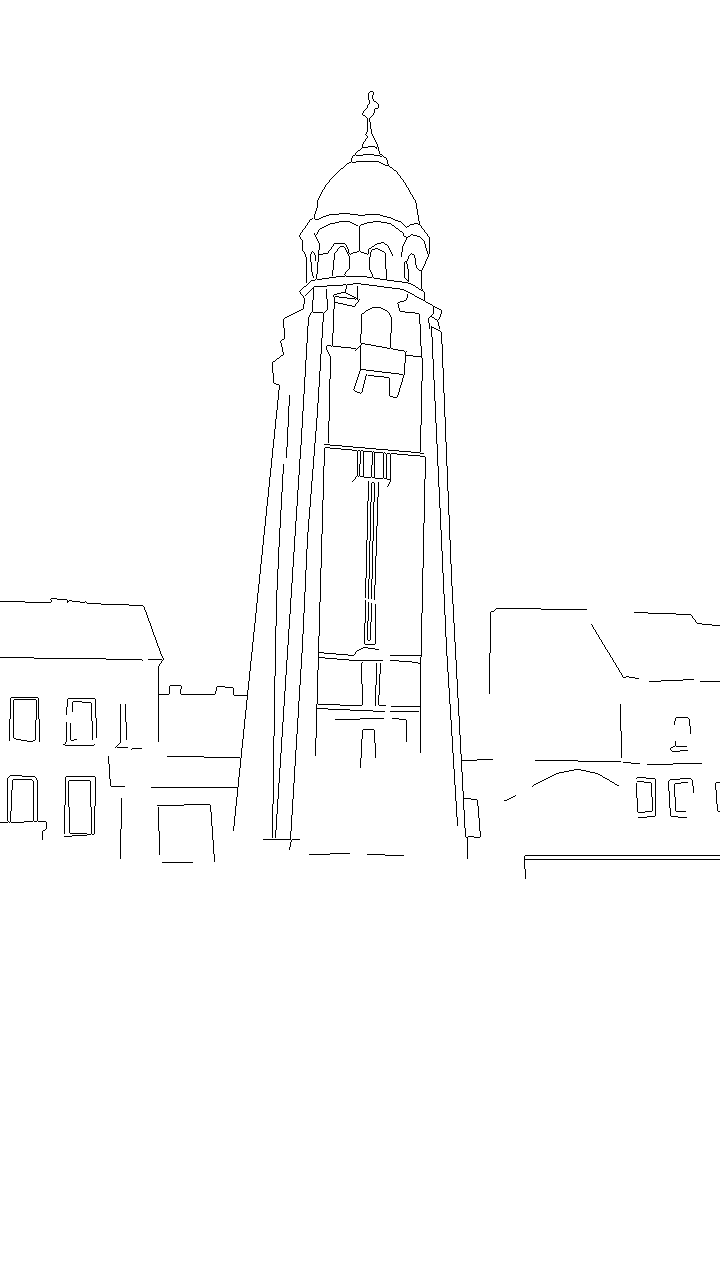}
\\
		\includegraphics[height=0.1\textheight]{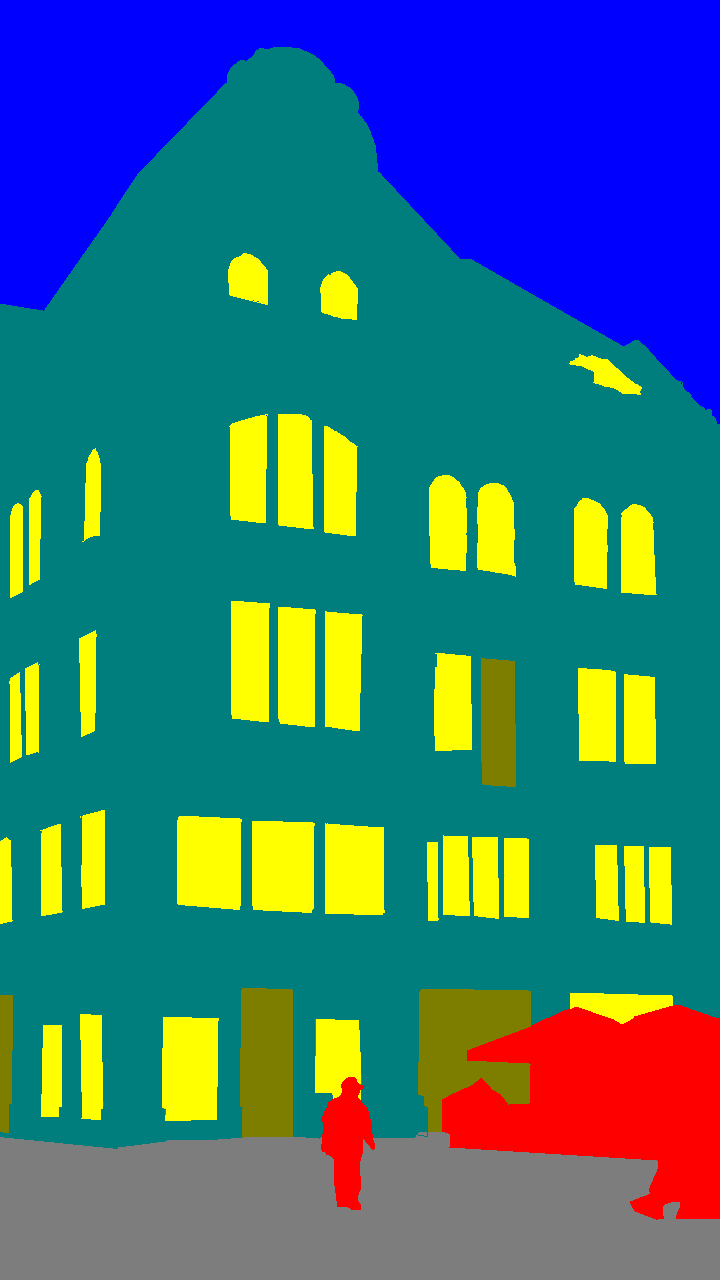}
	&
		\includegraphics[height=0.1\textheight]{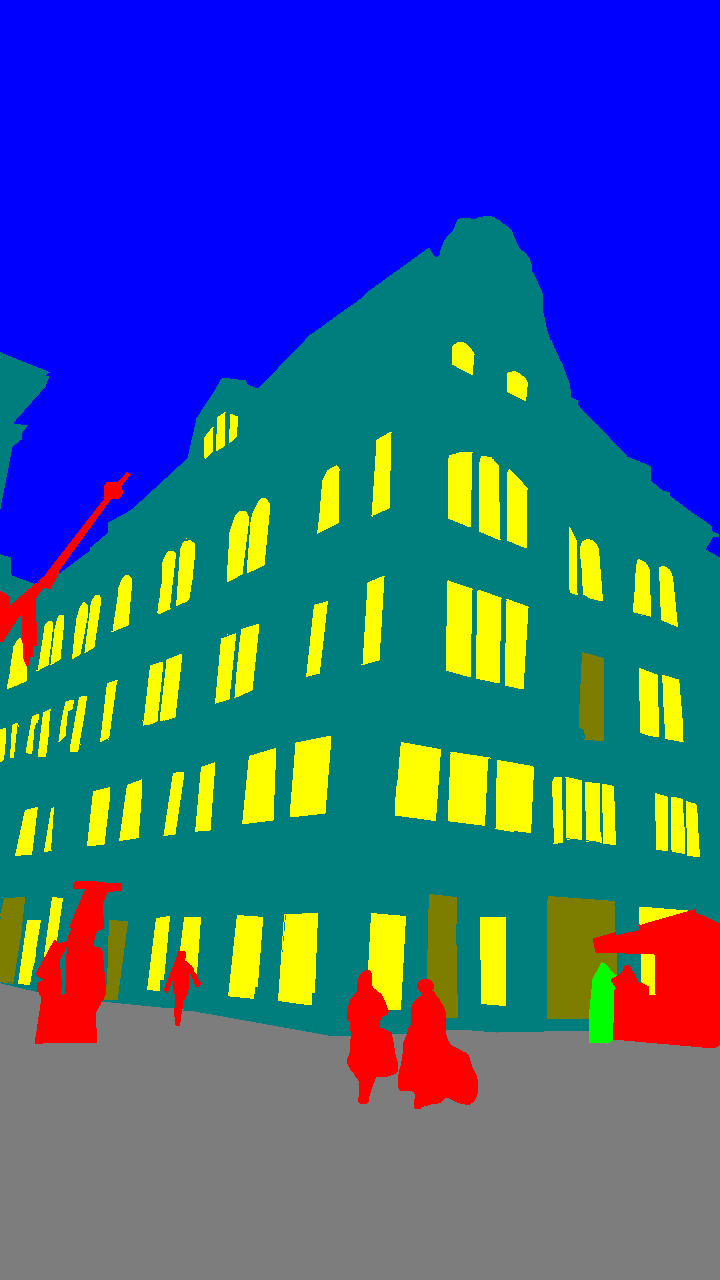}
	&
	    \includegraphics[height=0.1\textheight]{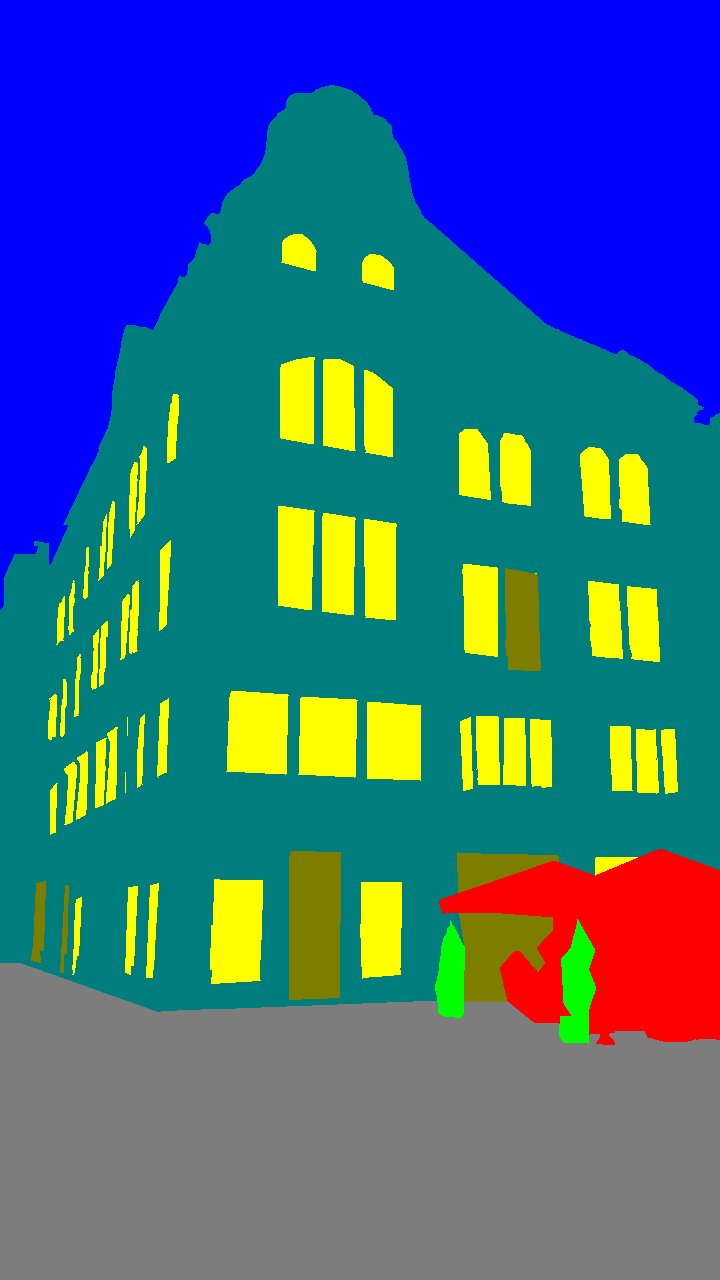}
	&
		\includegraphics[height=0.1\textheight]{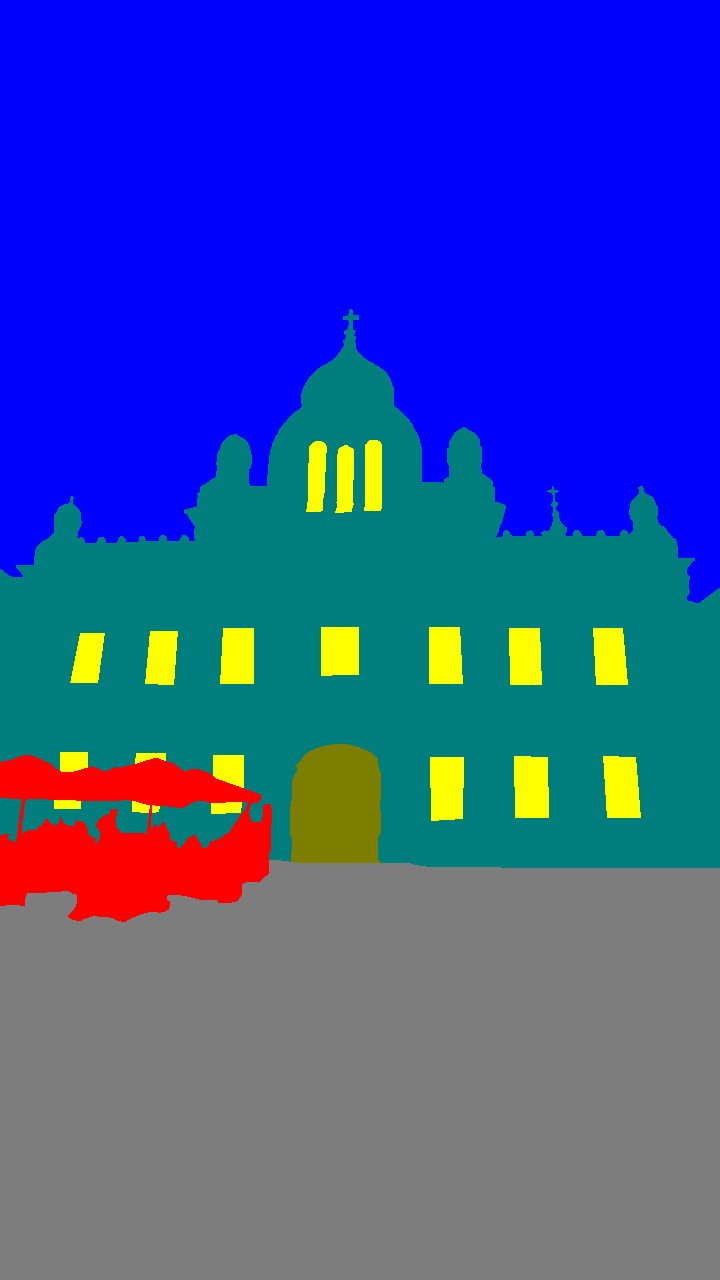}
	&
		\includegraphics[height=0.1\textheight]{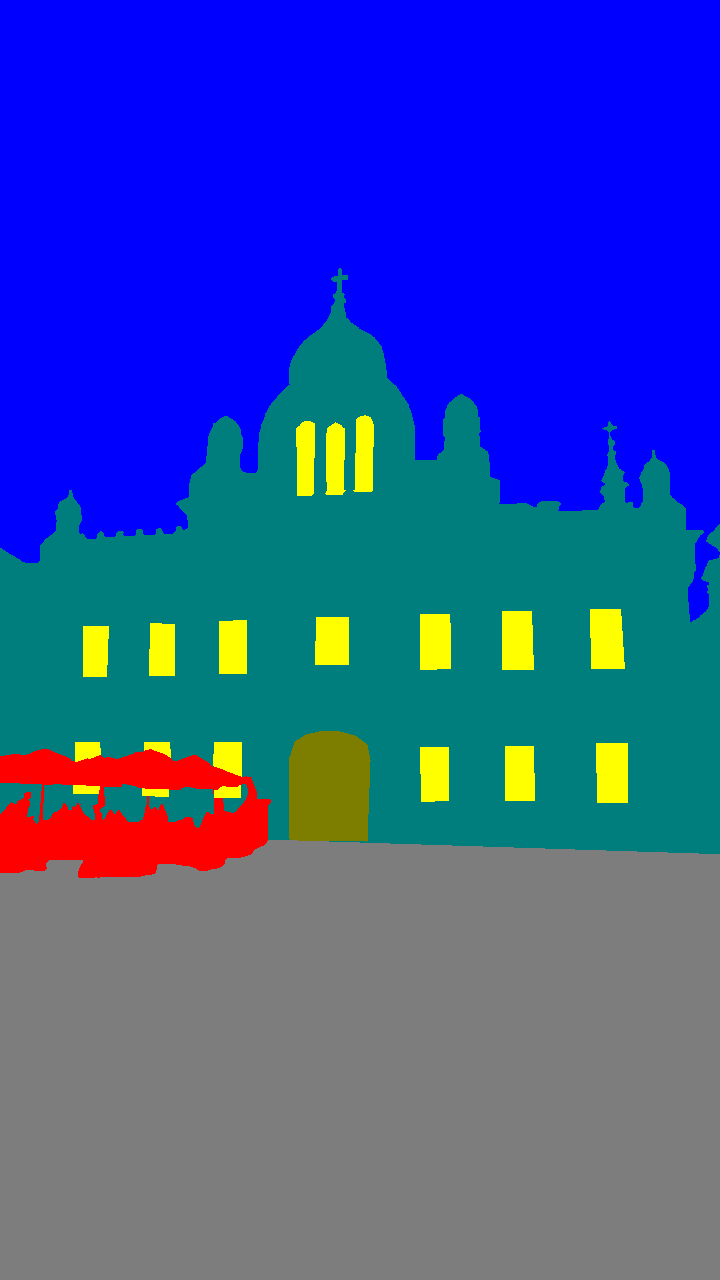}
	&
		\includegraphics[height=0.1\textheight]{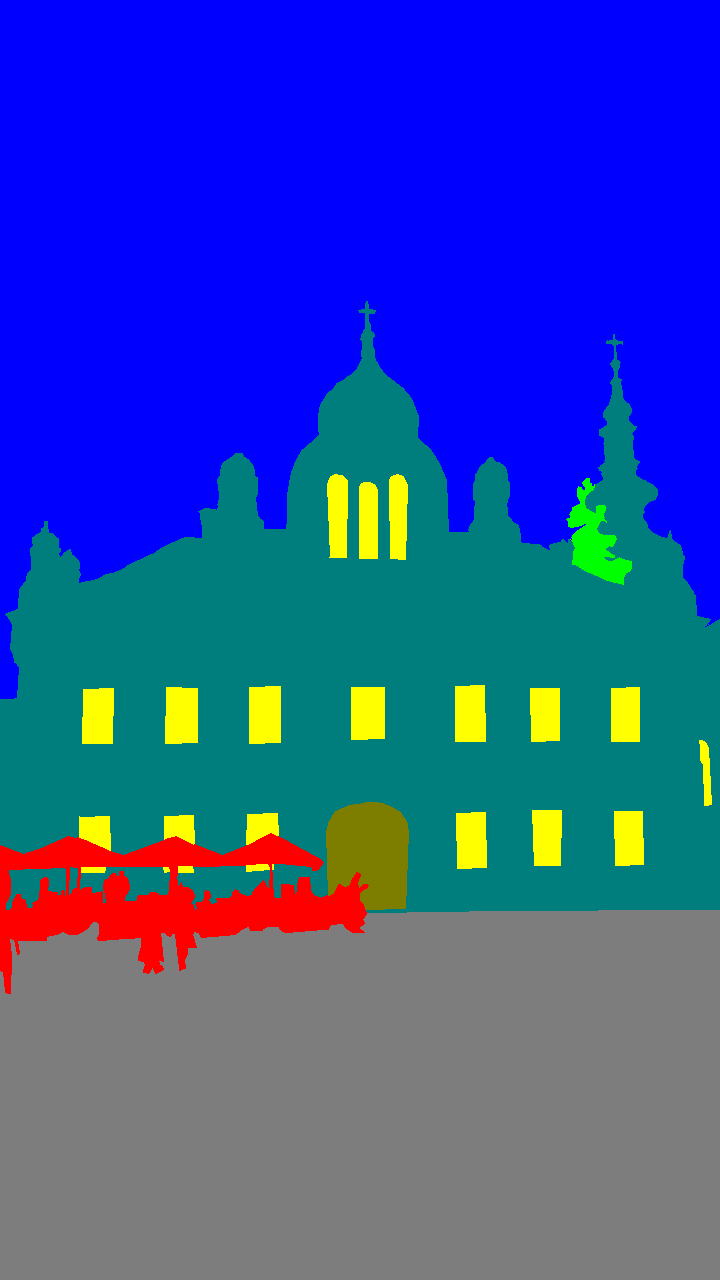}
	&
		\includegraphics[height=0.1\textheight]{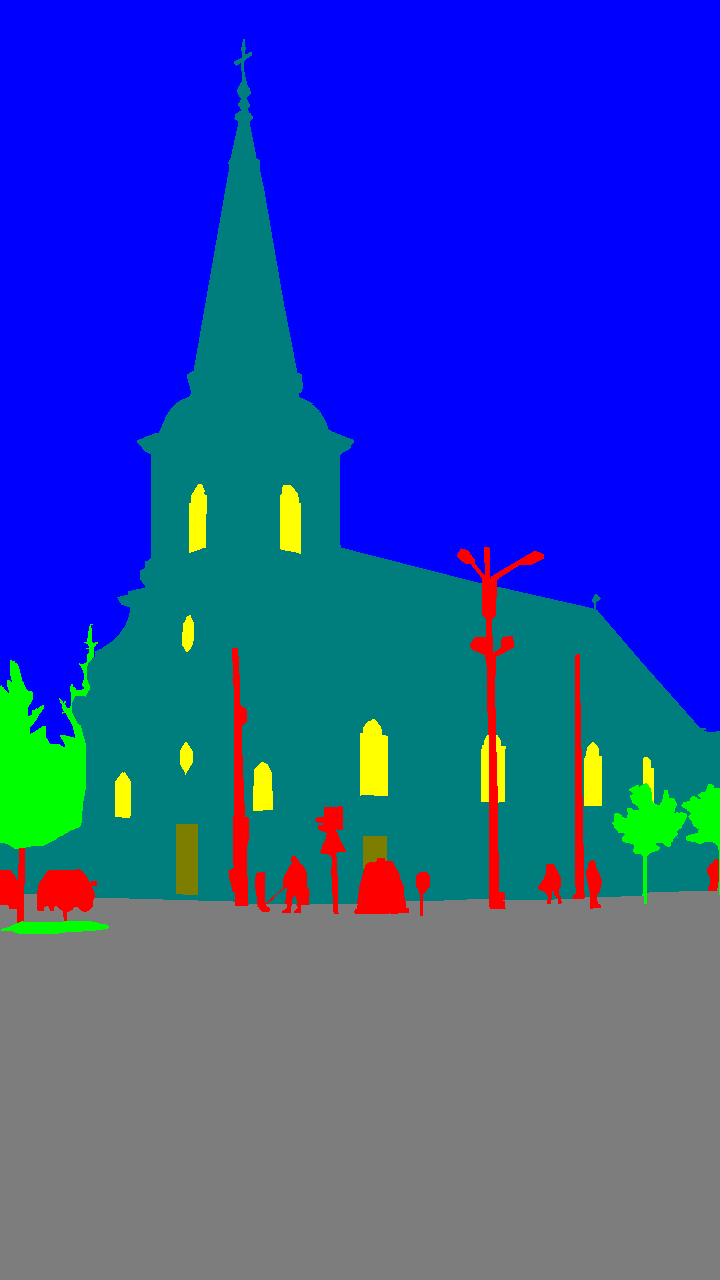}
	&
		\includegraphics[height=0.1\textheight]{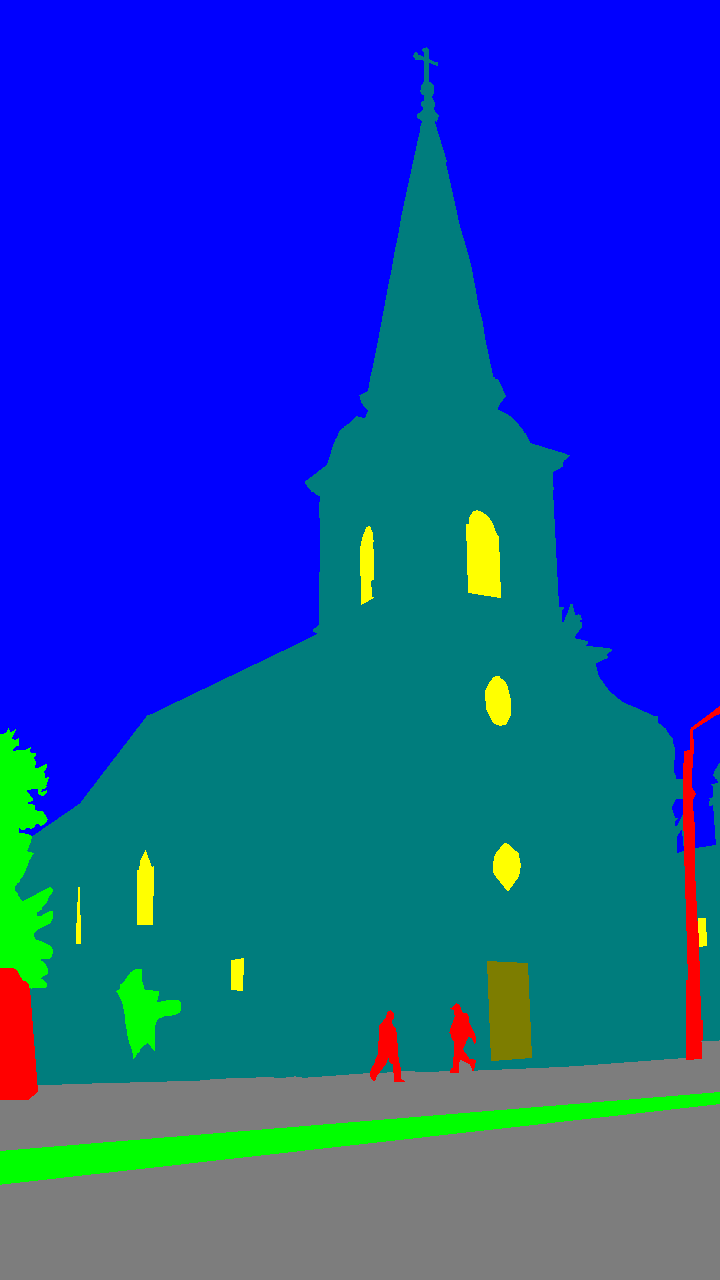}
	&
		\includegraphics[height=0.1\textheight]{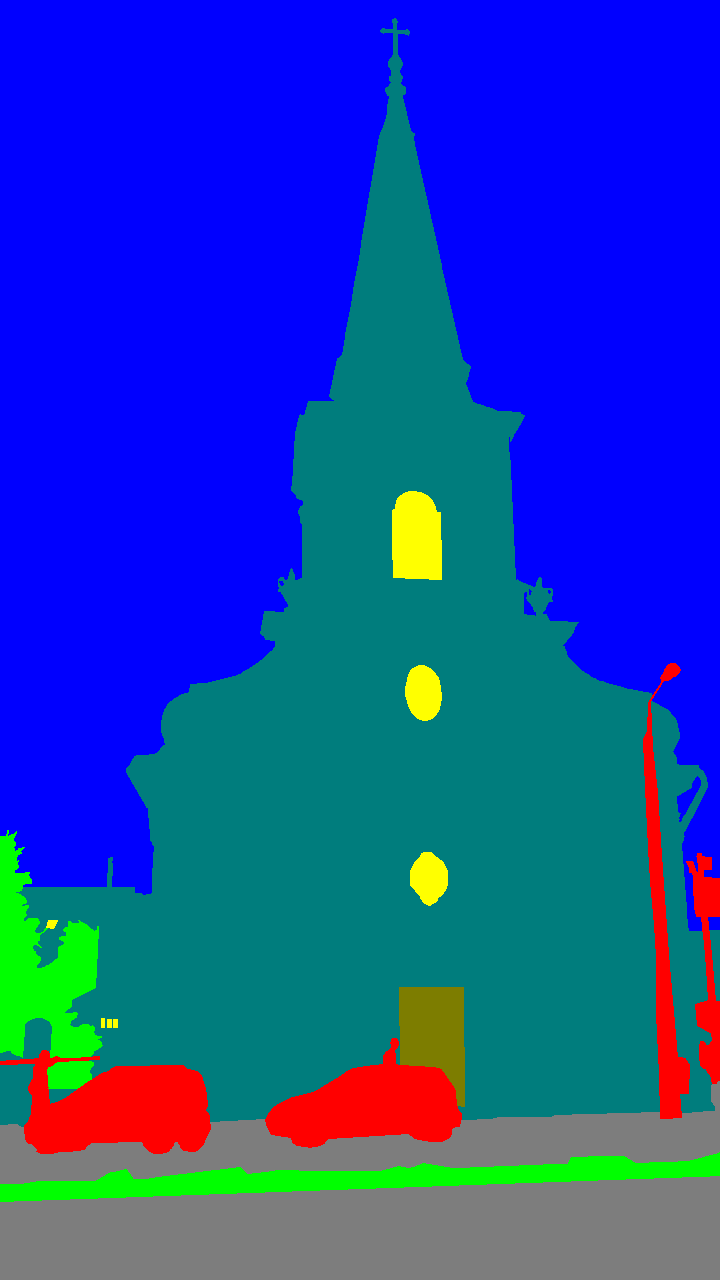}
	&
		\includegraphics[height=0.1\textheight]{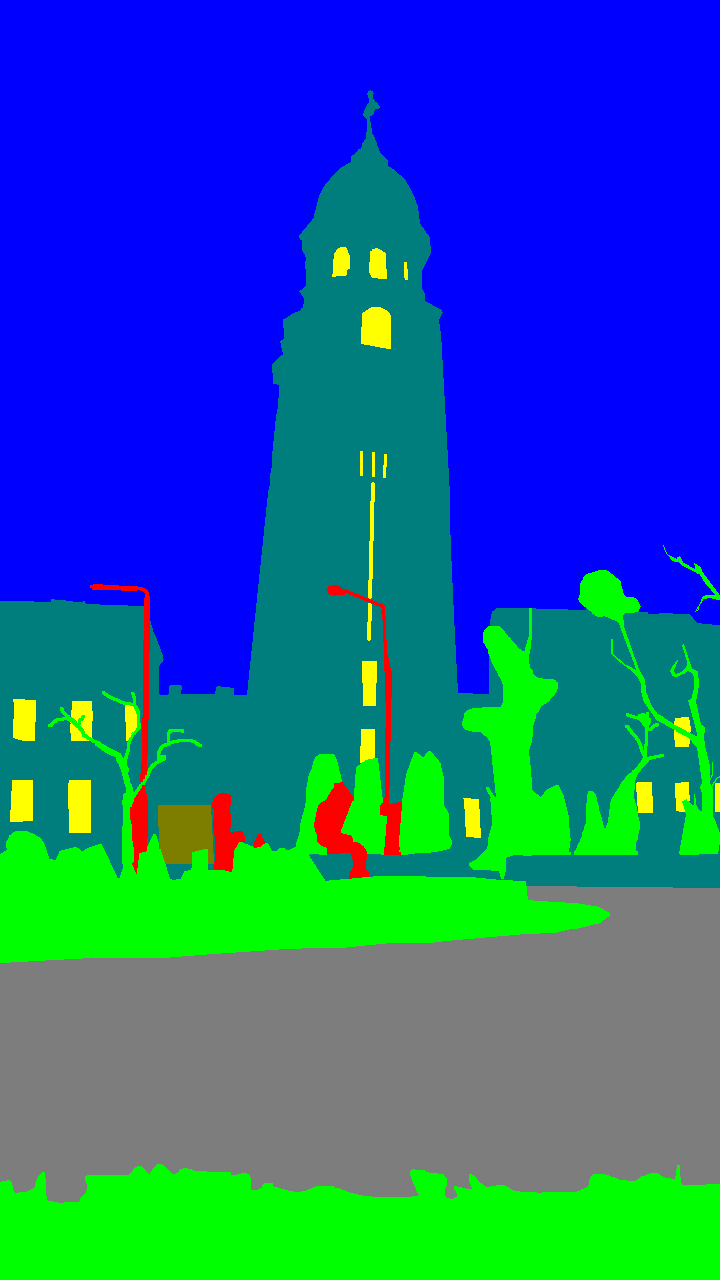}

\end{tabular}
\caption{Images from proposed dataset. Rows: original image, edge ground-truth, label ground-truth }
\label{fig:other_img}
\end{figure}%



\end{document}